\documentclass[twoside,11pt]{article}
 
\usepackage[preprint]{jmlr2e}

\usepackage{amsmath,amsfonts}

\usepackage{bm}
\usepackage{graphicx}
\usepackage{epstopdf}
\usepackage{color}
\usepackage{float}
\usepackage{pifont}
\usepackage{algorithm}
\usepackage{algpseudocode}

\makeatother

\usepackage{thmtools}
\usepackage{thm-restate}

\usepackage{enumitem}

\usepackage{titletoc}

\usepackage{endnotes}
\usepackage{setspace}
\usepackage{multirow}
\usepackage{url}
\usepackage{mathtools}

\usepackage{multicol}
\usepackage[table]{colortbl}
\usepackage{wrapfig}
\usepackage{footnote}
\makesavenoteenv{table}

\allowdisplaybreaks[4]

\numberwithin{theorem}{section}

\numberwithin{equation}{section}

\usepackage{booktabs}
\usepackage{stfloats}

\newcommand{\thetaAlgBeta}{\theta^{ \mathrm{B}}}
\newcommand{\alphaAlgBeta}{\alpha^{ \mathrm{B}}}
\newcommand{\betaAlgBeta}{\beta^{ \mathrm{B}}}
\newcommand{\rmKL}{\mathrm{KL}}
\newcommand{\numAlgBeta}{N^{ \mathrm{B}}}
\newcommand{\bmThetaAlgBeta}{{\bm \theta}^{ \mathrm{B}}}

\newcommand{\calA}{\mathcal{A}}

\newcommand{\prHlin}{\frac{d}{t^2}} 
\newcommand{\clin} { c(\lambda) }  
\newcommand{\tlin} { t(\lambda) } 

\newcommand{\algCasLin}{{\sc LinTS-Cascade($\lambda$)}}
\newcommand{\thetaLin}{\rho}
\newcommand{\muLin}{\psi}
\newcommand{\NLin}{M}
\newcommand{\ZLin}{\xi}
\newcommand{\bLin}{b}
\newcommand{\RLin}{}
\newcommand{\Wfeedback}{A}
\newcommand{\stdLin}{\hat{\sigma}}
\newcommand{\cZeroLin}{1}

\newcommand{\calF}{\mathcal{F}}
\newcommand{\bbR}{\mathbb{R}}
\newcommand{\bbE}{\mathbb{E}}

\usepackage{hyperref}

\definecolor{darkgreen}{rgb}{0.0,0.7,0.0}

\usepackage{longtable}
\usepackage{hhline}
\makeatletter
\newcommand{\thickhline}{%
    \noalign {\ifnum 0=`}\fi \hrule height 1.5pt
    \futurelet \reserved@a \@xhline
}
\newcolumntype{"}{@{\hskip\tabcolsep\vrule width 1pt\hskip\tabcolsep}}
\makeatother

\ShortHeadings{Thompson Sampling Algorithms for Cascading Bandits}{ Zhong, Cheung and Tan}
\firstpageno{1}

\begin{document}

\title{Thompson Sampling Algorithms for Cascading Bandits}


\author{\name Zixin Zhong  \email zixin.zhong@u.nus.edu \\
       \addr Department of Mathematics \\
       National University of Singapore\\
       119076, Singapore
        \AND
        \name Wang Chi Chueng \email isecwc@nus.edu.sg \\
       \addr Department of Industrial Systems and Management\\
       National University of Singapore\\
       117576, Singapore
       \AND
       \name Vincent Y. F. Tan  \email vtan@nus.edu.sg \\
       \addr Department of Electrical and Computer Engineering and \\
       Department of Mathematics \\
       National University of Singapore\\
       117583, Singapore
       }

\editor{NA}       


\maketitle

\begin{abstract}
%
%

 Motivated by the pressing need for efficient optimization in online recommender systems, we revisit the cascading bandit model proposed by Kveton et al.\ (2015). 
 While Thompson sampling (TS) algorithms have been  shown to be empirically superior to Upper Confidence Bound (UCB) algorithms for cascading bandits, theoretical guarantees are only known for the latter.
In this paper, we first provide a problem-dependent
upper bound on the regret of a TS algorithm with Beta-Bernoulli updates; this upper bound is tighter than a recent derivation under a more general setting by \cite{pmlr-v89-huyuk19a}.
Next,
we design and analyze another TS algorithm with Gaussian updates, {\sc TS-Cascade}. {\sc TS-Cascade} achieves the state-of-the-art regret bound for cascading bandits. 
Complementarily,
we consider a linear generalization of the cascading bandit model, which allows efficient learning in large cascading bandit problem instances. We introduce and analyze a TS algorithm, 
which enjoys a regret bound that depends on the dimension of the linear model but not the number of items.
Finally,
by using information-theoretic techniques and judiciously constructing cascading bandit instances,
we derive a nearly matching regret lower bound for the standard model.
Our paper establishes the first theoretical guarantees on TS algorithms for stochastic combinatorial bandit problem model with partial feedback.  
Numerical experiments demonstrate the superiority of the proposed 
TS algorithms compared to existing UCB-based ones.
\end{abstract}


\section{Introduction}\vspace{-.1in} \label{cascadeTS_sec:intro}
Online recommender systems seek to recommend a small list of items (such as movies or hotels) to users based on a larger ground set $[L] := \{1, \ldots, L\}$ of items. Optimizing the performing of these systems is of fundamental importance in the e-service industry, where companies such as Yelp and Spotify strive to maximize users' satisfaction by catering to their taste. The model we consider in this paper is the {\em cascading bandit} model  \citep{kveton2015cascading}, which models online learning in the standard {\em cascade}  model by~\cite{Craswell08}. The latter model  \citep{Craswell08} is widely used in information retrieval and online advertising.

In the cascade model, a recommender offers a list of items to a user. The user then scans through it in a sequential manner. She looks at the first item, and if she is {\em attracted} by it, she {\em clicks} on it. If not, she skips to the next item and clicks on it if she finds it attractive. This process stops when she clicks on one item in the list or when she  comes to the end of the list, in which case she is {\em not attracted} by {\em any} of the items. The items that are in the ground set but not in the chosen list and those in the list that come after the attractive one are {\em unobserved}. Each item $i\in [L]$, which has a certain {\em click probability} $w(i)\in [0, 1]$, attracts the user independently of other items. Under this assumption, the optimal solution is the list of items that maximizes the probability that the user finds an attractive item. This is precisely the list of the most attractive items.

In the cascading bandit model, which is a {\em multi-armed bandit} version of the cascade model by \citep{Craswell08}, the click probabilities $\bm w := \{w(i)\}^L_{i=1}$  are {\em unknown} to the learning agent, and should be learnt over time. Based on the lists  previously chosen and the rewards obtained thus far, the agent tries to learn the click probabilities (exploration) in order to adaptively and judiciously recommend other lists of items (exploitation) to maximize his overall reward over $T$ time steps.

Apart from the cascading bandit model where each item's click probability is learnt individually (we call it the \emph{standard cascading bandit problem}), we also consider a {\em linear generalization of the cascading bandit model}, which is called the  linear cascading bandit problem, proposed by \citep{ZongNSNWK16}. In the linear model, each click probability $w(i)$ is known to be equal to $x(i)^\top \beta\in [0, 1]$, where the feature vector $x(i)$ is {\em known} for each item $i$, 
but the latent $\beta\in \bbR^d$ is {\em unknown} and should be learnt over time. The feature vectors $\{x(i)\}_{i = 1}^L$ represents the prior knowledge of the learning agent. When we have $d < L$, the learning agent can potentially harness the prior knowledge $\{x(i)\}_{i = 1}^L$ to learn $\{w(i)\}^L_{i=1}$ by estimating $\beta$, which is more efficient than by estimating each $w(i)$ individually.

\textbf{Main Contributions.} 
First, we analyze the Thompson sampling algorithm with the Beta-Bernoulli update, which is the application of Combinatorial Thompson Sampling ({\sc CTS}) Algorithm~\citep{pmlr-v89-huyuk19a} in the cascading bandit setting.  We provide a concise analysis which leads to a tighter upper bound compared to the original one reported in \cite{pmlr-v89-huyuk19a}.
 
 
Our next contribution is the design and analysis of {\sc TS-Cascade}, a Thompson sampling algorithm \citep{Thompson33} for the standard cascading bandit problem. 
Our motivation is to design an algorithm that can be generalized and applied to the linear bandit setting, since it is not so natural to design a linear generalization of the {\sc CTS} algorithm.
The linear generalization of {\sc TS-Cascade} is presented in Section~\ref{cascadeTS_sec:lin_alg_cas}.
    Our design involves two novel features. First, the Bayesian estimates on the vector of latent click probabilities $\bm w$ are constructed by a univariate Gaussian distribution. Consequently, in each time step, {\sc Ts-Cascade} conducts exploration in a suitably defined one-dimensional space. 
    Second, inspired by \cite{AudibertMS09}, we judiciously incorporate the empirical variance of each item's click probability in the Bayesian update. The allows efficient exploration on item $i$ when $w(i)$ is close to $0$ or~$1$.
    
We establish a problem-independent regret bound for our proposed algorithm {\sc TS-Cascade}.      Our regret bound matches the state-of-the-art regret bound for UCB algorithms on the standard cascading bandit model  \citep{wang2017improving}, up to a multiplicative logarithmic factor in the number of time steps $T$, when $T \geq L$. 
Our regret bound is the first theoretical guarantee on a Thompson sampling algorithm for the standard cascading bandit problem model, or for any stochastic combinatorial bandit problem model with partial feedback (see literature review). 

Our consideration of Gaussian Thompson sampling is primarily motivated by \cite{ZongNSNWK16}, who report the empirical effectiveness of Gaussian Thompson sampling on cascading bandits, and raise its theoretical analysis as an open question. In this paper, we answer this open question 
under both the standard setting and its linear generalization,
 and overcome numerous analytical challenges. 
We carefully design estimates on the latent mean reward (see inequality~\eqref{cascadeTS_eq:crucial_ineq}) to handle the subtle statistical dependencies between partial monitoring and Thompson sampling.
We reconcile the statistical inconsistency in using Gaussians to model click probabilities by considering a certain truncated version of the Thompson samples (Lemma~\ref{cascadeTS_lemma:gap_bound}).
Our framework provides useful tools for analyzing Thompson sampling on stochastic combinatorial bandits with partial feedback in other settings.

Subsequently, we extend {\sc TS-Cascade} to \algCasLin, a Thompson sampling algorithm for the linear cascading bandit problem and derive a problem-independent upper bound on its regret.
 According to \cite{kveton2015cascading} and our own results, the regret on standard cascading bandits grows linearly with $\sqrt{L}$. Hence, the standard cascading bandit problem is intractable when $L$ is large, which is typical in real-world applications. This motivates us to consider linear cascading bandit problem~\citep{ZongNSNWK16}. 
 \cite{ZongNSNWK16} introduce {\sc CascadeLinTS}, an implementation of Thompson sampling in the linear case, but present no analysis. In this paper, we propose {\algCasLin}, a Thompson sampling algorithm different from {\sc CascadeLinTS}, and upper bound its regret. We are the first to theoretically establish a regret bound for Thompson sampling algorithm on the linear cascading bandit  problem.

Finally, we derive a problem-independent lower bound on the regret incurred by any online algorithm for the standard cascading bandit problem. Note that a problem-dependent one is provided in \cite{kveton2015cascading}.

	\textbf{Literature Review. }Our work is closely related to existing works on the class of stochastic combinatorial bandit (SCB) problems and Thompson sampling. In an SCB model, an arm corresponds to a subset of a ground set of items, each associated with a latent random variable. The corresponding reward depends on the constituent items' realized random variables. SCB models with \emph{semi-bandit feedback}, where a learning agent observes all random variables of the items in a pulled arm, are extensively studied in existing works. Assuming semi-bandit feedback, \cite{AnantharamVW87} study the case when the arms constitute a uniform matroid, \cite{KvetonWAEE14} study the case of general matroids, \cite{GaiKJ10} study the case of permutations, and \cite{GaiKJ12}, \cite{ChenWY13}, \cite{CombesTPL15},  and \cite{KvetonWAS15a} investigate various general SCB problem settings. More general settings with contextual information \citep{LiChuLangford10,QinCZ14} and linear generalization \citep{WenKA15,agrawal2013further} are also studied. When \cite{WenKA15} consider both UCB and TS algorithms, \cite{agrawal2013further} focus on TS algorithms; all other works above hinge on UCBs.

	Motivated by numerous applications in recommender systems and online advertisement, SCB models have been studied under a more challenging setting of \emph{partial feedback}, where a learning agent only observes the random variables for a subset of the items in the pulled arm. A prime example of SCB model with partial feedback is the cascading bandit model, which is first introduced by \cite{kveton2015cascading}. Subsequently, \cite{KvetonWAS15b},  \cite{KatariyaKSW16}, \cite{LagreeVC16} and \cite{ZoghiTGKSW17}  study the cascading bandit model in various general settings. Cascading bandits with contextual information \citep{LiWZC16} and linear generalization \citep{ZongNSNWK16} are also studied. \cite{wang2017improving} provide a general algorithmic framework on SCB models with partial feedback.  All of the works listed above only involve the analysis of UCB algorithms.
	
	On the one hand, UCB has been extensively applied for solving various SCB problems. On the other hand, {\em Thompson sampling} \citep{Thompson33, ChapelleL09, RussoRKOW18}, an online algorithm based on Bayesian updates,  has been shown to be empirically superior compared to UCB and $\epsilon$-greedy algorithms in various bandit models. The empirical success has motivated a series of research works on the theoretical performance guarantees of Thompson sampling on multi-armed bandits \citep{AgrawalG12,KaufmannKM12,agrawal2017near, AgrawalG17}, linear bandits \citep{AgrawalG13b}, generalized linear bandits \citep{AbeilleL17}, etc. Thompson sampling has also been studied for SCB problems with semi-bandit feedback. \cite{KomiyamaHN15} study the case when the combinatorial arms constitute a uniform matroid; \cite{WangC18} investigate the case of general matroids, and  \cite{GopalanMM14}  and  \cite{pmlr-v89-huyuk19a} consider settings with general reward functions. In addition, SCB problems with semi-bandit feedback are also studied in the Bayesian setting \citep{RussoV14}, where the latent model parameters are assumed to be drawn from a known prior distribution. Despite existing works, an analysis of Thompson sampling for an SCB problem in the more challenging case of partial feedback is yet to be done. Our work fills in this gap in the literature, and our analysis provides tools for handling the statistical dependence between Thompson sampling and partial feedback in the cascading bandit models.
    
\textbf{Outline. }In Section~\ref{cascadeTS_sec:prob_setup}, we formally describe the setup of the standard and the linear cascading bandit problems. 
In Section~\ref{cascadeTS_sec:beta_alg_cas}, we provide an upper bound on the regret of {\sc CTS} algorithm and a proof sketch.
In Section~\ref{cascadeTS_sec:gauss_alg_cas}, we propose our first algorithm {\sc TS-Cascade} for the standard problem, and present an asymptotic upper bound on its regret. Additionally, we compare our algorithm to the state-of-the-art algorithms in terms of their regret bounds. 
We also provide an outline to prove a regret bound of {\sc TS-Cascade}.
In section~\ref{cascadeTS_sec:lin_alg_cas}, we design {\algCasLin} for the linear setting, provide an upper bound on its regret and the proof sketch.
In Section~\ref{cascadeTS_sec:lb_cas_reg}, we provide a lower bound on the regret of any algorithm in the cascading bandits with its proof sketch.
In Section~\ref{cascadeTS_sec:exp_cas_reg_min_linear}, we evaluate the {\algCasLin} algorithm with experiments and show the regrets grow as predicted by our theory.
 We discuss future work and conclude in Section~\ref{cascadeTS_sec:summary_cas_reg_min}. Besides, details of proofs 
as given in Appendix~\ref{cascadeTS_sec:proofs}.
    
    
%

\section{Problem Setup}    \label{cascadeTS_sec:prob_setup}  
    Let there be $L\in\mathbb{N}$ ground items, denoted as $[L]:= \{1, \ldots, L\}$. Each item $i\in [L]$ is associated with a weight $w(i)\in [0, 1]$, signifying the item's click probability.  In the standard cascading bandit problem, 
    the click probabilities $w(1), \ldots, w(L)$ are 
    not known to the agent. 
    The agent needs to learn these probabilities and construct an accurate estimate for each of them individually. 
  In the linear generalization setting,
the agent possesses certain linear contextual knowledge about each of the click probability terms $w(1), \ldots, w(L)$.
%
The agent knows that, for each item $i$ it holds that $$w(i) = \beta^\top x(i),$$ where $x(i)\in \mathbb{R}^d$ is the feature vector of item $i$, and $\beta\in \mathbb{R}^d$ is a common vector shared by all items. While the agent knows $x(i)$ for each $i$, he does not know $\beta$. And the problem now reduces to estimating $\beta\in \bbR^d$. 
The linear generalization setting represents a setting when the agent could harness certain prior knowledge (through the features $x(1), \ldots, x(L)$ and their linear relationship to the respective click probability terms) to learn the latent click probabilities efficiently. Indeed, in the linear generalization setting, the agent could jointly estimate $w(1), \ldots, w(L)$ by estimating $\beta$, which could allow more efficient learning than estimating $w(1), \ldots, w(L)$ individually when $d < L$. 
This linear parameterization is a generalization of the standard case in the sense that when $\{ x(i) \}_{i=1}^L$ is the set of $(d=L)$-dimensional standard basis vectors, $\beta$ is  a length-$L$ vector filled with click probabilities, i.e., $\beta= [ w(1), w(2),\ldots, w(L) ]^T$, then the problem reverts to the standard case. This standard basis case actually represents the case when the $x(i)$ does not carry useful information.

    At each time step $t \in [T]$, the agent recommends a list of $K\le L$ items $S_t: = (i^t_1, \ldots, i^t_K)\in [L]^{(K)} $ to the user, where $[L]^{(K)}$ denotes the set of all $K$-permutations of $[L]$. The user examines the items from $i_1^t$ to $i_K^t$ by  examining each item one at a time, until one item is clicked or all items are examined. For $1\leq k\leq K$,  $W_t(i^t_k) \sim \text{Bern}\left(w(i^t_k)\right)$ are i.i.d.\ and $W_t(i^t_k)=1$ iff   user clicks on $i^t_k$ at time~$t$. 
    
    The {\em instantaneous reward} of the agent at time $t$ is 
    $$R(S_t|{\bm w}): = 1 - \prod^K_{k=1}(1 - W_t(i^t_k)) \in\{0,1\}.$$
    In other words, the agent gets a reward of $R(S_t|{\bm w}) = 1$ if $W_t(i^t_k) = 1$ for some $1\leq k\leq K$, and a reward of $R(S_t|{\bm w}) = 0$ if  $W_t(i^t_k) = 0$ for all $1\leq k\leq K$.
    
    The {\em feedback} from the user at time $t$ is defined as 
    $$k_t := \min\{1\leq k\leq K: W_t(i^t_k) = 1 \},$$
    where we assume that the minimum over an empty set is $\infty$.  If $k_t < \infty$, then the agent observes $W_t(i^t_k) = 0$ for $1\leq k < k_t$, and also observes $W_t(i^t_k) = 1$, but does not observe $W_t(i^t_k)$ for $k> k_t$; otherwise, $k_t = \infty$, then the agent observes $W_t(i^t_k) = 0$ for $1\leq k \leq K$.
    
    As the agent aims to maximize the sum of rewards over all steps, an expected cumulative regret is defined to evaluate the performance of an algorithm. First, the {\em expected instantaneous reward} is
    $$r(S |{\bm w}) = \mathbb{E}[R(S |{\bm w}) ]        
    = 1 - \prod_{i \in S}(1 - w(i )).$$
    Note that the expected reward is permutation invariant, but the set of observed items is not.  Without loss of generality, we assume that $w(1)\geq w(2)\geq \ldots \geq w(L)$. Consequently, any permutation of $\{1, \ldots, K\}$ maximizes the expected reward. We let $S^* = (1, \ldots , K )$ be the optimal ordered $K$-subset that maximizes the expected reward. Additionally, we let items in $S^*$ be optimal items and others be suboptimal items. In $T$ steps, we aim to minimize the {\em expected cumulative regret}:
    \begin{equation*}
        \mathrm{Reg}(T) := T\cdot r(S^* |{\bm w}) - \sum^T_{t=1}r(S_t | {\bm w}),
    \end{equation*}
while the vector of click probabilities ${\bm w}\in [0, 1]^L$ is not known to the agent, and the recommendation list $S_t$ is chosen online, i.e., dependent on the previous choices and previous rewards.

\section{The {\sc CTS} Algorithm} \label{cascadeTS_sec:beta_alg_cas}

In this section, we analyze a Thompson sampling algorithm for the cascading bandit problem and consider the posterior sampling based on the Beta-Bernoulli conjugate pair. 
The algorithm, dubbed {\sc CTS} algorithm, was first analyzed by \cite{pmlr-v89-huyuk19a}. We derive a tighter bound on the regret than theirs. The algorithm and the bound are as follows.


%

    \begin{algorithm}[H]
		\caption{{\sc CTS}, Thompson Sampling for Cascading Bandits with Beta Update}\label{cascadeTS_alg:ts_beta}
		\begin{algorithmic}[1]
			\State Initialize $\alphaAlgBeta_1(i)=1$ and $\betaAlgBeta_1(i)=1$ for all $i\in [L]$.
			\For{$t = 1, 2, \ldots$}
			    \State Generate Thompson sample $\thetaAlgBeta_t (i)\sim \mathrm{Beta}( \alphaAlgBeta_t(i), \betaAlgBeta_t(i) )$ for all $i\in [L]$. \label{cascadeTS_alg:sample_beta}
				\For{$k\in [K]$}
				    \State Extract $i^t_k\in \text{argmax}_{i\in [L]\setminus \{i^t_1, \ldots, i^t_{k-1}\}} \thetaAlgBeta_t(i)$.
				\EndFor
				\State Pull arm $S_t = (i^t_1, i^t_2, \ldots, i^t_K)$. \label{cascadeTS_alg:ts_opt}
				\State Observe click $k_t \in \{ 1, \ldots,K, \infty \}$.
				\State For each $i\in [L]$, if $W_t(i)$ is observed, define 
				$\alphaAlgBeta_{t+1}(i) = \alphaAlgBeta_{t}(i) + W_t(i)$, $\betaAlgBeta_{t+1}(i) = \betaAlgBeta_{t}(i) + 1 - W_t(i)$.
				Otherwise, $\alphaAlgBeta_{t+1}(i) = \alphaAlgBeta_{t}(i) $, $\betaAlgBeta_{t+1}(i) = \betaAlgBeta_{t}(i) $.
			\EndFor
		\end{algorithmic}
	\end{algorithm}


\begin{theorem}[Problem-dependent bound for {\sc CTS}] \label{cascadeTS_thm:upbd_dep_beta}
	Assume $ w (1 ) \ge \ldots \ge  w (K ) >  w (K+1)  \ge \ldots \ge w (L) $ and  $\Delta_{i,j} = w(i) - w(j) $ for $i \in S^*$, $j \notin S^*$,
	the expected cumulative regret of {\sc CTS} is bounded as:\\
\resizebox{  \textwidth}{!}{$
    \displaystyle
     \mathrm{Reg} (T)
    \le L \!  -\!  K \!  + \!  16 \cdot \! \! \! \! \!  \sum_{i=K+1}^L \!  \bigg( \frac{ 1  }{ \Delta_{K,i}^2 }
        +  \frac{   \log T  }{  \Delta_{K,i} } \bigg) \! + \! \! \sum_{i=K+1}^L  \sum_{j=1}^K \Delta_{j,i}
        \! +   \alpha_2 \cdot \!  \sum_{j=1}^K \Delta_{j,L}  \bigg( \frac{ \mathsf{1} \big[ \Delta_{j,K+1} \! < \! \frac{1}{4}  \big]  }{  \Delta_{j,K+1}^3 } + \frac{ \log T }{ \Delta_{j,K+1}^2  }  \bigg),
$}
\\
where $a_2 $
 is a constant
\footnote{ 
It is common to involve a problem-independent constant in the upper bound \citep{WangC18}.
}
independent of the problem instance.
\end{theorem}
	\begin{table*}[ht]
	  \centering
	  \caption{Problem-dependent upper bounds on the $T$-regret of  {\sc CTS},  {\sc CascadeUCB1} and {\sc CascadeKL-UCB}; the problem-dependent lower bound apply to  {\em all} algorithms for the cascading bandit model. 
	    }	  
	  \vspace{1em}
	    \begin{tabular}{lll}
		    \textbf{Algorithm/Setting}     & \textbf{Reference} &  \textbf{Bounds}    \\
		    \hline
		    {\sc CTS}  & Theorem~\ref{cascadeTS_thm:upbd_dep_beta} & $ O \big(  L\log T /\Delta +  L / \Delta^2 \big) \vphantom{\Big( } $    \\
		    {\sc CTS} &\cite{pmlr-v89-huyuk19a} & $O\big( KL\log T/[(1-w(L))^{K-1}w(L)\Delta]  \big)  \vphantom{\Big( } $  \\
		    {\sc CascadeUCB1} &\cite{kveton2015cascading}  & $O\big( {(L-K)(\log T)}/{\Delta} + L \big)  \vphantom{\Big( } $   \\
		    {\sc CascadeKL-UCB} & \cite{kveton2015cascading}  & $O\big( {  (L-K)(\log T)\log (1/\Delta)}/{ \Delta }  \big)  \vphantom{\Big( } $  \\
		    \hline
		    Cascading Bandits &\cite{kveton2015cascading}& $\Omega\big( { (L-K)}( \log T)/{ \Delta } \big)  \vphantom{\Big( }  $ (Lower Bd)  
	    \end{tabular}%
	  \label{cascadeTS_tab:bound_dep}%
	\end{table*}

	Next, in Table~\ref{cascadeTS_tab:bound_dep}, we compare our regret bound for the {\sc CTS} algorithm to those in the existing literature. 
	Note that the upper bound for {\sc CTS} in \cite{pmlr-v89-huyuk19a} can be specified to the cascading bandits as above, and the two other algorithms are based on the UCB idea  \citep{kveton2015cascading}.  
		To present the results succinctly, 
	we consider the problem where $\Delta:=\Delta_{j,i}  $. In other words, the gap between any optimal item $j$ and any suboptimal item $i$ is $\Delta$, which does not depend on $i$ and $j$. 
%
	The parameter $\Delta$ measures the difficulty of the problem.
 
	Table~\ref{cascadeTS_tab:bound_dep} implies that our upper bound grows like $\log T$, similarly to the other bounds for the other algorithms. 
When $\log T \ge 1/\Delta$,
since $L \log T/\Delta$ 
is the dominant term in our bound,	
	we can see that it is smaller than the term $KL\log T/[(1-w(L))^{K-1}w(L)\Delta]$  in the bound by~\cite{pmlr-v89-huyuk19a}. 
	Besides, the former matches the bound for {\sc CascadeUCB1} in the term involving $\log T$ and is slightly better than the one for  {\sc CascadeKL-UCB}~\citep{kveton2015cascading}. 
	Furthermore, our bound matches the lower bound provided in~\cite{kveton2015cascading}. 
	Therefore, our derived bound on {\sc CTS}, the Thompson sampling algorithm with Beta-Bernoulli update, is tight.

Moreover, apart from the problem-dependent bound, we also notice that it is valuable to derive a problem-independent bound for CTS. However,  in the cascading bandit model, it cannot be trivially reduced from a problem-dependent one, and hence we leave it as an open question.
For instance, \cite{kveton2015cascading} (the first paper on cascading bandits) provided an problem-dependent regret bound, but left the problem-independent bound as an open question.

    
%


To derive an upper bound, the analytical challenge is to carefully handle the partial feedback from the user in the cascading bandits. Though our analysis 
has similarity to those in~\cite{WangC18} (semi-bandit feedback) and~\cite{pmlr-v89-huyuk19a} (partial feedback), the difference is manifested in the following aspects: 
\begin{itemize}
    \item[(i)] To handle the partial feedback from the user, we handcraft the design of events $\calA_{t}(\ell, j, k)$,   $\tilde{\calA}_{t}( \ell, j, k)$, $\calA_t( \ell, j)$ in \eqref{cascadeTS_eq:beta_event_def}, and apply the property of the reward to decompose the regret in terms of the observation of a suboptimal items in Lemma~\ref{cascadeTS_lemma:beta_1st_term_part} and \ref{cascadeTS_lemma:beta_3rd_term_part}.
    \item[(ii)] Additionally, our algorithm crucially requires the items in $S_t$ to be in the 
 descending order of $\thetaAlgBeta_t(i)$ (see Line 4--7 of Algorithm~\ref{cascadeTS_alg:ts_beta}), which is an important algorithmic property that enables our analysis (see Lemma~\ref{cascadeTS_lemma:beta_3rd_term_part}).%
\end{itemize}
Our analytical framework, which judiciously takes into the impact of partial feedback, is substantially different from \cite{WangC18}, which assume semi-feedback and do not face the challenge of partial feedback.
Besides, since \cite{pmlr-v89-huyuk19a} consider a less specific problem setting, they define more complicated events than in \eqref{cascadeTS_eq:beta_event_def} to decompose the regret. Moreover, we carefully apply the mechanism of {\sc CTS} algorithm to derive Lemma~\ref{cascadeTS_lemma:beta_3rd_term_part}, which leads to a tight bound on the regret. 
Overall, our concise analysis fully utilizes the properties of cascading bandits and Algorithm~\ref{cascadeTS_alg:ts_beta}.


    Now we outline the proof of Theorem~\ref{cascadeTS_thm:upbd_dep_beta}; some details are deferred to Appendix~\ref{cascadeTS_sec:proofs}. 
     Note that a pull of any suboptimal item contributes to the loss of expected instantaneous reward. 
    Meanwhile, we choose $S_t$ according to the Thompson samples $\thetaAlgBeta_t(i)$ instead of the unknown click probability $w(i)$.
    We expect that as time progresses, the empirical mean $\hat{w}_t(i)$ will approach $w(i)$ and the Thompson sample $\thetaAlgBeta_t(i)$ which is drawn from $\mathrm{Bern}( \thetaAlgBeta_t(i) )$ will concentrate around $w(i)$ with high probability.
    In addition, this probability depends heavily on the number of observations of the item.
    Therefore, the number of observations of each item affects the distribution of the Thompson samples and hence, also affects the instantaneous rewards. Inspired by this observation, we decompose the the regret in terms of the number of observations of the suboptimal items in the chain of equations including \eqref{cascadeTS_eq:beta_decomp_reg_by_N_ij}  to follow. 
     
      
    We first define $\pi_t$ as follows. For any $k$, if the $k$-th item in $S_t$ is optimal, we place this item at position $k$, $\pi_t(k) = i^t_k$. The remaining optimal items are positioned arbitrarily. Since $S^*$ is optimal with respect to $w$, $w(i^t_k) \le w(\pi_t(k))$ for all $k$. Similarly, since $S_t$ is optimal with respect to $\thetaAlgBeta_t$, $\thetaAlgBeta_t(i^t_k) \ge \thetaAlgBeta_t( \pi_t(k))$ for all $k$. 

   Next, for any optimal item $j$ and suboptimal item $\ell$, index $1\le k \le K$, time step $t$, we say $ {\calA}_{t}(\ell, j, k)$ holds when suboptimal item $\ell$ is the $k$-th item to pull, optimal item $j$ is not to be pulled,  and $\pi_t$ places $j$ at position $k$.
   We say $\tilde{\calA}_{t}(\ell, j, k)$ holds when $ {\calA}_{t}( \ell, j, k)$ holds and item $ \ell$ is observed.
   Additionally, $\calA_t( \ell, j)$ is the union set of events $\tilde{\calA}_{t}( \ell, j, k)$ for $k=1,\ldots,K$. We can also write these events as follows:
    \begin{align}
      \calA_{t}( \ell, j, k) &= \left\{  i_{k}^{t}= \ell, \pi_t(k) = j \right\},   \nonumber \\
            \tilde{\calA}_{t}( \ell, j, k) &= \left\{  i_{k}^{t}= \ell, \pi_t(k) = j,  W_{t}( i_{1}^{t} ) = \ldots = W_{t}( i_{k-1}^{t} ) =0 \right\}  \nonumber \\
                & = \left\{  i_{k}^{t}=\ell, \pi_t(k) = j,  \text{Observe $W_t( \ell )$ at $t$}  \right\},  \nonumber \\
      \calA_t(\ell, j) &= \bigcup_{k=1}^K  \tilde{\calA}_{t} (\ell, j, k) =  \left\{ \exists 1 \leq k \leq K ~s.t.~ i_{k}^{t}=\ell, \pi_t(k) = j, \text{Observe $W_t(\ell)$ at $t$}  \right\}.\label{cascadeTS_eq:beta_event_def}
    \end{align}
Then we have
   \begin{align}
        & r(S^* |{\bm w}) - r(S_t |{\bm w})  \nonumber  \\
        & = r( \pi_t(1),\ldots,\pi_t(K)  |{\bm w}) - r(S_t |{\bm w})  \label{cascadeTS_eq:beta_cas_reg_decomp_one}  \\
        & = \sum^K_{k=1}\left[\prod^{k-1}_{m=1} [1 - w(i_m^t)) \right] \cdot [ w(  \pi_t(k) ) - w(i_k^t) ] \cdot \left[\prod^{K}_{n=k+1}(1 - w( \pi_t(n) ) ]  \right] \label{cascadeTS_eq:beta_cas_reg_decomp_two} \\
        & \le  \sum^K_{k=1}\left[\prod^{k-1}_{m=1}(1 - w(i_m^t)) \right] \cdot [ w(  \pi_t(k) ) - w(i_k^t) ]    \nonumber \\
        & \le \sum^K_{k=1}  \sum_{j=1}^{K} \sum_{ \ell =K+1}^{L}  \mathbb{E} \left[   \Big[\prod^{k-1}_{m=1}(1 - w(i_m^t)) \Big] \cdot [ w(  \pi_t(k) ) - w(i_k^t) ] ~| ~ {\calA}_t(  \ell,j,k)   \right] \cdot \mathbb{E} \left[\mathsf{1}\left\{ {\calA}_t( \ell,j,k)  \right\}\right]    \nonumber \\
        &  =  \sum_{j=1}^{K} \sum_{ \ell =K+1}^{L} \sum_{k=1}^K \Delta_{j, \ell}   
            \cdot  \mathbb{E} \left[  ~\mathsf{1}\left\{ W_{t}( i_{1}^{t} ) = \ldots = W_{t}( i_{k-1}^{t} ) =0  \right\}  |  {\calA}_t( \ell,j,k)   \right]
            \cdot \mathbb{E} \left[\mathsf{1}\left\{   {\calA}_t( \ell, j,k)  \right\}\right]    \nonumber \\
        & = \sum_{j=1}^{K} \sum_{ \ell=K+1}^{L} \Delta_{ j, \ell }  \cdot  \sum_{k=1}^K  \mathbb{E} \left[\mathsf{1}\left\{ \tilde{\calA}_t( \ell ,j,k) \right\}\right]   \nonumber  \\
        & = \sum_{j=1}^{K} \sum_{ \ell=K+1}^{L} \Delta_{j,  \ell }  \cdot \mathbb{E} \left[\mathsf{1}\left\{ \calA_t(\ell, j) \right\}\right] .    \nonumber 
        %
   \end{align}
Line~\eqref{cascadeTS_eq:beta_cas_reg_decomp_one} results from the fact that any permutation of $\{1, \ldots, K\}$ maximizes the expected reward and $( \pi_t(1), \ldots, \pi_t(K) )$ is one permutation of $\{1, \ldots, K\}$. Line~\eqref{cascadeTS_eq:beta_cas_reg_decomp_two} applies the property of the reward (Lemma~\ref{cascadeTS_lemma:zong_sysmmetry_reg_decomp}).
The subsequent ones follow from the various events defined in \eqref{cascadeTS_eq:beta_event_def}.
   
Let $\numAlgBeta(i,j)$ denote the number of time steps that $\calA_t(i, j)$ happens for optimal item $j$ and suboptimal item $i$ within the whole horizon, then
\begin{align}
    &
    \numAlgBeta(i,j) = \sum_{t=1}^T  \mathbb{E} \left[\mathsf{1}\left\{ \calA_t(i, j) \right\}\right],\quad
  \nonumber \\
  & \text{Reg}(T)  
  =   T\cdot  r(S^* |{\bm w}) -  \sum_{t=1}^T r(S_t |{\bm w}) 
  \le \sum_{j=1}^{K} \sum_{ i=K+1}^{L} \Delta_{j, i} \cdot  \bbE [ \numAlgBeta(i,j) ].
      \label{cascadeTS_eq:beta_decomp_reg_by_N_ij}
\end{align}  
To bound the expectation of $\numAlgBeta(i,j)$, we define two more events taking into account the concentration of $\thetaAlgBeta_{t}(i)$ and $\hat{w}_{t}(i)$:
\begin{align*}
    & \mathcal{B}_{t}(i)=\left\{\hat{w}_{t}(i)> w(i)+\varepsilon_i \right\} , \quad   
     \mathcal{C}_{t}(i,j)=\left\{\thetaAlgBeta_{t}(i)>  w(j) -\varepsilon_j \right\} ,\quad \ 1\le j \le K < i \le L.
\end{align*}    
with
\begin{align*}
    \varepsilon_i = \frac{1}{2} \cdot \Big(  \Delta_{K+1,i} +  \frac{ \Delta_{K,K+1} }{2}  \Big),
    \quad
    \varepsilon_j = \frac{1}{2} \cdot \Big(  \Delta_{j,K} +  \frac{ \Delta_{K,K+1} }{2}  \Big),
    \quad \ 1\le j \le K < i \le L.
\end{align*}
Then $ \varepsilon_i \ge \Delta_{K,i}/4$, $\varepsilon_j \ge \Delta_{j,K+1}/4$, $\varepsilon_i+ \varepsilon_j = \Delta_{j,i}/2 $,
and we have
\begin{align} 
    & \mathbb{E}\left[\numAlgBeta (i, j) \right]
    =   \mathbb{E}\left[\sum_{t=1}^{T} \mathsf{1}\left[\calA_t(i, j)\right]\right] \nonumber \\
    &  \leq \mathbb{E}\left[\sum_{t=1}^{T} \mathsf{1}\left[\mathcal{A}_t(i, j) \wedge \mathcal{B}_{t}(i)\right]\right]
     +\mathbb{E}\left[\sum_{t=1}^{T} \mathsf{1}\left[\calA_t(i, j) \wedge \neg \mathcal{B}_{t}(i) \wedge \mathcal{C}_{t}(i,j)\right]\right] 
       +\mathbb{E}\left[\sum_{t=1}^{T} \mathsf{1}\left[\calA_t(i, j) \wedge \neg \mathcal{C}_{t}(i,j)\right]\right] . \label{cascadeTS_eq:beta_decomp_N_ij}
\end{align}

Subsequently, we bound the three terms in~\eqref{cascadeTS_eq:beta_decomp_N_ij} separately.
Since we need to count the number of observations of items in the subsequent analysis, we also use $N_t(i)$ to denote one plus the number of observations of item $i$ before time $t$ (consistent with other sections).


{\bf The first term.}
We first derive a lemma which facilitates us to bound the probability the event $\calA_t(i, j) \wedge \mathcal{B}_{t}(i)$.

\begin{restatable}
{lemma}{cascadeTSlemmaBetaFirstTermPart}
\label{cascadeTS_lemma:beta_1st_term_part}
    For any suboptimal item $K <  i \leq L$, 
$$
\mathbb{E}\left[~ \exists t : 1 \leq t \leq T, i \in S_t,| \hat{w}_t(i)- w(i) | >\varepsilon_i  , \text{Observe $W_t(i)$ at $t$} ~\right] \leq 1+\frac{1}{\varepsilon_i^{2}}.
$$

\end{restatable}

Next, summing over all possible pairs $(i,j)$ and then applying Lemma~\ref{cascadeTS_lemma:beta_1st_term_part}, the regret incurred by the first term can be bounded as follows:
\begin{align*}
    & \sum_{j=1}^{K} \sum_{ i=K+1}^{L} \mathbb{E}\left[\sum_{t=1}^{T} \mathsf{1}\left[\calA_t(i, j) \wedge \mathcal{B}_{t}(i)\right]\right] \\
    & = \sum_{i=K+1}^L \mathbb{E}\left[\sum_{t=1}^{T} \sum_{j=1}^{K} \mathsf{1}\left[\calA_t(i, j) \wedge \mathcal{B}_{t}(i)\right]\right] \\
    & \le \sum_{i=K+1}^L \mathbb{E}\left[\sum_{t=1}^{T} \mathsf{1}\left[  \mathcal{B}_{t}(i) , \text{Observe $W_t(i)$ at $t$}  \right]\right] \\
    &  \le \sum_{i=K+1}^L  \mathbb{P}\left[~ \exists t : 1 \leq t \leq T, i \in S(t),| \hat{w}_t(i)-w(i)| >\varepsilon  , \text{Observe $W_t(i)$ at $t$} ~\right] \\
    & \le L - K + \sum_{i=K+1}^L  \frac{ 1 }{ \varepsilon_i^2 }
        \le L - K + \sum_{i=K+1}^L  \frac{ 16 }{ \Delta_{K,i}^2 }.
\end{align*}

{\bf The second term.} 
%
%
Considering the occurrence of  $\calA_t(i, j) \wedge \neg \mathcal{B}_{t}(i) \wedge \mathcal{C}_{t}(i,j)$,
we bound the regret incurred by the second term as follows (see Appendix~\ref{cascadeTS_pf:eq_beta_2nd_term} for details): for any  $0<\varepsilon < \Delta_{K,K+1}/2$,
\begin{align}
    & \sum_{i=K+1}^L \sum_{j=1}^K \Delta_{j,i} \cdot \mathbb{E}\left[\sum_{t=1}^{T} \mathsf{1}\left[\calA_t(i, j) \wedge \neg \mathcal{B}_{t}(i) \wedge \mathcal{C}_{t}(i,j) \right]\right]  
     \le  \sum_{i=K+1}^L  \frac{  16\log T  }{  \Delta_{K,i} } + \sum_{i=K+1}^L  \sum_{j=1}^K \Delta_{j,i}.\label{cascadeTS_eq:beta_2nd_term}
\end{align}

%
%

{\bf The third term.}
We use $\thetaAlgBeta(-i)$ to denote the vector $\bmThetaAlgBeta$ without the $i$-th element $\thetaAlgBeta(i)$. For any optimal item $j$, let $W_{i,j}$ be the set of all possible values of
$\bmThetaAlgBeta$ satisfying that $\bar{\mathcal{A}}_{i, j}(t) \wedge \neg \mathcal{C}_{t}(i,j)$ happens for some $i$, and $W_{i,-j}=\left\{\thetaAlgBeta(-j) : {\bmThetaAlgBeta} \in W_{i,j}\right\}$.
We now state Lemma~\ref{cascadeTS_lemma:beta_3rd_term_part} whose proof depends on a delicate design of Algorithm~\ref{cascadeTS_alg:ts_beta}: to pull items in their descending order of Thompson samples $\thetaAlgBeta_t(i)$.

\begin{restatable}
{lemma}
{lemmaBetaThirdTermPart}
\label{cascadeTS_lemma:beta_3rd_term_part}
    For any optimal item $1\le j \le K$, 
    we have
$$
\mathbb{E}\left[\sum_{t=1}^{T} \operatorname{Pr}\left[ ~\exists i~s.t.~ \bmThetaAlgBeta_t \in W_{i,j}, \text{Observe $W_t(i)$ at $t$}  ~\right]\right] \leq
c \cdot \bigg( \frac{ \mathsf{1} [ \varepsilon_j < 1/16  ]    }{ \varepsilon_j^3 } + \frac{ \log T }{ \varepsilon_j^2  }  \bigg),
$$
where $c $ is a constant which does not depend on the problem instance.
\end{restatable}

Recalling the definition of $W_{i,j}$, we bound the regret incurred by the third term 
\begin{align*} 
    & \sum_{i=K+1}^L \sum_{j=1}^K \Delta_{j,i} \cdot \mathbb{E}\left[\sum_{t=1}^{T} \mathsf{1}\left[\calA_t(i, j) \wedge \neg \mathcal{C}_{t}(i,j)\right]\right]  \\
    &\le \sum_{j=1}^K \Delta_{j,L} \cdot \mathbb{E}\left[\sum_{t=1}^{T} \sum_{i=K+1}^L \mathsf{1}\left[\calA_t(i, j) \wedge \neg \mathcal{C}_{t}(i,j)\right]\right] \\ 
    &= \sum_{j=1}^K\Delta_{j,L} \cdot \mathbb{E}\left[\sum_{t=1}^{T} \sum_{i=K+1}^L \mathsf{1}\left[ [ \bmThetaAlgBeta_t \in W_{i,j}, \text{Observe $W_t(i)$ at $t$}  ~ \right]\right] \\ 
    & = \sum_{j=1}^K \Delta_{j,L} \cdot \mathbb{E}\left[\sum_{t=1}^{T} \mathsf{1}\left[    ~\exists i~s.t.~  \bmThetaAlgBeta_t \in W_{i,j}, \text{Observe $W_t(i)$ at $t$}  ~ \right]\right] \\ 
    & \leq c \cdot \sum_{j=1}^K \Delta_{j,L} \cdot \bigg( \frac{ \mathsf{1} [ \varepsilon_j < 1/16  ]   }{ \varepsilon_j^3 } + \frac{ \log T }{ \varepsilon_j^2  }  \bigg)
    \leq \alpha_2 \cdot \sum_{j=1}^K \Delta_{j,L} \cdot \bigg( \frac{ \mathsf{1} [ \Delta_{j,K+1} < 1/4  ]  }{  \Delta_{j,K+1}^3 } + \frac{ \log T }{ \Delta_{j,K+1}^2  }  \bigg)   
\end{align*}
with $\alpha_2  = 8c $ (a constant which does not depend on the problem instance).

{\bf Summation of all the terms.}
Overall, we derive an upper bound on the total regret:
%
\begin{align*}
    & \mathrm{Reg} (T)
    \\ &
    \le L - K + \sum_{i=K+1}^L  \frac{ 16 }{ \Delta_{K,i}^2 }
        + \sum_{i=K+1}^L  \frac{  16\log T  }{  \Delta_{K,i} } + \sum_{i=K+1}^L  \sum_{j=1}^K \Delta_{j,i}
        + \alpha_2' \cdot \sum_{j=1}^K \Delta_{j,L}  \bigg( \frac{ \mathsf{1} \big[ \Delta_{j,K+1} \! < \! \frac{1}{4}  \big]  }{  \Delta_{j,K+1}^3 } + \frac{ \log T }{ \Delta_{j,K+1}^2  }  \bigg)
    \\ &
    \le L - K + 16 \cdot \! \! \sum_{i=K+1}^L \!  \bigg( \frac{ 1  }{ \Delta_{K,i}^2 }
        +  \frac{   \log T  }{  \Delta_{K,i} } \bigg) + \!  \sum_{i=K+1}^L  \! \sum_{j=1}^K \Delta_{j,i}
        + \alpha_2' \cdot \sum_{j=1}^K \Delta_{j,L}  \bigg( \frac{ \mathsf{1} \big[ \Delta_{j,K+1} \! < \! \frac{1}{4}  \big]  }{  \Delta_{j,K+1}^3 } + \frac{ \log T }{ \Delta_{j,K+1}^2  }  \bigg).
\end{align*}

\section{The {\sc TS-Cascade} Algorithm} \label{cascadeTS_sec:gauss_alg_cas}


The {\sc CTS} algorithm cannot be applied directly to the linear generalization setting, since the Beta-Bernoullli update is incompatible with the linear generalization model on the click probabilities
(see detailed discussions after Lemma~\ref{cascadeTS_lemma:transfertwo} for an explanation).
We propose another Thompson sampling algorithm, {\sc TS-Cascade}. Instead of the Beta distribution, 
{\sc TS-Cascade} designs appropriate Thompson samples that are Gaussian distributed and uses these in the posterior sampling steps.
The linear generalization of {\sc TS-Cascade} is presented in Section~\ref{cascadeTS_sec:lin_alg_cas}.


    Our algorithm {\sc TS-Cascade} is presented in Algorithm~\ref{cascadeTS_alg:ts}. Intuitively, to minimize the expected cumulative regret, the agent aims to learn the true weight $w(i)$ of each item $i\in [L]$ by exploring the space to identify $S^*$ (i.e., exploitation) after a hopefully small  number of steps. In our algorithm, we approximate the true weight $w(i)$ of each item $i$ 
with
 a Bayesian statistic $\theta_t(i)$ at each time step $t$. This statistic is known as the {\em Thompson sample}. To do so, first, we sample a one-dimensional standard Gaussian $Z_t\sim {\cal N}(0, 1)$, define the empirical variance $ \hat{\nu}_t(i)  =\hat{w}_t(i) (1 - \hat{w}_t(i))$ 
of each item,
 and calculate $\theta_t(i)$. 
  Secondly, we select $S_t = (i^t_1, i^t_2, \ldots, i^t_K)$ such that 
     $\theta_t(i_1^t) \ge \theta_t(i_2^t)  \ge \ldots \ge \theta_t(i_K^t)  \ge \max_{j \not\in S_t} \theta_t(j)$; this is reflected in Line 10 of Algorithm~\ref{cascadeTS_alg:ts}. 
    Finally,   we update the parameters for each observed item $i$ in a standard manner  by applying Bayes rule on the mean of the Gaussian (with conjugate prior being another Gaussian) in Line 14.
%
   The algorithm results in the following theoretical guarantee.

       \begin{algorithm}[t]
		\caption{{\sc TS-Cascade}, Thompson Sampling for Cascading Bandits with Gaussian Update}\label{cascadeTS_alg:ts}
		\begin{algorithmic}[1]
			\State Initialize $\hat{w}_1(i) = 0$,  $N_1(i) = 0$ for all $i\in [L]$.
			\For{$t = 1, 2, \ldots$}
			    \State Sample a $1$-dimensional random variable $Z_t\sim {\cal N}(0, 1)$. \label{cascadeTS_alg:ts_gaussian}
				\For{$i\in [L]$}		    
                    \State Calculate   the  empirical variance  
$\hat{\nu}_t(i) =\hat{w}_t(i) (1 - \hat{w}_t(i)).$
                            \State Calculate standard  deviation of the Thompson sample 
                           $
                            \sigma_t(i)   = \max\left\{\sqrt{\frac{\hat{\nu}_t(i)\log (t+1)}{N_t(i) + 1}}, \frac{\log (t+1)}{N_t(i) + 1}\right\}.  
                        $
                    \State Construct the Thompson sample 
                        $
                             \theta_t(i) = \hat{w}_t(i) + Z_t  \sigma_t(i).
                       $
				\EndFor
				\For{$k\in [K]$}
				    \State Extract $i^t_k\in \text{argmax}_{i\in [L]\setminus \{i^t_1, \ldots, i^t_{k-1}\}} \theta_t(i)$.
				\EndFor
				\State Pull arm $S_t = (i^t_1, i^t_2, \ldots, i^t_K)$. \label{cascadeTS_alg:ts_opt}
				\State Observe click $k_t \in \{ 1, \ldots,K, \infty \}$.
				\State For each $i\in [L]$, if $W_t(i)$ is observed, define 
				$\hat{w}_{t+1}(i) = \frac{N_t(i)\hat{w}_t(i) + W_t(i)}{N_t(i) + 1}$ and 
				$N_{t+1}(i) = N_t(i)  +1.$
				Otherwise (if $W_t(i)$ is not observed), $\hat{w}_{t+1}(i) = \hat{w}_t(i), N_{t+1}(i) = N_t(i)$.
			\EndFor
		\end{algorithmic}
	\end{algorithm}

        \begin{theorem}[Problem-independent]\label{cascadeTS_thm:main}
        Consider the cascading bandit problem. Algorithm {\sc TS-Cascade}, presented in Algorithm~\ref{cascadeTS_alg:ts}, incurs an expected regret at most
        $$ O(\sqrt{KLT}\log T + L \log^{5/2}T),$$
        where the big $O$ notation hides a constant factor that is independent of $K, L, T, \bm w$.
    \end{theorem}

	\begin{table*}[ht]
	  \centering
	  \caption{Problem-independent upper bounds on the $T$-regret of  {\sc TS-Cascade} and {\sc CUCB}; the problem-independent lower bound apply to {\em all} algorithms for the cascading bandit model. }
	  \vspace{1em}
	    \begin{tabular}{lll}
		    \textbf{Algorithm/Setting}     & \textbf{Reference} &  \textbf{Bounds}    \\
		    \hline
		    {\sc TS-Cascade} & Theorem~\ref{cascadeTS_thm:main} & $ O(\sqrt{KLT}\log T + L \log^{5/2}T)  \vphantom{\Big( } $ \\
		    {\sc CUCB} & \cite{wang2017improving}  & $O\big( \sqrt{KLT\log T  \vphantom{\bar{d} }  }  ~\big)  \vphantom{\Big(  } $  \\
		    \hline
		    		    Cascading Bandits & Theorem~\ref{cascadeTS_thm:main_lb} & $\Omega\big(\sqrt{LT/K}\big) \vphantom{\Big( }  $ (Lower Bd) 
	    \end{tabular}%
	  \label{cascadeTS_tab:bound_indep}%
	\end{table*}
    \setcitestyle{authoryear,round}	

    In practical applications, $T\gg L$ and so the regret bound is essentially $\tilde{O}(\sqrt{KLT}) $. 
    Besides, we note that in problem-independent bounds for TS-based algorithms, the implied constants in the $O(\cdot)$-notation are usually large (see ~\cite{agrawal2013further, agrawal2017near}). These large constants result from the anti-concentration bounds used in the analysis.
    In this work, we mainly aim to show the dependence of regret on $T$ similarly to other papers on Thompson Sampling in which constants are not explicitly shown. hence. In the spirit of those papers on Thompson Sampling, we made no further attempt to optimize for the best constants in our regret bounds.
    Nevertheless we show via experiment that in practice, {\sc TS-Cascade}'s empirical performance outperforms what the finite-time bounds (with large constants) suggest. 
    We elaborate on the main features of the algorithm and the guarantee. 
    
In a nutshell, {\sc TS-Cascade} is a Thompson sampling Algorithm \citep{Thompson33}, based on prior-posterior updates on Gaussian random variables with refined variances. The use of the Gaussians is useful,  since it allows us to readily generalize the algorithm and analyses to the contextual setting \citep{LiChuLangford10,LiWZC16}, as well as the linear bandit setting \citep{ZongNSNWK16} for handling a large $L$.
We consider the linear setting in Section~\ref{cascadeTS_sec:lin_alg_cas} and plan to study the contextual setting in the future. 
To this end, we remark that the posterior update of TS can be done in a variety of ways. While the use of a Beta-Bernoulli update to maintain a Bayesian estimate on $w(i)$ is a natural option \citep{RussoRKOW18}, we use Gaussians instead, which is in part inspired by \cite{ZongNSNWK16}, in view of their use in generalizations and its empirical success in the linear bandit setting. 
Indeed, the conjugate prior-posterior update is not the only choice for TS algorithms for complex multi-armed bandit problems. For example, the posterior update in Algorithm 2 in \cite{agrawal2017thompson} for the multinomial logit bandit problem is not conjugate.



While the use of Gaussians is useful for generalizations, the analysis of Gaussian Thompson samples in the cascading setting comes with some difficulties, as $\theta_t(i) $ is not in $[0,1]$ with probability one. We perform a truncation of the Gaussian Thompson sample in the proof of Lemma~\ref{cascadeTS_lemma:gap_bound} to show that this replacement of the Beta by the Gaussian does not incur any significant loss in terms of the regret and the analysis is not affected significantly.

We elaborate on the refined variances of our Bayesian estimates. Lines 5--7 indicate  that the Thompson sample $\theta_t(i)$ is constructed to be a Gaussian random variable with mean $\hat{w}_t(i)$ and  variance being the maximum of $ \hat{\nu}_t(i)\log (t+1) /  (N_t(i) + 1) $ and  $ [ \log (t+1)/ (N_t(i) + 1) ]^2$.   Note that $\hat{\nu}_t(i)$ is the variance of a Bernoulli distribution with mean $\hat{w}_t(i)$. In Thompson sampling algorithms, the choice of the variance is of crucial importance. 
We considered a na\"ive TS implementation initially. However, the theoretical and empirical results were unsatisfactory, due in part to the large variance of the Thompson sample variance; this motivated us to   improve on the algorithm leading to Algorithm~\ref{cascadeTS_alg:ts}.
The reason why we choose the variance in this manner is to (i) make the Bayesian estimates behave like Bernoulli random variables and to (ii) ensure that it is tuned so that the regret bound has a dependence on $\sqrt{K}$ (see Lemma~\ref{cascadeTS_lemma:transfertwo}) and does not depend on any pre-set parameters. We utilize a key result by \cite{AudibertMS09} concerning the analysis of using the empirical variance in multi-arm bandit problems to achieve (i). In essence, in Lemma~\ref{cascadeTS_lemma:transfertwo}, the Thompson sample is shown to depend only on a {\em single} source of randomness, i.e., the Gaussian random variable $Z_t$ (Line~3 of Algorithm~\ref{cascadeTS_alg:ts}). This shaves off a factor of $\sqrt{K}$ vis-\`a-vis a more na\"ive analysis where the variance is pre-set in the relevant probability in Lemma~\ref{cascadeTS_lemma:transfertwo} depends on $K$ independent random variables.
In detail, if instead of Line 3--7 in Algorithm~\ref{cascadeTS_alg:ts}, we sample independent Thompson samples as follows:
\begin{align*}
    \theta_t(i) \sim \mathcal{N} \left( \hat{w}_t(i),  \frac{ K \log(t+1) }{ N_t(i) + 1 } \right)  \quad \forall i\in [L], 
\end{align*}
a similar analysis to our proof of Theorem~\ref{cascadeTS_thm:main} leads to an upper bound 
$ O(  K \sqrt{LT}  \log T) $, which is $\sqrt{K}$ worse than the upper bound for {\sc TS-Cascade}.
Hence, we choose to design our Thompson samples  based on a single one-dimensional Gaussian random variable $Z_t$.

		Next, in Table~\ref{cascadeTS_tab:bound_indep}, we present various problem-independent bounds on the regret for cascading bandits. Table~\ref{cascadeTS_tab:bound_indep} shows that our upper bound grows like $\sqrt{T}$ just like the other bounds. Our bound also matches the state-of-the-art UCB bound (up to log factors) by \cite{wang2017improving}, whose algorithm, when suitably specialized to the cascading bandit setting, is the same as   {\sc CascadeUCB1}  in~\cite{kveton2015cascading}. For the case in which  $T\ge L$, our bound is a $\sqrt{\log T}$ factor worse  than the problem-independent bound in  \cite{wang2017improving}, but we are the first to analyze Thompson sampling for the  cascading bandit problem.

		Moreover, Table~\ref{cascadeTS_tab:bound_indep} shows that the major difference between the upper bound in Theorem~\ref{cascadeTS_thm:main} and the lower bound in Theorem~\ref{cascadeTS_thm:main_lb} is manifested in their dependence on $K$ (the length of recommendation list).  
		(i) On one hand , when $K$ is larger, the agent needs to learn more optimal items. However, instead of the full feedback in the semi-bandit setting, the agent only get partial feedback in the cascading bandit setting and therefore he is not guaranteed to observe more items at each time step. Therefore, a larger $K$ may lead to a larger regret.
		(ii) On the other hand, when $K$ is smaller, the maximum possible amount of observations that the agent can obtain at each time step diminishes. Hence, the agent may obtain less information from the user, which may also result in a larger regret.
		Intuitively, the effect of $K$ on the regret is not clear. We leave the exact dependence of optimal algorithms on $K$ as an open question. 
Sequentially, we present a proof sketch of Theorem~\ref{cascadeTS_thm:main} with several lemmas in order, when some details are postponed to Appendix~\ref{cascadeTS_sec:proofs}.


During the iterations, we update  $\hat{w}_{t+1}(i)$ such that it approaches $w(i)$ eventually. To do so, we select a set $S_t$ according to the order of $\theta_t(i)$'s at each time step. Hence, if $\hat{w}_{t+1}(i)$, $\theta_t(i)$ and $w(i)$   are close enough, then we are likely to select the optimal set. This motivates us to define two  ``nice events'' as follows:
    \begin{align*}
        {\cal E}_{\hat{w}, t}  := \left\{\forall i\in [L] \; : \; \left| \hat{w}_t(i) - w(i) \right| \leq g_t(i)\right\}, \
        {\cal E}_{\theta, t}  := \left\{\forall i\in [L]\; : \; \left| \theta_t(i) - \hat{w}_t(i) \right| \leq h_t(i)\right\},
    \end{align*}
where $\hat{\nu}_t(i)$ is defined in  Line~5 of Algorithm~\ref{cascadeTS_alg:ts}, and
    \begin{align*}
        g_t(i) :=  \sqrt{\frac{16\hat{\nu}_t(i)\log (t + 1)}{N_t(i) + 1}} + \frac{24\log(t + 1)}{N_t(i) + 1}, \
h_t(i) := \sqrt{\log (t + 1)}g_t(i).
\end{align*} 

\begin{restatable}{lemma}{cascadeTSclaimconc}\label{cascadeTS_claim:conc}
    For each $t\in [T]$, $H_t \in {\cal E}_{\hat{w}, t}$, we have
        $$\Pr\left[{\cal E}_{\hat{w}, t}\right] \geq 1- \frac{3L}{(t+1)^3},\quad  
        \Pr\left[{\cal E}_{\theta, t} | H_t \right]\geq 1- \frac{1}{2(t+1)^2}.
        $$
\end{restatable}
Demonstrating that  ${\cal E}_{\hat{w}, t}$ has high probability requires the concentration inequality in Theorem~\ref{cascadeTS_thm:emp_berstein}; this is a specialization of a result in \cite{AudibertMS09} to Bernoulli random variables.  Demonstrating that ${\cal E}_{\hat{\theta}, t}$ has high probability requires the concentration property of Gaussian random variables (cf.\ Theorem~\ref{cascadeTS_thm:gaussian}).


To start our analysis, define
\begin{align}
    F(S,t) :=\sum^K_{k=1}\left[\prod^{k-1}_{j=1}(1\!-\!w(i_j))\right]  \left(g_t(i_k) \!+ \!h_t(i_k)\right)\label{cascadeTS_eq:defF}
\end{align}
and the  the set  
$$
    {\cal S}_t := \Big\{S = (i_1, \ldots, i_K)\in \pi_K(L):
     F(S,t)\geq r(S^* | \bm w) - r(S |\bm w) \Big\}. 
$$
Recall that $w(1)\geq w(2)\geq \ldots \geq w(L)$. As such, ${\cal S}_t$ is non-empty, since  $S^* = (1, 2, \ldots, K)\in {\cal S}_t$. 

\textbf{Intuition behind the set  ${\cal S}_t$:} Ideally, we expect the user to click an item in ${\cal S}_t$ for every time step $t$. Recall that $g_t(i)$ and $h_t(i)$ are decreasing in $N_t(i)$, the number of time steps $q$'s in $1, \ldots, t-1$ when we get to \emph{observe} $W_q(i)$. Naively, arms in ${\cal S}_t$ can be thought of as arms that ``lack observations'', while arms in $\bar{\cal S}_t$ can be thought of as arms that are ``observed enough'', and are believed to be suboptimal. Note that $S^*\in {\cal S}_t$ is a prime example of an arm that is under-observed.

To further elaborate, $g_t(i) + h_t(i)$ is the ``statistical gap'' between the Thompson sample $\theta_t(i)$ and the latent mean $w(i)$. The gap shrinks with more observations of $i$. To balance exploration and exploitation, for any suboptimal item $i\in [L]\setminus [K]$ and any optimal item $k\in [K]$, we should have $g_t(i) + h_t(i) \geq w(k) - w(i)$. However, this is too much to hope for, and it seems that hoping for $S_t\in {\cal S}_t$ to happen would be more viable. (See the forthcoming Lemma~\ref{cascadeTS_lemma:transferone}.)

\textbf{Further notations.} In addition to set  ${\cal S}_t$, we define $H_t$ as the collection of observations of the agent, from the beginning until the end of time   $t-1$. 
More precisely, we define $H_t := \{S_q\}^{t-1}_{q=1} \cup \{\left(i^q_k, W_q(i^q_k)\right)^{\min\{k_t, \infty\}}_{k=1}\}^{t-1}_{q=1}$. Recall that $S_q \in \pi_K(L)$ is the arm  pulled during time step $q$, and $\left(i^q_k, W_q(i^q_k)\right)^{\min\{k_t, \infty\}}_{k=1}$ is the collection of observed items and their respective values during time step $q$.  At the start of time step $t$, the agent has observed everything in $H_{t}$, and determine the arm $S_t$ to pull accordingly (see Algorithm~\ref{cascadeTS_alg:ts}). Note that event ${\cal E}_{\hat{w}, t}$ is $\sigma(H_t)$-measurable. For the convenience of discussion, we define $\mathcal{H}_{\hat{w}, t} := \{H_t : \text{Event ${\cal E}_{\hat{w}, t}$ is true in $H_t$}\}$. The first statement in Lemma~\ref{cascadeTS_claim:conc} is thus $\Pr[H_t\in \mathcal{H}_{\hat{w}, t}] \geq 1 - 3L/(t+1)^3$.

The performance of Algorithm~\ref{cascadeTS_alg:ts} is analyzed using the following four Lemmas. To begin with, Lemma~\ref{cascadeTS_lemma:transferone} quantifies a set of conditions on $\hat{\bm \mu}_t$ and ${\bm \theta}_t$ so that the pulled arm $S_t$ belongs to ${\cal  S}_t$, the collection of arms that lack observations and should be explored. We recall from Lemma~\ref{cascadeTS_claim:conc} that the events ${\cal E}_{\hat{w}, t}$ and ${\cal E}_{\theta, t}$ hold with high probability. Subsequently, we will crucially use our definition of the Thompson sample ${\bm \theta}_t$ to argue that inequality~\eqref{cascadeTS_eq:crucial_ineq} holds with non-vanishing probability when $t$ is sufficiently large. 

\begin{restatable}{lemma}{cascadeTSlemmatransferone}\label{cascadeTS_lemma:transferone}
    Consider a time step $t$. Suppose that events ${\cal E}_{\hat{w}, t},{\cal E}_{\theta, t}$ and inequality 
     \begin{equation}\label{cascadeTS_eq:crucial_ineq}
         \sum^K_{k=1}\left[\prod^{k-1}_{j=1}(1 - w(j))\right]   \theta_t(k)\ge\sum^K_{k=1}\left[\prod^{k-1}_{j=1}(1- w(j))\right]\! w(k)
     \end{equation}
    hold, then the event $\{S_t\in {\cal S}_t\}$ also holds.
\end{restatable}
  In the following, we condition on $H_t$ and show that $\bm{\theta}_t$ is ``typical'' w.r.t.\  $\bm{w}$ in the sense of \eqref{cascadeTS_eq:crucial_ineq}. Due to the conditioning on $H_t$, the only source of the randomness of the pulled arm $S_t$ is from the Thompson sample.   Thus,  by analyzing a suitably weighted version of the Thompson samples in inequality~\eqref{cascadeTS_eq:crucial_ineq}, we disentangle the statistical dependence between partial monitoring and Thompson sampling.
Recall that $\bm{\theta}_t$ is normal with $\sigma(H_t)$-measurable mean and variance (Lines 5--7 in Algorithm~\ref{cascadeTS_alg:ts}).

\begin{restatable}{lemma}{cascadeTSlemmatransfertwo}\label{cascadeTS_lemma:transfertwo}
    There exists an absolute constant $c \in (0, 1)$ independent of $\bm w, K, L, T$ such that, for any time step $t$ and any historical observation $H_t\in \mathcal{H}_{\hat{w}, t}$, the following inequality holds: 
    $$\Pr_{{\bm \theta}_t}\left[\text{${\cal E}_{\theta, t }$ and \eqref{cascadeTS_eq:crucial_ineq} hold}  ~ | ~ H_t\right] \geq  c - \frac{1}{2(t+1)^3}.$$ 
\end{restatable}

\begin{proof} \label{cascadeTS_pf:lemmatransfertwo}
We prove the Lemma by setting the absolute constant $c$ to be $1/(4\sqrt{\pi}e^{8064}) > 0$.  

For brevity, we define  $\alpha(1) := 1$, and $\alpha(k) = \prod^{k-1}_{j=1}(1 - w(j))$ for $2\leq k\leq K$. By the second part of Lemma~\ref{cascadeTS_claim:conc}, we know that $\Pr [ {\cal E}_{\theta, t} | H_t ] \geq 1 - 1/2(t+1)^3$, so to complete this proof, it suffices to show that $\Pr [ \text{(\ref{cascadeTS_eq:crucial_ineq}) holds} | H_t]  \geq  c$. For this purpose, consider
\begin{align}
    &\Pr_{{\bm \theta}_t}\left[\sum^K_{k=1}\alpha(k)\theta_t(k) \geq \sum^K_{k=1}\alpha(k) w(k) ~ \bigg | ~ H_t\right]  
        = \Pr_{Z_t} \left[\sum^K_{k=1}\alpha(k) \big[ \hat{w}_t(k) \! +\!   Z_t  \sigma_t(k) \big] \! \geq \!  \sum^K_{k=1}\alpha(k) w(k) ~ \bigg | ~ H_t \right] \label{cascadeTS_eq:equiv_gaussian}\\
    &\geq \Pr_{Z_t}\left[\sum^K_{k=1}\alpha(k) \left[w(k) - g_t(k)\right] + Z_t \cdot \sum^K_{k=1} \alpha(k)  \sigma_t(k) \geq  \sum^K_{k=1}\alpha(k) w(k) ~ \bigg |~ H_t \right] \label{cascadeTS_eq:by_E_mu_hat}\\
   & =\Pr_{Z_t} \left[ Z_t \cdot \sum^K_{k=1} \alpha(k)\sigma_t(k) \geq  \sum^K_{k=1}\alpha(k)  g_t(k) ~ \bigg | ~ H_t \right] \nonumber\\
   & \geq  \frac{1}{4\sqrt{\pi}}\exp\left\{-\frac{7}{2}\left[ \sum^K_{k=1}\alpha(k)  g_t(k) \bigg / \sum^K_{k=1} \alpha(k) \sigma_t(k)  \right]^2 \right\}\label{cascadeTS_eq:by_anti_conc}\\
    &\geq \frac{1}{4\sqrt{\pi}}\exp\left\{-\frac{7}{2}\left[ \sum^K_{k=1}\alpha(k)  g_t(k)\bigg /  \sum^K_{k=1} \alpha(k) \left(\frac{1}{2}\sqrt{\frac{\hat{\nu}_t(i)\log (t+1)}{N_t(i) + 1}} +  \frac{1}{2}\frac{\log (t+1)}{(N_t(i) + 1)}\right)  \right]^2 \right\}\nonumber \\
%
%
& \geq   \frac{1}{4\sqrt{\pi}e^{8064}}=c. \label{cascadeTS_eq:by_def_g_t}
\end{align}
Step (\ref{cascadeTS_eq:equiv_gaussian}) is by the definition of  $\{\theta_t(i)\}_{i\in L}$ in Line 7 in Algorithm~\ref{cascadeTS_alg:ts}. It is important to note that these samples share the {\em same} random seed $Z_t$.
Next, step~(\ref{cascadeTS_eq:by_E_mu_hat}) is by the Lemma assumption that $H_t\in \mathcal{H}_{\hat{w}, t}$, which means that $\hat{w}_t(k) \geq w_t(k) - g_t(k)$ for all $k\in [K]$. Step~(\ref{cascadeTS_eq:by_anti_conc}) is an application of the anti-concentration inequality of a normal random variable in Theorem~\ref{cascadeTS_thm:gaussian}. Step~(\ref{cascadeTS_eq:by_def_g_t}) is by applying the definition of $g_t(i)$.
\end{proof}

		        We would like to highlight an obstacle in deriving a problem-independent bound for the {\sc CTS} algorithm.
		        In the proof of Lemma~\ref{cascadeTS_lemma:transfertwo}, which is essential in  deriving the bound for {\sc TS-Cascade}, step \eqref{cascadeTS_eq:by_E_mu_hat} results from the fact that the {\sc TS-Cascade} algorithm generates Thompson samples with the \emph{same} random seed $Z_t$ at time step $t$ (Line 3 -- 8 in Algorithm \ref{cascadeTS_alg:ts}).
		        However, {\sc CTS} generate samples for each item \emph{individually}, which hinders the derivation of a similar lemma. 
		        Therefore, we conjecture  that a problem-dependent  bound for CTS cannot be derived using the same strategy as that for {\sc TS-Cascade}.
%


Combining Lemmas~\ref{cascadeTS_lemma:transferone} and~\ref{cascadeTS_lemma:transfertwo}, we conclude that there exists an absolute constant $c$ such that, for any time step $t$ and any historical observation $H_t\in \mathcal{H}_{\hat{w}, t}$, 
\begin{equation}\label{cascadeTS_eq:transfer_ineq}
    \Pr_{{\bm \theta}_t}\left[ S_t\in {\cal S}_t ~ \big | ~ H_t\right] \geq c - \frac{1}{2(t+1)^3}.
\end{equation}

Equipped with \eqref{cascadeTS_eq:transfer_ineq}, we are able to provide an upper bound on the regret of our Thompson sampling algorithm at every sufficiently large time step.
\begin{restatable}{lemma}{cascadeTSlemmagapbound}\label{cascadeTS_lemma:gap_bound}
    Let $c$ be an absolute constant such that Lemma~\ref{cascadeTS_lemma:transfertwo} holds. Consider a time step $t$ that satisfies $c - 1/(t + 1)^3 > 0$. Conditional on an arbitrary but fixed historical observation $H_t\in {\cal H}_{\hat{w}, t}$, we have
    \begin{align*}
        &\mathbb{E}_{\bm \theta_t}[ r(S^* |{\bm w}) - r(S_t |{\bm w}) | H_t] 
        \leq\left(1 + \frac{4}{c}\right) \mathbb{E}_{\bm \theta_t}\left[ F(S_t,t) ~\big | ~ H_t\right] +\frac{L}{2(t+1)^2}.
    \end{align*}
\end{restatable}
The proof of Lemma~\ref{cascadeTS_lemma:gap_bound} relies crucially on {\em truncating} the original  Thompson sample ${\bm \theta}_t   \in\mathbb{R}$ to $\tilde{{\bm \theta}}_t \in [0,1]^L$. Under this truncation operation,   $S_t$ remains optimal under $\tilde{{\bm \theta}}_t$ (as it was under ${{\bm \theta}}_t$) and $|\tilde{\theta}_t(i)-w(i)|\le|\theta_t(i)-w(i)|$, i.e., the distance from the truncated Thompson sample to the ground truth is not increased.

For any  $t$ satisfying $c - 1/(t+1)^3 > 0$, define 
\begin{align*}
    {\cal F}_{i, t}  := \left\{ \text{Observe $W_t(i)$ at $t$}\right \}, \quad
    G(S_t, {\bm W}_t)  := \sum^K_{k=1}\mathsf{1}\big( {\cal F}_{i_k^t, t} \big) \cdot ( g_t(i^t_k)  +  h_t(i^t_k))\big),
\end{align*}
we unravel the upper bound in Lemma~\ref{cascadeTS_lemma:gap_bound} to establish the expected regret at time step $t$:
\begin{align}
        & \mathbb{E}\left\{ r(S^* |{\bm w}) - r(S_t |{\bm w})  \right\} 
           \leq  \mathbb{E}\left[\mathbb{E}_{\bm \theta_t}[ r(S^* |{\bm w}) - r(S_t |{\bm w})~ | ~H_t]\cdot \mathsf{1}(H_t\in \mathcal{H}_{\hat{w}, t})\right]  
               + \mathbb{E}\left[ \mathsf{1}(H_t\not\in \mathcal{H}_{\hat{w}, t}) \right]\nonumber\\
    &\leq   \left(1 + \frac{4}{c}\right)\mathbb{E}\left[ \mathbb{E}_{\bm \theta_t}\left[ F(S_t,t) ~\big | ~ H_t\right] \mathsf{1}(H_t\in \mathcal{H}_{\hat{w}, t}) \right] + \frac{1}{2(t+1)^2}+ \frac{3L}{(t+1)^3}\nonumber\\
    &\leq  \left(1 + \frac{4}{c} \right)\mathbb{E}\left[ F(S_t,t) \right] + \frac{4L}{(t+1)^2}\label{cascadeTS_eq:assume_delta}\\
    &=   \left(1 + \frac{4}{c} \right) \mathbb{E}\left[ \mathbb{E}_{\bm W_t}[ G(S_t, {\bm W}_t) \big | H_t, S_t ]\right] + \frac{4L}{(t+1)^2}\nonumber\\
&=   \left(1 + \frac{4}{c} \right) \mathbb{E}\left[ G(S_t, {\bm W}_t) \right] + \frac{4L}{(t+1)^2}\label{cascadeTS_eq:to_sum},
\end{align}
where   (\ref{cascadeTS_eq:assume_delta}) follows by assuming $t$ is sufficiently large.

\begin{restatable}{lemma}{cascadeTSlemmatelescope}\label{cascadeTS_lemma:telescope}
    For any realization of historical trajectory ${\cal H}_{T+1}$,   we have  
   \begin{align*}
        & \sum^T_{t = 1} G(S_t, {\bm W}_t) \leq 6  \sqrt{K L T}\log T + 144L \log^{5/2}T. 
   \end{align*}
\end{restatable}
\begin{proof}
\label{cascadeTS_pf:lemmatelescope}
Recall that for each $i\in [L]$ and $t\in [T+1]$, $N_{t}(i) = \sum^{t-1}_{s=1}\mathsf{1}({\cal F}_{i,s})$ is the number of rounds in $[t-1]$ when we get to observe the outcome for item $i$.  Since $G(S_t, {\bm W}_t)$ involves $g_t(i)+h_t(i)$, we first bound this term. The definitions of $g_t(i)$ and $h_t(i)$ yield that
 \begin{align*}
    g_t(i) + h_t(i) 
    &  \leq \frac{12\log (t+1)}{\sqrt{N_t(i) + 1}} + \frac{72\log^{3/2}(t+1)}{N_t(i) + 1}.
\end{align*} 
Subsequently, we decompose $\sum^T_{t = 1} G(S_t, {\bm W}_t) $ according to its definition. For a fixed but arbitrary item $i$, consider the sequence $(U_t(i))^T_{t=1} = ( \mathsf{1}\left( {\cal F}_{i, t}\right) \cdot ( g_t(i)  +  h_t(i)) )^T_{t = 1}$. Clearly, $U_t(i) \neq 0$ if and only if the decision maker observes the realization $W_t(i)$ of item $i$ at $t$. Let $t = \tau_1 < \tau_2 < \ldots < \tau_{N_{T+1}}$ be the time steps when $U_t(i) \neq 0$. We assert that $N_{\tau_n}(i) = n-1$ for each $n$. Indeed, prior to time steps $\tau_n$, item $i$ is observed precisely in the time steps $\tau_1, \ldots, \tau_{n-1}$. Thus, we have 
\begin{align}
 &     \sum^T_{t = 1}\mathsf{1}\left( {\cal F}_{i, t}\right) \cdot ( g_t(i)  +  h_t(i)) = \sum^{N_{T+1}(i)}_{n=1} ( g_{\tau_n}(i)  +  h_{\tau_n}(i)) 
 \leq   \sum^{N_{T+1}(i)}_{n=1} \frac{12\log T}{\sqrt{n}} + \frac{72\log^{3/2}T}{n}.\label{cascadeTS_eq:bound_G}
 \end{align}
 Now we complete the proof as follows:
\begin{align}
    &\sum^T_{t =  1} \sum^K_{k=1}\mathsf{1}\big( {\cal F}_{i_k^t, t}\big) \cdot ( g_t(i^t_k)  +  h_t(i^t_k)) 
    = \sum_{i\in [L]}\sum^T_{t = 1}\mathsf{1}\left( {\cal F}_{i, t}\right) \cdot ( g_t(i)  +  h_t(i))\nonumber\\ 
&    \leq \sum_{i\in [L]}\sum^{N_{T+1}(i)}_{n = 1}\frac{12\log T}{\sqrt{n}} + \frac{72\log^{3/2}T}{n}\label{cascadeTS_eq:observation_accounting}
    \leq   6 \sum_{i\in [L]} \sqrt{N_{T+1}(i)}\log T + 72 L \log^{3/2}T (  \log T + 1) \\* 
&\leq  6  \sqrt{L \sum_{i\in [L]} N_{T+1}(i)}\log T + 72 L \log^{3/2}T (  \log T + 1)
    \leq   6  \sqrt{KLT}\log T + 144 L \log^{5/2}T  , \label{cascadeTS_eq:final_bounded_set}
\end{align}
where step~(\ref{cascadeTS_eq:observation_accounting}) follows from step~\eqref{cascadeTS_eq:bound_G}, 
the first inequality of step~\eqref{cascadeTS_eq:final_bounded_set} follows from the Cauchy-Schwarz inequality,and the second one is because the decision maker can observe at most $K$ items at each time step, hence $\sum_{i\in [L]}N_{T+1}(i)\leq KT$.\end{proof} 

Finally, we bound the total regret from above by considering the time step $t_0 := \lceil 1/c^{1/3} \rceil $, and then bound the regret for the time steps before $t_0$ by 1 and the regret for time steps after by inequality (\ref{cascadeTS_eq:to_sum}), which holds for all $t>t_0$: 
\begin{align}
    \mathrm{Reg}(T) 
    & \leq \bigg \lceil \frac{1}{c^{1/3}} \bigg \rceil + \sum^{T}_{t= t_0 + 1}\mathbb{E}\left\{ r(S^* |{\bm w}) - r(S_t |{\bm w})\right\} \nonumber\\
    & \leq  \bigg \lceil \frac{1}{c^{1/3}} \bigg \rceil +  \left(1 + \frac{4}{c} \right) \mathbb{E}\left[\sum^{T}_{t= 1} G(S_t, {\bm W}_t) \right] + \sum^T_{t = t_0+1}\frac{4L}{(t+1)^2}.\nonumber
\end{align}
It is clear that the third term is $O(L)$, and by Lemma~\ref{cascadeTS_lemma:telescope}, the second term is $O(\sqrt{KLT}\log T + L \log^{5/2}T)$.  Altogether, Theorem~\ref{cascadeTS_thm:main} is proved.

\section{Linear generalization} \label{cascadeTS_sec:lin_alg_cas}
 
In the standard cascading bandit problem, when the number of ground items $L$ is large, the algorithm converges slowly. 
In detail, the agent learns each $w(i)$ individually, and the knowledge in $w(i)$ does not help in the estimation of $w(j)$ for $j\neq i$. 
	To ameliorate this problem, we consider the linear generalization setting as described in Section~\ref{cascadeTS_sec:intro}. Recall that $w(i)= x(i)^T \beta$, where $x(i)$'s are known feature vectors, but $\beta$ is an unknown universal vector. Hence, the problem reduces to estimating $\beta\in \bbR^d$, which, when $d<L$, is easier than estimating the $L$ distinct weights $w(i)$'s in the standard cascading bandit problem.

    
We construct a Thompson Sampling algorithm {\algCasLin} to solve the cascading bandit problem, and provide a theoretical guarantee for this algorithm.

\begin{algorithm}[h]
	\caption{\algCasLin}\label{cascadeTS_alg:casLinTS'}
		\begin{algorithmic}[1]
			\State Input feature matrix $X_{d\times L } = [ x(1)^T, \ldots, x(L)^T ]$ and parameter $\lambda$. Set $R =1/2$.
			\State Initialize $\bLin_1 = \mathbf{0}_{d\times 1}$,  $\NLin_1 = I_{d\times d}$,  $\hat{\muLin}_1 = \NLin_1^{-1} \bLin_1$.
			\For{$t = 1, 2, \ldots$}
			    \State Sample a $d$-dimensional random variable $\ZLin_t\sim {\cal N}( \mathbf{0}_{d\times 1}, I_{d\times d})$. \label{cascadeTS_alg:ts_gaussian_lin}
			    \State Construct the Thompson sample 
                	$
                    	v_t =3\RLin \sqrt{d\log t},\ \thetaLin_t = \hat{\muLin}_t+ \lambda v_t \sqrt{K}\NLin_t^{-1/2} \ZLin_t.
                    $
				\For{$k\in [K]$}
				    \State Extract $i^t_k\in \text{argmax}_{i\in [L]\setminus \{i^t_1, \ldots, i^t_{k-1}\}} x(i)^T \thetaLin_t$.
				\EndFor
				\State Pull arm $S_t = (i^t_1, i^t_2, \ldots, i^t_K)$. \label{cascadeTS_alg:ts_opt'}
				\State Observe click $k_t \in \{ 1, \ldots,K, \infty \}$.
				\State Update $M_t, B_t$ and $\hat{\muLin}_t$ with observed item
                \begin{align*}
				    \NLin_{t+1} = \NLin_{t} +  \sum_{ k=1}^{ \min\{k_t, K \} }   x(i_k) x(i_k)^T ,\
				    \bLin_{t+1} = \bLin_{t} + \sum_{ k=1}^{ \min\{k_t, K \} } x(i_k) W_t(i_k),\
				    \hat{\muLin}_{t+1} = \NLin_{t+1}^{-1} \bLin_{t+1}.
				\end{align*}
			\EndFor
		\end{algorithmic}
	\end{algorithm}
	
   
        \begin{theorem}[Linear generalization] \label{cascadeTS_thm:mainLin}
        Assuming the $l_2$ norms of feature parameters are bounded by
$1/\sqrt{K}$, i.e., $\| x(i) \| \le 1/\sqrt{K}$
        for all $i \in [L]$,
        \footnote{
        It is realistic to assume that $\| x(i) \| \le 1/\sqrt{K}$ for all $i\in[L]$ since this can always be ensured by rescaling the feature vectors.
        If instead, we assume that  $\| x(i) \| \le 1$ as in \cite{GaiKJ12,WenKA15,ZongNSNWK16}, the result in Lemma~\ref{cascadeTS_lemma:telescopeLin} will be $\sqrt{K}$ worse, which would lead to an upper bound larger than the bound in Theorem~\ref{cascadeTS_thm:mainLin} by a factor of $\sqrt{K}$.
        } 
        Algorithm \algCasLin, presented in Algorithm~\ref{cascadeTS_alg:casLinTS'}, incurs an expected regret at most
        $$O\left( \min \{ \sqrt{d}, \sqrt{\log L} \} \cdot K d\sqrt{T} (\log T)^{3/2} \right) .$$
        Here the big $O$ notation depends only on $\lambda$.
    \end{theorem}

The upper bound in Theorem~\ref{cascadeTS_thm:mainLin} diminishes when $\lambda$ grows.
Recall that for {\sc CascadeLinUCB}, \cite{ZongNSNWK16} derived an upper bound in the form of $\tilde{O}(Kd\sqrt{T} )$; and  for  Thompson sampling applied to linear contextual bandits, \cite{AgrawalG13b} proved a high probability regret bound of $\tilde{O}(d^{3/2}\sqrt{T} )$. Our result is a factor $\sqrt{d}$ worse than the bound on {\sc CascadeLinUCB}. However, our bound matches that of Thompson sampling for contextual bandits with $K=1$ up to log factors.
As \cite{ZongNSNWK16} proposed {\sc CascadeLinTS} without analysis, we are the first to analyze a Thompson sampling algorithm for the linear cascading bandit setting.
Different from {\sc CascadeLinTS}, the agent needs to choose a parameter $\lambda$ for \algCasLin. As shown in Line 5 of Algorithm~\ref{cascadeTS_alg:casLinTS'}, we apply $\lambda$ to scale the variance of Thompson samples.
We see that when the given set of feature vectors are ``well-conditioned'', i.e., the set of feature vectors of optimal items and suboptimal items are far away from each other in the feature space $[0,1]^d$, then the agent should should choose a small $\lambda$ to simulate exploitation.
Otherwise, a large $\lambda$ encourages more exploration.


Now we provide a proof sketch of Theorem~\ref{cascadeTS_thm:mainLin}, which is similar to that of Theorem~\ref{cascadeTS_thm:main}. More details are given in Appendix~\ref{cascadeTS_sec:proofs}.

    To begin with, we observe that a certain noise random variable possesses the sub-Gaussian property~(same as Appendix A.1. of  \cite{filippi2010parametric}), and we assume $x(i)$ is bounded for each~$i$:

\begin{remark}[Sub-Gaussian property]\label{cascadeTS_remark:sub_gauss}
    Define $\eta_{t}(i) := W_t(i) - x(i)^T \beta$ and let $R=1/2$. The fact that $\bbE W_t(i) = w(i) =x(i)^T \beta  $ and $W_t(i)\in [0,1]$ yields that $\bbE \eta_t (i) = \bbE W_t(i) - w(i) = 0$ and $\eta_{t}(i) \in [  - w(i), 1 - w(i) ]$. Hence, $\eta_{t}(i)$ is $0$-mean and $R$-sub-Guassian, i.e., 
    $$\bbE[ \exp(\lambda \eta_t) | \calF_{t-1} ] \le \exp \Big( \frac{\lambda^2 R^2}{2} \Big) \quad \forall \lambda \in \bbR .$$
\end{remark}

During the iterations, we update  $\hat{\muLin}_{t+1}$ such that $x(i)^T \hat{\muLin}_{t+1}$ approaches $x(i)^T\beta$ eventually. To do so, we select a set $S_t$ according to the order of $x(i)^T \thetaLin_t$'s at each time step. Hence, if $x(i)^T \hat{\muLin}_{t+1}$, $x(i)^T \thetaLin_t$ and $x(i)^T \beta$ are close enough, then we are likely to select the optimal set. 
 
Following the same ideas as in the standard case, this motivates us to define two  ``nice events'' as follows:
    \begin{align*}
        &
        {\cal E}_{\hat{\muLin}, t}   := \left\{\forall i\in [L] \; : \; \left| x(i)^T \hat{\muLin}_t - x(i)^T \beta \right| \leq g_t(i)\right\}, \quad 
        \\ &
        {\cal E}_{\thetaLin, t}  := \left\{\forall i\in [L]\; : \; \left|   x(i)^T \thetaLin_t -  x(i)^T \hat{\muLin}_t \right| \leq h_t(i)\right\},
    \end{align*}
    where
    \begin{align*}
    	& v_t: = 3\RLin \sqrt{d \log t},\quad
    	  l_t: = \RLin \sqrt{3d \log t} + \cZeroLin, \quad
    	  u_t(i): = \sqrt{ x(i)^T \NLin_t^{-1} x(i) }, \\
    	& g_t(i): = l_t u_t(i), \quad
    	  h_t(i): = \min \{ \sqrt{d}, \sqrt{\log L} \} \cdot \sqrt{4K\log t}\cdot \lambda v_t u_t(i).
    \end{align*}
Analogously to Lemma~\ref{cascadeTS_claim:conc}, Lemma~\ref{cascadeTS_claim:concLin} illustrates that  ${\cal E}_{\hat{\muLin}, t}$ has high probability. This follows from the concentration inequality in Theorem~\ref{cascadeTS_thm:linear}
, and the analysis is similar to that of Lemma 1 in \cite{AgrawalG13b}. It also illustrates that ${\cal E}_{ {\thetaLin}, t}$ has high probability, which is a direct consequence of the concentration bound in Theorem~\ref{cascadeTS_thm:gaussian}.
\begin{restatable}{lemma}{cascadeTSclaimconcLin}\label{cascadeTS_claim:concLin}
    For each $t\in [T]$, we have 
        $$\Pr\left[{\cal E}_{\hat{\muLin}, t}\right] \geq 1- \frac{1}{t^2},\quad  
        \Pr\left[{\cal E}_{\thetaLin, t} | H_t \right]\geq 1 - \prHlin.
        $$
\end{restatable}

Next, we revise the definition of $F(S,t)$ in equation~\eqref{cascadeTS_eq:defF} for the linear generalization setting: 
\begin{align}
    F(S,t) &:= \sum^K_{k=1}\left[\prod^{k-1}_{j=1}(1\!-\!x(i_j)^T \beta )\right]  \left(g_t(i_k) +h_t(i_k)\right),\label{cascadeTS_eq:defFLin}\\*
    {\cal S}_t &:= \Big\{S = (i_1, \ldots, i_K)\in \pi_K(L): F(S,t)\geq r(S^* |  w) - r(S | w) \Big\}.
\end{align}
Further, following the intuition of Lemma~\ref{cascadeTS_lemma:transferone} and Lemma~\ref{cascadeTS_lemma:transfertwo}, we state and prove Lemma~\ref{cascadeTS_lemma:transferoneLin} and Lemma~\ref{cascadeTS_lemma:transfertwoLin} respectively. These lemmas are similar in spirit to Lemma~\ref{cascadeTS_lemma:transferone} and Lemma~\ref{cascadeTS_lemma:transfertwo} except that $\theta_t(k)$ in Lemma~\ref{cascadeTS_lemma:transferone} is replaced by $x(k)^T \thetaLin_t$ in Lemma~\ref{cascadeTS_lemma:transferoneLin}, when inequality~\eqref{cascadeTS_eq:crucial_ineq} and $c$ in Lemma~\ref{cascadeTS_lemma:transfertwo} are replaced by inequality~\eqref{cascadeTS_eq:crucial_ineqLin} and $\clin$ in Lemma~\ref{cascadeTS_lemma:transfertwoLin} respectively.

\begin{restatable}{lemma}{cascadeTSlemmatransferoneLin}\label{cascadeTS_lemma:transferoneLin}
    Consider a time step $t$. 
    Suppose that events ${\cal E}_{\hat{\muLin}, t},{\cal E}_{\thetaLin, t}$ and inequality 
     \begin{equation}\label{cascadeTS_eq:crucial_ineqLin}
         \sum^K_{k=1}\left[\prod^{k-1}_{j=1}(1 - w(j))\right]   x(k)^T \thetaLin_t\ge\sum^K_{k=1}\left[\prod^{k-1}_{j=1}(1- w(j))\right]\! w(k)
     \end{equation}
    hold, then the event $\{S_t\in {\cal S}_t\}$ also holds.
\end{restatable}

%

\begin{restatable}{lemma}{cascadeTSlemmatransfertwoLin}\label{cascadeTS_lemma:transfertwoLin}
    There exists an absolute constant $\clin =1/(4\sqrt{\pi}e^{7/2\lambda^2})\in (0, 1)$ independent of $ w, K, L, T$ such that, for any time step $t\ge 4$ and any historical observation $H_t\in \mathcal{H}_{\hat{\muLin}, t}$, the following inequality holds: 
    $$\Pr_{{ \thetaLin}_t}\left[\text{${\cal E}_{\thetaLin, t }$ and \eqref{cascadeTS_eq:crucial_ineqLin} hold}  ~ | ~ H_t\right]~\geq~\clin- \prHlin.$$ 
\end{restatable}

Above yields that there exists an absolute constant $\clin$ such that, for any time step $t$ and any historical observation $H_t\in \mathcal{H}_{\hat{\muLin}, t}$, 
\begin{equation}\label{cascadeTS_eq:transfer_ineq_lin}
    \Pr_{{ \thetaLin}_t}\left[ S_t\in {\cal S}_t ~ \big | ~ H_t\right]~\geq~\clin - \prHlin.
\end{equation}

This allows us to upper bound the regret of {\algCasLin} at every sufficiently large time step. The proof of this lemma is similar to that of Lemma~\ref{cascadeTS_lemma:gap_bound} except that $\theta_t(k)$ in Lemma~\ref{cascadeTS_lemma:gap_bound} is replaced by $x(k)^T \theta_t$. We omit the proof for the sake of simplicity.
 
\begin{restatable}{lemma}{cascadeTSlemmagapboundLin}\label{cascadeTS_lemma:gap_boundLin}
    Let $\clin$ be an absolute constant such that Lemma~\ref{cascadeTS_lemma:transfertwoLin} holds. Consider a time step $t$ that satisfies $\clin > 2d/t^2$, $t>4$. Conditional on an arbitrary but fixed historical observation $H_t\in {\cal H}_{\hat{\muLin}, t}$, we have
    \begin{align*}
        &\mathbb{E}_{ \thetaLin_t}[ r(S^* |{ w}) - r(S_t |{ w}) | H_t]
         \leq\left(1 + \frac{4}{\clin }\right) \mathbb{E}_{ \thetaLin_t}\left[ F(S_t,t) ~\big | ~ H_t\right] + \prHlin. 
    \end{align*}
\end{restatable}


For any  $t$ satisfying $\clin > 2/t^2$, $t>4$, recall that
\begin{align*}
    {\cal F}_{i, t}  := \left\{ \text{Observe $W_t(i)$ at $t$}\right \}, \quad
    G(S_t, { W}_t)  := \sum^K_{k=1}\mathsf{1}\big( {\cal F}_{i_k^t, t} \big) \cdot ( g_t(i^t_k)  +  h_t(i^t_k))\big),
\end{align*}
we unravel the upper bound in Lemma~\ref{cascadeTS_lemma:gap_boundLin} to establish the expected regret at time step $t$:
\begin{align}
        \mathbb{E}\left\{ r(S^* |{ w}) - r(S_t |{ w})  \right\} 
    &\leq  \mathbb{E}\left[\mathbb{E}_{ \thetaLin_t}[ r(S^* |{ w}) - r(S_t |{ w})~ | ~H_t]\cdot \mathsf{1}(H_t\in \mathcal{H}_{\hat{\muLin}, t})\right]   + \mathbb{E}\left[ \mathsf{1}(H_t\not\in \mathcal{H}_{\hat{\muLin}, t}) \right]\nonumber\\
    &\leq   \left(1 + \frac{4}{c}\right)\mathbb{E}\left[ \mathbb{E}_{ \thetaLin_t}\left[ F(S_t,t) ~\big | ~ H_t\right] \mathsf{1}(H_t\in \mathcal{H}_{\hat{\muLin}, t}) \right] +\prHlin + \frac{1}{t^2}\label{cascadeTS_eq:assume_delta_Lin}\\
&=   \left(1 + \frac{4}{c} \right) \mathbb{E}\left[ G(S_t, { W}_t) \right] + \frac{d+1}{t^2},\label{cascadeTS_eq:to_sum_Lin}
\end{align}
where step (\ref{cascadeTS_eq:assume_delta_Lin}) follows by assuming $t$ is sufficiently large.
\begin{restatable}{lemma}{cascadeTSlemmatelescopeLin}\label{cascadeTS_lemma:telescopeLin}
    For any realization of historical trajectory ${\cal H}_{T+1}$,   we have  
   \begin{align*}
        & \sum^T_{t = 1} G(S_t, { W}_t) \leq  40\RLin(1+\lambda) \min \{ \sqrt{d}, \sqrt{\log L} \} \cdot K d\sqrt{T} (\log T)^{3/2} .
   \end{align*}
\end{restatable}
Note that here similarly to Lemma~\ref{cascadeTS_lemma:telescope}, we prove a worst case bound without needing the expectation operator.
\begin{proof}
\label{cascadeTS_pf:lemmatelescopeLin}
Recall that for each $i\in [L]$ and $t\in [T+1]$, $N_{t}(i) = \sum^{t-1}_{s=1}\mathsf{1}({\cal F}_{i,s})$ is the number of rounds in $[t-1]$ when we get to observe the outcome for item $i$.  Since $G(S_t, { W}_t)$ involves $g_t(i)+h_t(i)$, we first bound this term. The definitions of $g_t(i)$ and $h_t(i)$ yield that
 \begin{align*}
    & g_t(i) + h_t(i)   
      = u_t(i) [l_t + \min \{ \sqrt{d}, \sqrt{\log L} \} \cdot \lambda \sqrt{4K\log t}\cdot v_t  ]\\
     & = u_t(i) [\RLin \sqrt{3d \log t} + \cZeroLin + \min \{ \sqrt{d}, \sqrt{\log L} \} \cdot \lambda \sqrt{4K\log t}\cdot 3\RLin \sqrt{d \log t}  ]\\
     & \le 8 \RLin (1+\lambda) u_t(i) \log t\sqrt{dK} \cdot \min \{ \sqrt{d}, \sqrt{\log L} \}.      
\end{align*} 
Further, we have
 \begin{align*}
    & \sum^T_{t = 1} G(S_t, { W}_t)  
        = \sum^T_{t =  1} \sum^K_{k=1}\mathsf{1}\big( {\cal F}_{i_k^t, t}\big) \cdot [ g_t(i^t_k)  +  h_t(i^t_k) ] 
     = \sum^T_{t =  1} \sum^{k_t}_{k=1} [ g_t(i^t_k)  +  h_t(i^t_k) ]\\
     &  \le 8\RLin(1+\lambda)\log T\sqrt{dK} \cdot \min \{ \sqrt{d}, \sqrt{\log L} \} \sum^T_{t =  1} \sum^{k_t}_{k=1} u_t(i^t_k) .
\end{align*}

Let $x(i) = (x_1(i), x_2(i),\ldots, x_d(i) )$, $y_t = ( y_{t,1}, y_{t,2}, \ldots, y_{t,d} )$ with
\begin{align*}
    y_{t,l} = \sqrt{ \sum^{k_t}_{k=1} x_l(i^t_k)^2 }, \ \forall l = 1,2,\ldots,d,
\end{align*}
then $\NLin_{t+1} = \NLin_t + y_t y_t^T$, $\| y_t \| \le 1$.


Now, we have
\begin{align}
    & \sum_{t=1}^T \sqrt{ y_t^T \NLin_t^{-1} y_t } \le 5\sqrt{dT\log T}, \label{cascadeTS_eq:bound_y_other}
\end{align}
which can be derived along the lines of Lemma 3 in \cite{chu2011contextual} using Lemma~\ref{cascadeTS_lemma:bound_y_other} (see Appendix~\ref{cascadeTS_sec:bound_y_other} for details).
Let $\NLin(t)^{-1}_{m,n}$ denote the $(m,n)^{\mathrm{th}}$-element of $\NLin(t)^{-1}$. Note  that
 \begin{align}
      & \sum_{k=1}^{k_t} u_t(i_k^t)  =  \sum_{k=1}^{k_t} \sqrt{ x(i_k^t)^T \NLin(t)^{-1} x(i_k^t) } \label{cascadeTS_eq:def_u_t}\\
      & \le \sqrt{ k_t  \cdot \sum_{k=1}^{k_t} x(i_k^t)^T \NLin(t)^{-1} x(i_k^t) } \label{cascadeTS_eq:cauchy_u}  
          \le \sqrt{ K} \sqrt{  \sum_{1\le m,n \le d}  \NLin(t)^{-1}_{m,n} \left[ \sum_{k=1}^{k_t} x_m(i_k^t) x_n(i_k^t)\right] } \\ 
      & \le \sqrt{ K} \sqrt{  \sum_{1\le m,n \le d}  \NLin(t)^{-1}_{m,n} \sqrt{ \left[ \sum_{k=1}^{k_t} x_m(i_k^t)^2 \right] \cdot \left[ \sum_{k=1}^{k_t} x_n(i_k^t)^2 \right] } } \label{cascadeTS_eq:cauchy_x_element} 
          = \sqrt{ K} \sqrt{  \sum_{1\le m,n \le d}  \NLin(t)^{-1}_{m,n} y_{t,m} y_{t,n} } \\
      & = \sqrt{K} \sqrt{ y_t^T \NLin_t^{-1} y_t }. \nonumber 
      %
 \end{align}
 where step~\eqref{cascadeTS_eq:def_u_t} follows from the definition of $u_t(i)$, \eqref{cascadeTS_eq:cauchy_u} and \eqref{cascadeTS_eq:cauchy_x_element} follow from Cauchy-Schwartz inequality.
 
 Now we complete the proof as follows:
 \begin{align*}
    & 
    \sum^T_{t = 1} G(S_t, { W}_t)  
    \le 8\RLin(1+\lambda)\log T\sqrt{dK} \cdot \min \{ \sqrt{d}, \sqrt{\log L} \} \sum^T_{t =  1} \sum^{k_t}_{k=1} u_t(i^t_k) 
    \\ &
    \le 8\RLin(1+\lambda)\log T\sqrt{dK} \cdot \min \{ \sqrt{d}, \sqrt{\log L} \} \cdot \sum^T_{t =  1} \sqrt{K} \cdot \sqrt{ y_t^T \NLin_t^{-1} y_t }
    \\ &
    \le 40\RLin(1+\lambda) \min \{ \sqrt{d}, \sqrt{\log L} \} \cdot K d\sqrt{T} (\log T)^{3/2} .
\end{align*}

\end{proof}

Finally, we bound the total regret from above by considering the time step $\tlin := \lceil  2/ \clin \rceil =\lceil 8\sqrt{\pi}e^{7/2\lambda^2}\rceil$, (noting $\clin > 2d/t^2$, $t>4$ when $t\ge \tlin$) and then bound the regret for the time steps before $t_0$ by 1 and the regret for time steps after by inequality (\ref{cascadeTS_eq:to_sum_Lin}), which holds for all $t>\tlin$: 
\begin{align}
     \mathrm{Reg}(T) & \leq 
        \bigg \lceil \frac{2}{\clin} \bigg \rceil + \sum^{T}_{t= \tlin + 1}\mathbb{E}\left\{ r(S^* |{ w}) - r(S_t |{ w})\right\} \nonumber\\* 
    &\leq  \bigg \lceil \frac{2}{\clin} \bigg \rceil +  \left(1 + \frac{4}{\clin} \right) \mathbb{E}\left[\sum^{T}_{t= 1} G(S_t, { W}_t) \right]   + \sum^T_{t = \tlin+1} \frac{d+1}{t^2}  .\nonumber
\end{align}
It is clear that the first term is $O(1)$ and the third term is $O(d)$, and by Lemma~\ref{cascadeTS_lemma:telescopeLin}, the second term is 
 $ O\big( \min \{ \sqrt{d}, \sqrt{\log L} \} \cdot K d\sqrt{T} (\log T)^{3/2} \big).$ Altogether, Theorem~\ref{cascadeTS_thm:mainLin} is proved. 
   



\section{Lower bound on regret of the standard setting}
\label{cascadeTS_sec:lb_cas_reg}
We derive a problem-independent minimax lower bound on the regret for the standard cascading bandit problem, following the exposition in \cite[Theorem~3.5]{bubeck2012regret}.
	\begin{theorem}[Problem-independent] \label{cascadeTS_thm:main_lb}
        Any algorithm for the cascading bandit problem incurs an expected regret at least
        $$\Omega\left( \sqrt{LT/K} \right).$$
        Here the big $\Omega$ notation hides a constant factor that is independent of $K, L, T,  \bm{w}$.
    \end{theorem}

    Now we outline the proof of Theorem~\ref{cascadeTS_thm:main_lb}; the accompanying lemmas are proved in Appendix~\ref{cascadeTS_sec:proofs}.  The theorem is derived probabilistically: we construct several difficult instances such that the difference between click probabilities of optimal and suboptimal items is subtle in each of them, and hence the regret of any algorithm must be large when averaged over these distributions. 

{\bf Notations.} Before presenting the proof, we remind the reader of the definition of the KL divergence~\citep{cover2012elements}. For two probability mass functions 
$P_X$ and $P_Y$ defined on the same discrete probability space 
${\cal A}$, 
$$\mathrm{KL}(P_X,P_Y) = \sum_{ x\in{\cal A}} P_X(x) \log \frac{ P_X (x) }{ P_Y(x) }$$
 denotes the KL divergence between the probability mass functions $P_X$, $P_Y$. 
 Lastly, when $a$ and $b$ are two real numbers between $0$ and $1$, $\mathrm{KL}(a,b)= \mathrm{KL} \left( \text{Bern}(a) \|\text{Bern}(b) \right)$, i.e., $\mathrm{KL}(a,b)$ denotes the KL divergence between $\text{Bern}(a)$ and $\text{Bern}(b)$.


\subsection*{First step: Construct instances}
\begin{figure}[t]
	\centering
	\includegraphics[width=\textwidth]{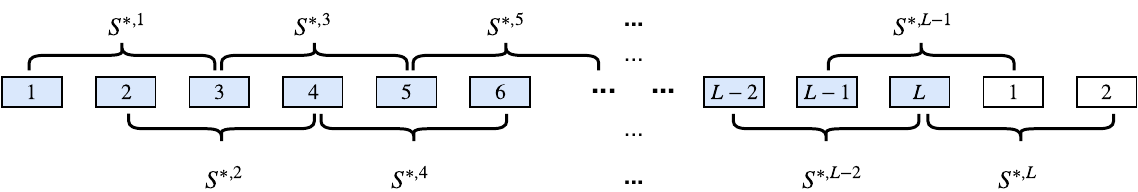}	
	\caption{An example of instances with $K=3$}
	\label{cascadeTS_pic:instance_lb}
\end{figure}

{\bf Definition of instances.} To begin with, we define a class of $L+1$ instances, indexed by $\ell = 0, 1, \ldots, L$ as follows:
\begin{itemize}
	\setlength{\itemindent}{0pt}
	\item Under instance 0, we have $w^{(0)}(i) = \frac{1-\epsilon}{K}$ for all $i \in [L]$.
	\item Under instance $\ell$, where $1\leq \ell\leq L$, we have $w^{(\ell)}(i) = \frac{1 + \epsilon}{K}$ if $i \in \{\underline{\ell}, \underline{\ell + 1}, \ldots, \underline{\ell +K - 1}\}$, where $\underline{a} = a$ if $1\leq a\leq L$, and $\underline{a} = a - L$ if $L < a \leq 2L$. We have $w^{(\ell)}(i) = \frac{1 - \epsilon}{K}$ if $i \in [L]\setminus \{\underline{\ell}, \underline{\ell+ 1}, \ldots, \underline{\ell +K - 1}\}$,
\end{itemize}
where 
$\epsilon \in [0, 1]$ is a small and positive number such that $0 <(1-\epsilon) / K < (1+\epsilon) / K < 1/4$.
The expected reward of a list $S$ under instance $\ell (0\le \ell \le L)$ is 
	$$ r(S | w^{(\ell)}) = 1 - \prod_{i\in S}(1 - w^{(\ell)}(i)).$$ 
Under instance $0$, every list $S \in [L]^{(K)}$ is optimal, while under instance $\ell$ $(1\le \ell \le L)$, a list $S$ is optimal iff $S$ is an ordered $K$-set comprised of items $\{\underline{\ell}, \underline{\ell+ 1}, \ldots, \underline{\ell+K - 1}\}$. Let $S^{*, \ell} = (\underline{\ell}, \underline{\ell + 1}, \ldots, \underline{\ell + K - 1})$ be an optimal list under instance $\ell$ for $\ell\in \{0, \ldots, L\}$. Evidently, it holds that $r(S^{*,\ell} | w^{(\ell)}) = 1 - [1 - (1 + \epsilon)/K]^K$ if $\ell \in \{1, \ldots, L\}$, and 
$r(S^{*,0} | w^{(0)}) = 1 - [1 - (1  -\epsilon)/K]^K$.
Note that $S^{*, 1}, \ldots, S^{*, L}$ `uniformly cover' the ground set $\{1, \ldots, L\}$, in the sense that each $i\in \{1, \ldots, L\}$ belongs to exactly $K$ sets among $S^{*, 1}, \ldots, S^{*, L}$. 
In Figure~\ref{cascadeTS_pic:instance_lb}, we give an example of $S^{*,1}, S^{*,2}, \ldots, S^{*,L}$ to illustrate how we construct these instances. 


{\bf Notations regarding to a generic online policy $\pi$.} A policy $\pi$ is said to be deterministic and non-anticipatory if  the list of items $S^\pi_t$ recommended by policy $\pi$ at time $t$ is completely determined by the observation history $\{S^\pi_s, O^\pi_s\}^{t-1}_{s=1}$. That is, $\pi$ is represented as a sequence of functions $\{\pi_t\}^{\infty}_{t=1}$, where $\pi_t : ([L]^{(K)} \times \{0, 1, \star\}^K)^{t-1} \rightarrow [L]^{(K)}$, and $S^\pi_t = \pi_t(S^\pi_1, O^\pi_1, \ldots, S^\pi_{t-1}, O^\pi_{t-1})$. We represent $O^\pi_t$ as a vector in $\{0, 1, \star\}^K$, where $0, 1, \star$ represents observing no click, observing a click and no observation respectively. For example, when $S^\pi_t = (2, 1, 5, 4)$ and $O^\pi_t = (0, 0, 1, \star)$, items $2,1,5,4$ are listed in the displayed order; items $2,1$ are not clicked, item 5 is clicked, and the response to item 4 is not observed. By the definition of the cascading model, the outcome  $O^\pi_t = (0, 0, 1, \star)$ is in general a (possibly empty) string of $0$s, followed by a $1$~(if the realized reward is $1$), and then followed by a possibly empty string of $\star$s.

%
%


Next, for any instance $\ell$, we lower bound the instantaneous regret of any arm $S$ with the number of suboptimal items within the arm: 

\begin{restatable}{lemma}{cascadeTSclaimrewarddiff}\label{cascadeTS_claim:rewarddiff}
    Let $\epsilon \in [0, 1]$, integer $K$ satisfy $0 <(1-\epsilon) / K < (1+\epsilon) / K < 1/4$. Consider instance $\ell$, where $1\leq \ell\leq L$. For any order $K$-set $S$, we have
$$
r(S^{*, \ell} | w^{(\ell)}) - r(S | w^{(\ell)}) \geq \frac{2  \left| S \setminus S^{*, \ell} \right|\epsilon}{e^{4}K}.
$$
\end{restatable}


\subsection*{Second step: Pinsker's inequality}
After constructing the family of instances, we consider a fixed but arbitrary non-anticipatory policy $\pi$, and analyze the stochastic process $\{S^\pi_t, O^\pi_t\}^T_{t=1}$. For instance $\ell \in \{0, \ldots, L\}$, we denote $P^{(\ell)}_{\{S^\pi_t, O^\pi_t\}^T_{t=1}}$ as the probability mass function of $\{S^\pi_t, O^\pi_t\}^T_{t=1}$ under instance $\ell$, and we denote $\mathbb{E}^{(\ell)}$ as the expectation operator (over random variables related to $\{S^\pi_t, O^\pi_t\}^T_{t=1}$) under instance $\ell$. Now,
Lemma~\ref{cascadeTS_claim:rewarddiff} directly leads to 
a fundamental
lower bound on the expected cumulative regret of a policy under an instance $\ell \in \{1, \ldots, L\}$:
\begin{equation}\label{cascadeTS_eq:intermediate_step_1}
\mathbb{E}^{(\ell)}\left[\sum^T_{t=1} R(S^{*, \ell} | w^{(\ell)}) - R(S^{\pi}_t| w^{(\ell)}) \right] \geq \frac{2\epsilon}{e^4 K} \cdot \sum_{j\in [L]\setminus S^{*, \ell}}  \mathbb{E}^{(\ell)} \left[ \sum^T_{t=1} \mathsf{1}(j \in S^{\pi}_t)\right].
\end{equation}
To proceed, we now turn to lower bounding the expectation of the number of times that suboptimal items appear in the chosen arms during the entire horizon. We apply Pinsker's inequality in this step.
\begin{restatable}{lemma}{cascadeTSclaimlbregrettoKL}\label{cascadeTS_claim:lbregrettoKL}
	For any non-anticipatory policy $\pi$, it holds that
	\begin{align}
	&\frac{1}{L} \sum^L_{\ell = 1} \sum_{j\in [L]\setminus S^{*, \ell}}	\mathbb{E}^{(\ell)} \left[  \sum^T_{t=1} \mathsf{1}(j\in S^{\pi}_t)\right] \nonumber\\
	& \qquad \geq  K T \left\{1  - \frac{K}{L} - \sqrt{\frac{1}{2  L}\sum^L_{\ell =1} \mathrm{KL}\left( P^{(0)}_{\{S^\pi_t, O^\pi_t\}^T_{t=1}}, P^{(\ell)}_{\{S^\pi_t, O^\pi_t\}^T_{t=1}} \right)}   \right\}.\nonumber
\end{align}     
\end{restatable}

\noindent Recall that under the null instance $0$, each item has click probabilities $ \frac{1 - \epsilon}{K}$, and the only difference between instance $0$ and instance $\ell~(1\le \ell \le L)$ is that the optimal items under instance $\ell$ have click probabilities $ \frac{1 - \epsilon}{K}$ under instance $0$ but $ \frac{1 + \epsilon}{K}$ under instance $\ell$. Since the difference between optimal and suboptimal items under instance $\ell$ is small, it is difficult to distinguish the optimal items from suboptimal ones. Therefore, the set of chosen arms and outcomes under instance $\ell$ would be similar to that under the null instance $0$. More precisely, the distributions of the trajectory $\{ S^{\pi}_t, O^{\pi}_t \}^T_{t=1}$ and that of $\{S^{\pi, \ell}_t, O^{\pi, \ell}_t \}^T_{t=1}$ would be similar, which implies that the KL divergence is small and hence the lower bound on the regret would be large. This intuitively implies that we have constructed instances that are difficult to discriminate and hence maximizing the expected regret.

\subsection*{Third step: chain rule}
To lower bound the regret, it suffices to bound the   average of $\mathrm{KL}\left(\{ S^{\pi, 0}_t, O^{\pi, 0}_t \}^T_{t=1}, \{ S^{\pi, \ell}_t, O^{\pi, \ell}_t \}^T_{t=1}\right)
$ (over $\ell$) from above. 
Observe that the outcome $O^{\pi, \ell}_t$ in each time step $t$ must be one of the following $K+1$ possibilities:
$$o^0 = (1, \star, \ldots, \star), o^1 = (0, 1, \star, \ldots, \star), \ldots, o^{K-1} = (0, \ldots, 0, 1) \text{ ,and } o^K = (0, \ldots, 0).$$ 
We index each possible outcome by the number of zeros (no-clicks), which uniquely defines each of the possible outcomes, and denote ${\cal O} = \{o^k\}_{k=0}^K$. The set of all possible realization of each $(S^{\pi, \ell}_t, O^{\pi, \ell}_t)$ is simply $[L]^{(K)} \times {\cal O}$.

With these notations, we decompose the KL divergence according to how time progresses. 
In detail, we apply the chain rule for the KL divergence iteratively~\citep{cover2012elements} to decompose $\mathrm{KL}\left(\{ S^{\pi, 0}_t, O^{\pi, 0}_t \}^T_{t=1}, \{ S^{\pi, \ell}_t, O^{\pi, \ell}_t \}^T_{t=1}\right)$ into per-time-step KL divergences.
\begin{restatable}{lemma}{cascadeTSclaimKLdecomp}\label{cascadeTS_claim:KLdecomp}
	For any instance $\ell$, where $1\le \ell \le L$,
\begin{align}
	&
	\mathrm{KL}\left( P^{(0)}_{\{S^\pi_t, O^\pi_t\}^T_{t=1}}, P^{(\ell)}_{\{S^\pi_t, O^\pi_t\}^T_{t=1}} \right)
	\nonumber \\ &
	= \sum_{t=1}^T \sum_{s_t\in [L]^{(K)}}P^{(0)}[S^{\pi}_t = s_t]\cdot \mathrm{KL}\left( P^{(0)}_{ O_t^{\pi}  \mid S_t^{\pi}}(\cdot \mid s_t)  \,\Big\|\,  P^{(\ell)}_{ O_t^{\pi }  \mid S_t^{\pi}}(\cdot \mid s_t)  \right).\nonumber
\end{align}
\end{restatable}
	
\subsection*{Fourth step: conclusion}
The design of instances $1$ to $L$ implies that the optimal lists cover the whole ground set uniformly, and an arbitrary item in the ground set belongs to exactly $K$ of the $L$ optimal lists. 
This allows us to further upper bound the average of the KL divergence $\mathrm{KL}\left( P^{(0)}_{ O_t^{\pi,0 }  \mid S_t^{\pi,0}}(\cdot \mid s_t)  \,\Big\|\,  P^{(\ell)}_{ O_t^{\pi, \ell }  \mid S_t^{\pi,\ell}}(\cdot \mid s_t)  \right)$ as follows:
\begin{restatable}{lemma}{cascadeTSclaimKLupperbd}\label{cascadeTS_claim:KLupperbound}
	For any $t \in [T]$,
	\begin{align*}
	    & \frac{1}{ L}\sum^L_{\ell =1} \sum_{s_t\in [L]^{(K)}} P^{(0)} [S^{\pi}_t = s_t]\cdot \mathrm{KL}\left( P^{(0)}_{ O_t^{\pi}  \mid S_t^{\pi}}(\cdot \mid s_t)  \,\Big\|\,  P^{(\ell)}_{ O_t^{\pi }  \mid S_t^{\pi}}(\cdot \mid s_t)  \right) \\
	    & \le \frac{K}{L} \left[ (1-\epsilon) \log \left( \frac{ 1-\epsilon }{ 1+\epsilon } \right) + (K-1+\epsilon) \log \left(  \frac{K-1+\epsilon}{K-1-\epsilon} \right) \right]. 
	\end{align*}
\end{restatable}
	
Insert the results of Lemma~\ref{cascadeTS_claim:KLdecomp} and Lemma~\ref{cascadeTS_claim:KLupperbound} into inequality~\eqref{cascadeTS_eq:after_jens}, we have
\begin{align}
	    &\frac{1}{L}\sum^L_{\ell=1}\mathbb{E}\left[\sum^T_{t=1}R(S^{*, \ell} | w^{(\ell)}) - R(S^{\pi, \ell}_t | w^{(\ell)}) \right] \nonumber \\ 
	    &
	    \ge \frac{2\epsilon T}{e^4 } \left[\left(1  - \frac{K}{L} \right)- \sqrt{ \frac{1}{2  L} \sum_{t=1}^T \sum^L_{\ell =1} \sum_{s_t\in [L]^{(K)}} P^{(0)} [S^{\pi}_t = s_t]\cdot \mathrm{KL}\left( P^{(0)}_{ O_t^{\pi}  \mid S_t^{\pi}}(\cdot \mid s_t)  \,\Big\|\,  P^{(\ell)}_{ O_t^{\pi }  \mid S_t^{\pi}}(\cdot \mid s_t)  \right) } \  \right]
	    \nonumber \\
	    & = \Omega \left( \epsilon T \left( 1  - \frac{K}{L} - \epsilon\sqrt{ \frac{TK}{2L} } \right) \right)= \Omega( \sqrt{LT/K} ).
	    \nonumber
\end{align}
The first step in the last line follows from the fact that 
\begin{align*}
     (1-\epsilon) \log \left( \frac{ 1-\epsilon }{ 1+\epsilon } \right) + (K-1+\epsilon) \log \left(  \frac{K-1+\epsilon}{K-1-\epsilon} \right)  = O(\epsilon^2)
\end{align*}
when $\epsilon$ is a small positive number, and the second step follows from differentiation. Altogether, Theorem~\ref{cascadeTS_thm:main_lb} is proved.

\section{Numerical experiments}
\label{cascadeTS_sec:exp_result}

We evaluate the performance of our proposed algorithms {\sc TS-Cascade}, {\algCasLin}, analyzed algorithm {\sc CTS} \citep{pmlr-v89-huyuk19a},  as well as algorithms in some existing works  such as 
   {\sc CascadeUCB1}, {\sc CascadeKL-UCB} in \cite{kveton2015cascading} and 
    {\sc CascadeLinUCB}, {\sc CascadeLinTS} , {\sc RankedLinTS}
  in \cite{ZongNSNWK16}.
The purpose of this section is not to show that the proposed algorithms---which have strong theoretical guarantees---are  uniformly the best possible (for all choices of the parameters), but rather to highlight some features of each algorithm and in which regime(s) which algorithm performs well. We believe this empirical study will be useful for practitioners in the future. 

\subsection{Comparison of Performances of  {\sc TS-Cascade} and {\sc CTS} to UCB-based Algorithms for Cascading Bandits}
\label{cascadeTS_sec:exp_nonlinear}

To demonstrate the effectiveness of the TS algorithms {\sc TS-Cascade} and {\sc CTS}, we compare their expected cumulative regrets  to {\sc CascadeKL-UCB} and {\sc CascadeUCB1} \citep{kveton2015cascading}. 
Since there are many numerical evaluations of these algorithms in the existing literature~\citep{cheung2019thompson,pmlr-v89-huyuk19a}, here we only present a brief discussion, highlighting some salient features.

     \begin{figure}[ht]
        \centering
        \begin{tabular}{cc}
        \hspace{-.5em}
        \makebox[ .48\textwidth]{ $L=256,K=2$ } 
        \makebox[ .48\textwidth]{ $L=256,K=4$ }\\
        \includegraphics[width = .48\textwidth]{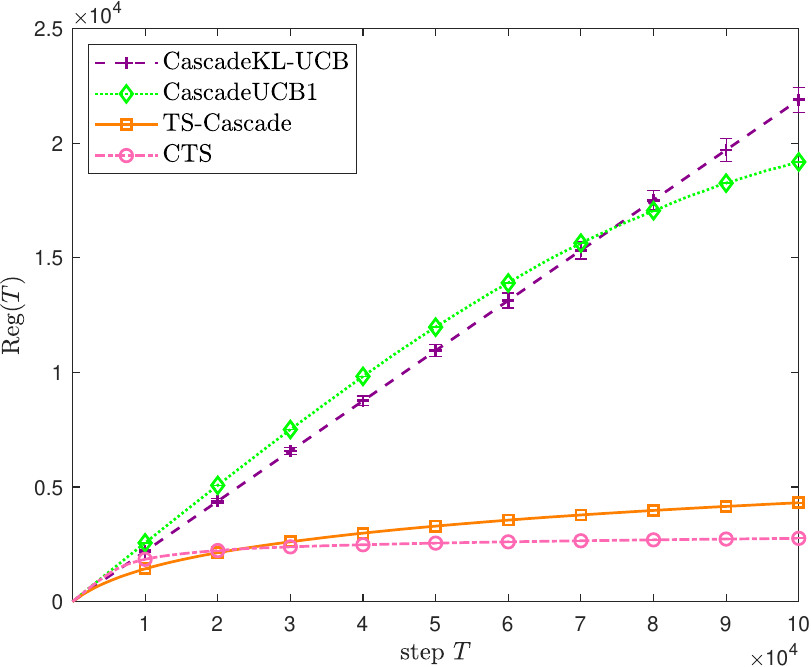} 
        \includegraphics[width = .48\textwidth]{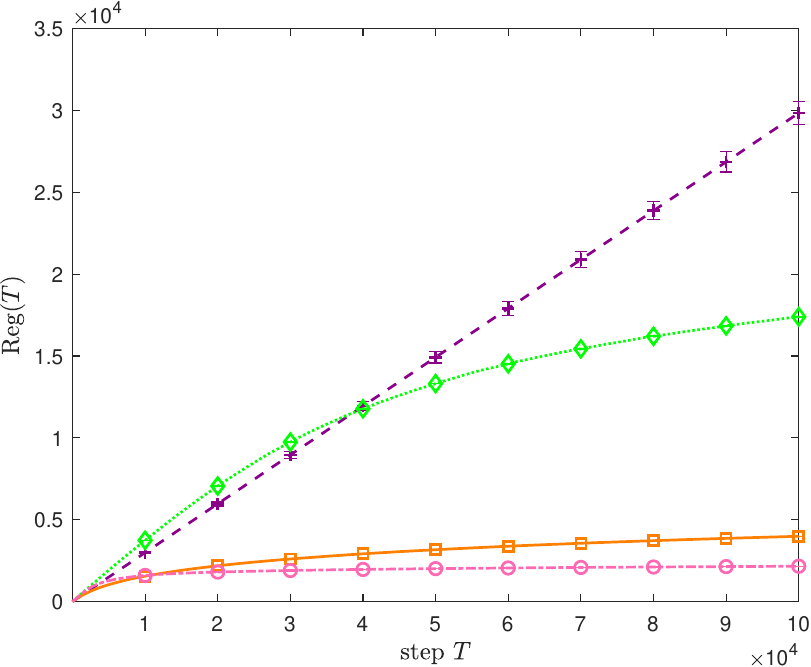} \\
        \makebox[ .48\textwidth]{ $L=2048,K=2$ } 
        \makebox[ .48\textwidth]{ $L=2048,K=4$ }\\
        \includegraphics[width = .48\textwidth]{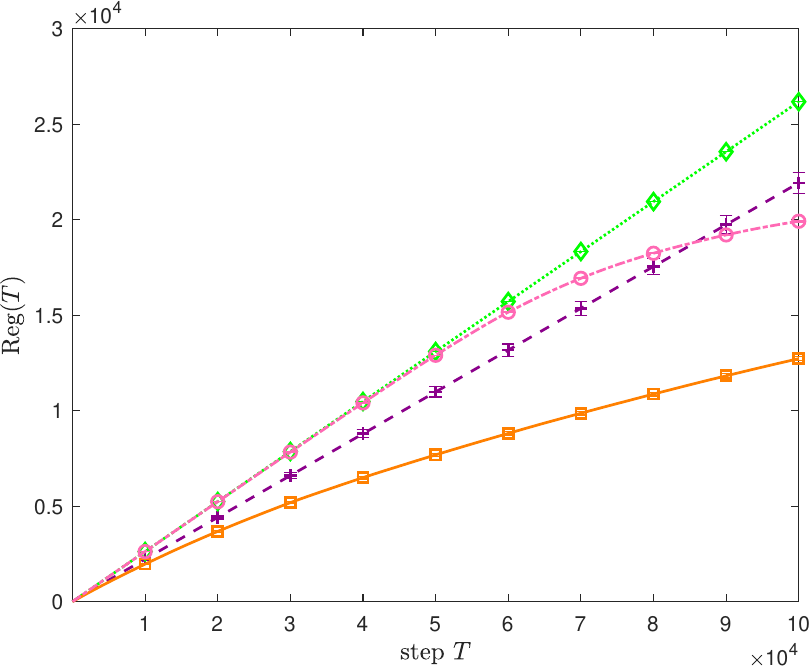} 
        \includegraphics[width = .48\textwidth]{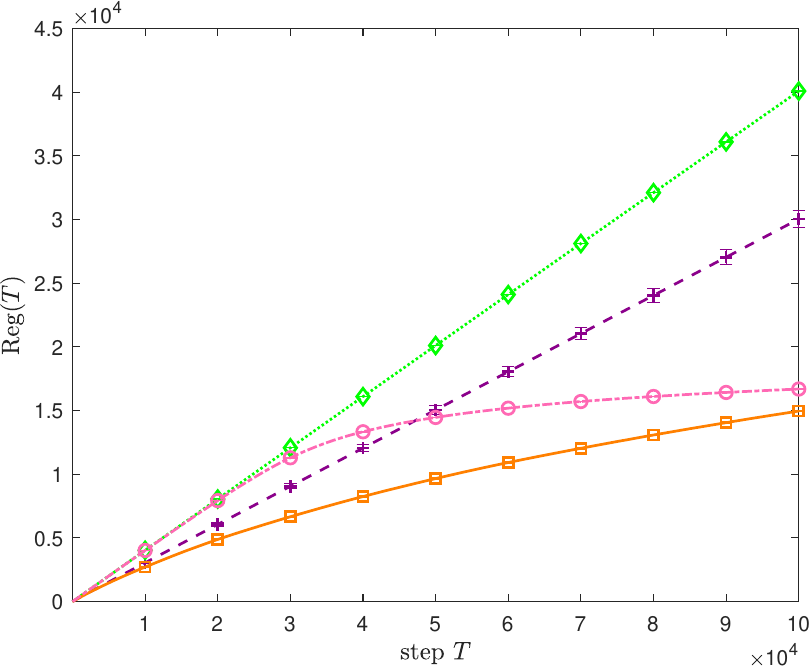} 
       \end{tabular}\vspace{-.1in}
        \caption{ $\mathrm{Reg}(T) $ of algorithms when $L\in\{ 256,2048\}$, $K\in\{ 2, 4 \}$. Each line indicates the average $\mathrm{Reg}(T)$ (over $20$ runs).}
        \label{cascadeTS_pic:nonlin_toy_probSet5}
    \end{figure}

 For the $L$ items, we set the click probabilities of $K$ of them to be $w_1=0.2$, the probabilities of another $K$ of them to be $w_2=0.1$, and the probabilities of the remaining $L-2K$ items as $w_3=0.05$. 
We vary $L\in \{ 16, 32, 64, 256, 512, 1024, 2048 \}$ and $K \in \{2 ,4 \}$. 
 We conduct $20$ independent simulations with each algorithm under each setting of $L$, $K$.
 We calculate the averages and standard deviations of $\mathrm{Reg}(T)$, and as well as the average running times of each experiment. 
 

Our experiments indicate that the two TS algorithms always outperform the two UCB algorithms in all settings considered (see Table~\ref{cascadeTS_tab:nonlin_toy_probSet5} in Appendix~\ref{cascadeTS_appdix:add_exp}) and Figure~\ref{cascadeTS_pic:nonlin_toy_probSet5}. 
Next, {\sc CascadeKL-UCB} outperforms {\sc CascadeUCB1} when $L \ge 512 $ while {\sc CascadeUCB1} outperforms {\sc CascadeKL-UCB} otherwise.
Similarly, {\sc TS-Cascade} outperforms {\sc CTS} when $L \ge 2048 $ while {\sc CTS} outperforms {\sc TS-Cascade} otherwise. Thus, for a large number of items,  {\sc TS-Cascade} is preferred to {\sc CTS}, again demonstrating the utility of the Gaussian updates over the seemingly more natural Bernoulli updates.
For simplicity,  we only present a representative subset of the results in Figure~\ref{cascadeTS_pic:nonlin_toy_probSet5}.
These observations imply that the best choice of algorithm depends on $L$ the total number of items in the ground set. 

\subsection{Performance of  {\algCasLin} compared to other algorithms for linear cascading bandits}
\label{cascadeTS_sec:exp_cas_reg_min_linear}

In this subsection, we compare 
 Algorithm~\ref{cascadeTS_alg:casLinTS'}  ({\algCasLin}) to other competitors using numerical simulations.
Note that the three competitors for cascading bandits with linear generalization {\sc CascadeLinUCB}, {\sc CascadeLinTS}, {\sc RankedLinTS}~\citep{ZongNSNWK16} involve tuning parameters (called $\sigma$). Thus to be fair to each algorithm,  we also tune the parameters for each of the algorithms to ensure that the best performances are obtained so that the comparisons are fair.  



We now describe how we set up these experiments. In real-world applications, good features that are extracted based on a large number of items are rarely obtained directly but rather learnt from historical data of users' behaviors~\citep{koren2009matrix}. 
One commonly used method to obtain the features of items is via low-rank matrix factorization.
In detail, let $\Wfeedback_{\text{train}} \in \{0, 1\}^{m\times L}$ be the   matrix representing feedback of $L$ items from $m$ users. The $(i,j)^{\mathrm{th}}$ element of feedback matrix $\Wfeedback_{\text{train}} (i,j)=1$ when user $i\in [m]$ is attracted to item $j \in [L]$, and $\Wfeedback_{\text{train}} (i,j)=0$ when user $i$ is not attracted to item $j$.  We apply Algorithm~\ref{cascadeTS_alg:generateFeature} to learn the features of the items from $\Wfeedback_{\text{train}}$.
 
    \begin{algorithm}[ht]
		\caption{ Generate feature matrix with historical data \citep{ZongNSNWK16} }\label{cascadeTS_alg:generateFeature}
		\begin{algorithmic}[1]
			\State Input historical data matrix $\Wfeedback_{\text{train}} \in \{0,1\}^{m\times L}$, feature length $d$.
			\State Conduct rank-$d$ truncated SVD of $\Wfeedback_{\text{train}}$: $\Wfeedback_{\text{train}} \approx U_{m\times d} S_{d\times d} V_{L\times d}^T$.
			\State Calculate the features of items $X_{d\times L}=SV^T$ to be used as an input in Algorithm \ref{cascadeTS_alg:casLinTS'}. 
		\end{algorithmic}
	\end{algorithm}

 
Our experiments reveal that, somewhat unsurprisingly,  the amount of training data has a noticeable impact on the performances of all algorithms assuming the linear bandit model. 
In detail, when the amount of training data $m$ is small, the feature matrix $X_{d\times L}$ contains a small amount of noisy information of the latent click probabilities, and thus taking into account the paucity of information, any algorithm assuming the linear bandit model cannot reliably identify the optimal items and hence suffers from a large (and often linear) regret. 
Conversely, when $m$ is large, $A_{\text{train}}$ contains a significant amount of information about the latent click probabilities $\{w(i)\}_{i=1}^L$. Thus, Algorithm~\ref{cascadeTS_alg:generateFeature} is likely to successfully project the unknown feature vectors $\{x(i)\}_{i=1}^L \subset \bbR^d$ onto an informative low-dimensional subspace, and to learn a linear separator in this subspace to classify the optimal and suboptimal items. This enables the algorithms assuming the linear bandit model with a small $d$ to quickly {\em commit} to the optimal set $S^*$ after small amount of exploration. 
However, in real-world applications, since the historical data is limited, we may not get sufficient training data but may require the algorithm itself to explore the items. 
Hence, for this problem setting, to illustrate the typical behavior of the algorithms under consideration, we set $m=200$ so that the amount of information contained in $A_{\text{train}}$ is ``appropriate''. 
However, obviously, other choices of $m$ will lead to other experimental outcomes, but since our objective is to compare the algorithms for the linear bandit model and not to scrutinize their performances as functions of $m$, we do not investigate the effect of $m$ any further.

To begin with, we generate independent Bernoulli random variables with   click probabilities $w_1$, $w_2$ or $w_3$ for each item to produce the matrix $A_{\text{train}}\in \{0,1\}^{m\times L}$ in which $m=200$. 
Next,  we vary the tuning parameters for each algorithm under consideration on a logarithmic scale.  We present the best performance of each algorithm for cascading bandits with linear generalization with tuning parameters in the finite set $\{ 0.01, 0.1, 1, 10, 100 \}$.

\begin{table}[t]
\setlength\extrarowheight{2pt}
  \centering
  \caption{The performances of  {\sc CascadeUCB1}, {\sc CascadeLinTS($100$)},  {\sc CascadeKL-UCB}, {\sc CTS}, {\sc TS-Cascade}, {\sc RankedLinTS($0.01$)}, { \sc CascadeLinUCB($0.1$)} and {\sc LinTS-Cascade($0.01$)} when $L=2048$, $K=4$.  The first column displays the mean and the standard deviation of $\mathrm{Reg}(T)$ and the second column displays the average running time in seconds.}
  \vspace{.5em}
    \begin{tabular}{llc}
           & $\mathrm{Reg}(T)$ & Running time \\
    \thickhline
    {\sc CascadeUCB1}  & $4.01 \times 10^4 \pm 4.18 \times 10^1$ & $5.57 \times 10^1 $ \\
    \hline
     {\sc CascadeLinTS($100$)} & $2.83 \times 10^4 \pm 2.06 \times 10^3$ & $2.36 \times 10^1 $ \\
    \hline
    {\sc CascadeKL-UCB} & $3.00 \times 10^4 \pm 1.00 \times 10^4$ & $5.47 \times 10^1 $ \\
    \hline
    {\sc \bf CTS} & $\bf 1.67 \times 10^4 \pm 3.99 \times 10^2$ & $\bf 1.25 \times 10^2 $ \\
    \hhline{===}
     {\sc \bf TS-Cascade} & $\bf 1.50 \times 10^4 \pm 7.27 \times 10^2$ & $ \bf 4.55 \times 10^1 $ \\
    \hline
     {\sc RankedLinTS($0.01$)} & $1.32 \times 10^4 \pm 5.03 \times 10^3$ & $2.64 \times 10^2 $ \\
    \hline
      {\sc CascadeLinUCB($0.1$)} & $7.71 \times 10^1 \pm 2.57 \times 10^2$ & $4.19 \times 10^3 $ \\
    \hline
     {\sc \bf LinTS-Cascade($0.01$)}   & $ \bf 7.12 \times 10^1 \pm 2.29 \times 10^2$ & $ \bf 5.08 \times 10^1 $ \\
    \end{tabular}%
  \label{cascadeTS_tab:lin_toy_probSet5}%
\end{table}%

\begin{figure}[t]
	\centering
	\begin{tabular}{cc}
    	\hspace{-.5em}
        \includegraphics[width = .45\textwidth]{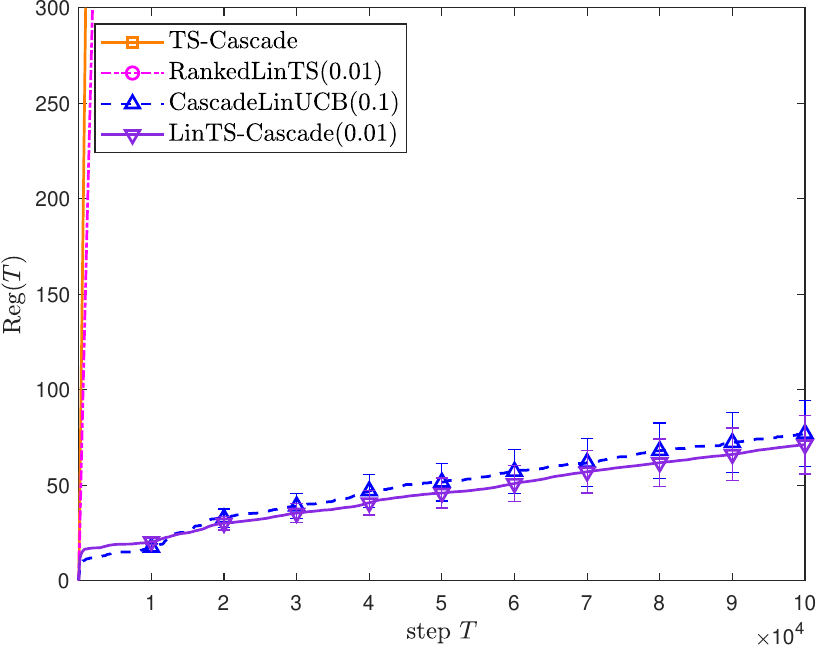} 
        \hspace{.5em}
        \includegraphics[width = .45\textwidth]{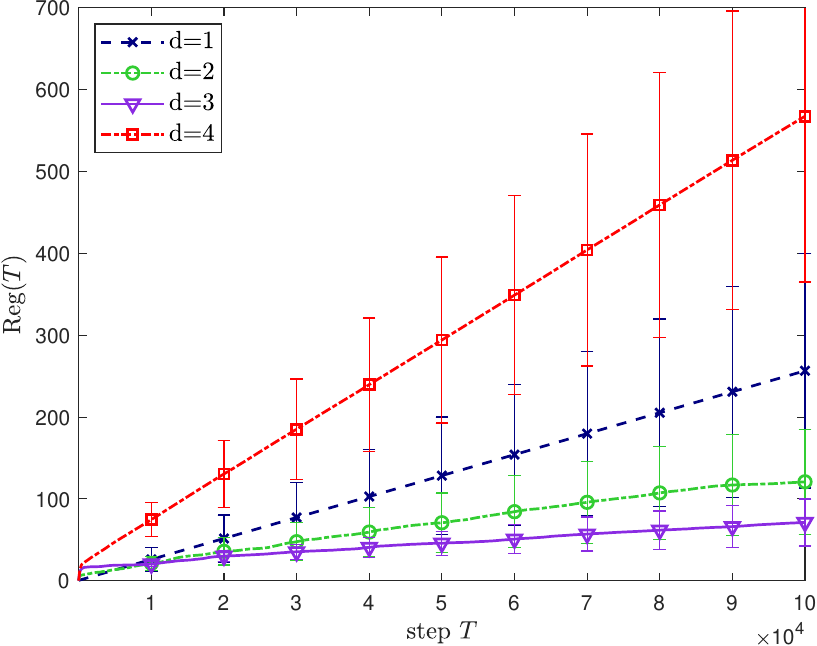} 
       \end{tabular}\vspace{-.1in}
        \caption{Left: $\mathrm{Reg}(T) $ of algorithms when $L=1024$, $K=2$, $d=3$; Right:  $\mathrm{Reg}(T) $ of  {\sc LinTS-Cascade($0.04$)} when $L=1024$, $K=2$, $d\in \{1,2,3,4\}$. Each line indicates the average $\mathrm{Reg}(T)$ (over $20$ runs), and  the length of each errorbar above and below each data point is the scaled standard deviation.}
        \label{cascadeTS_pic:lin_toy_probSet5}
    \end{figure}


Table~\ref{cascadeTS_tab:lin_toy_probSet5} displays the means and the standard deviations of $\mathrm{Reg}(T)$ as well as the average running times. 
The algorithms are listed in descending order of the average cumulative regrets.
 Note that we set $d=3$ for the algorithms based on the linear bandit model.   We will explain this choice later.

%

In the left  plot of Figure~\ref{cascadeTS_pic:lin_toy_probSet5}, we present the average cumulative regrets of the algorithms. Since {\sc CascadeUCB1},  {\sc CascadeKL-UCB}, {\sc CTS},  {\sc CascadeLinTS($100$)} incur much larger regrets compared to the remaining algorithms, we omit their regrets in the plot for clarity. 
For the algorithms assuming the linear bandit model and for this dataset, our experiments indicate that $d=3$ is the most appropriate choice.
 For clarity, we only display the regrets of the proposed {\algCasLin} algorithm as $d$ varies from $1$ to $4$ in the right plot of Figure~\ref{cascadeTS_pic:lin_toy_probSet5}. Figure~\ref{cascadeTS_pic:lin_toy_probSet5}  is a representative plot for all considered algorithms based on the linear bandits.  It can be seen that $d=3$ results in the best performance in terms of the expected cumulative regret, %
 i.e., $d = 3$ minimizes the average regret among the values of $d$ considered here. %
Roughly speaking, this implies that the optimal and suboptimal items can be separated when the data is projected onto a three-dimensional subspace (compared to subspaces of other dimensions). 
When $d>3$, the matrix that contains the extracted features  $X_{d\times L}$  is overfits to the historical data and due to this overfitting, our algorithm---as well as other algorithms---incurs a larger regret. Hence, in our comparison to other algorithms for linear bandits, we only consider $d=3$ features. 

Among the algorithms designed for   linear bandits, {\sc CascadeLinUCB} requires significantly more time to run (as compared to other algorithms) as suggested by  Table~\ref{cascadeTS_tab:lin_toy_probSet5} while {\sc RankedLinTS} lacks theoretical analysis. Our proposed {\algCasLin} algorithm is the {\em only} efficient algorithm with {\em theoretical guarantees}.

For {\algCasLin}, it is essential to pick an appropriate $\lambda$ to balance   exploration and exploitation. Varying $\lambda$ is equivalent to  tuning the variance of the Thompson samples~(see Line~5 of Algorithm~\ref{cascadeTS_alg:casLinTS'}). 
We note that other algorithms such as {\sc CascadeLinUCB} and {\sc CascadeLinTS}
~\citep{ZongNSNWK16} also contain tuning parameters (called $\sigma$), and thus our algorithm is similar to these other competing algorithms
 in this aspect. However, in Figure~\ref{cascadeTS_pic:lin_toy_probSet5}, we have displayed results in which the hyperparameters of all algorithms are tuned on a logarithmic scale to be fair.
 

Lastly, we submit that our {\algCasLin} algorithm does not outperform other algorithms under all problem settings. Indeed, the performances of all algorithms based on the linear bandit model are sensitive to the amount of training data $A_{\text{train}}\in \{0,1\}^{m\times L}$ (i.e., the number of rows $m$) and the choice of $d$. However, it is worth emphasizing that our proposed algorithm  is computationally efficient and possesses appealing (asymptotic) {\em theoretical guarantees}; cf.~Theorem~\ref{cascadeTS_thm:mainLin}. Indeed, it is the only Thompson sampling algorithm for linear cascading bandits that admits such theoretical guarantees.

\section{Summary  and Future work}\vspace{-.1in}
\label{cascadeTS_sec:summary_cas_reg_min}
This work presents a comprehensive theoretical analysis of various Thompson sampling algorithms for cascading bandits under the standard and linear models. %
First, the upper bound on the expected regret of {\sc TS-Cascade} we derived in Theorem~\ref{cascadeTS_thm:main} matches the state-of-the-art bound based on UCB by~\cite{wang2017improving} (which is identical to {\sc CascadeUCB1} in \cite{kveton2015cascading}). %
Secondly, we provide a tighter upper bound on the regret of {\sc CTS} compared to \cite{pmlr-v89-huyuk19a}. %
Thirdly, we propose {\algCasLin} for the linear generalization of the problem and upper bound its regret in Theorem \ref{cascadeTS_thm:mainLin} by refining proof techniques used for {\sc TS-Cascade}. 
  Finally, in Theorem \ref{cascadeTS_thm:main_lb}, we also derive a problem-independent lower bound on the expected cumulative regret by analyzing, using the information-theoretic technique of \cite{Auer02} as well as  a judiciously constructed of bandit instance. 
Lastly, %
we also demonstrate the computational efficiency and efficacy in terms of regret of {\algCasLin}.

In Section~\ref{cascadeTS_alg:ts_beta}, we analyze the Thompson sampling algorithm with Beta-Bernoulli update and provide a tighter bound than \cite{pmlr-v89-huyuk19a} in the cascading bandit setting. Our analysis can be readily generalized to the matroid bandit setting which is also widely considered in current literature as in \cite{KvetonWAEE14}. 
Next, since the attractiveness of items in real life may change with time, it is reasonable to consider the cascading bandit model with unknown change points of latent  click probabilities~\citep{garivier2011upper}. There are practical real-life implications in designing and analyzing algorithms that take into account the non-stationarity of the latent click probabilities, e.g., the attractiveness of hotels in a city depending on the season. 
Further, another interesting topic within the realm of multi-armed bandit problems concerns best arm identification~\citep{audibert2010best}. In the cascading model since an {\em arm} is a {\em list of items}, the objective here is to minimize the total number of pulls of all arms to identify items that are ``close to'' or exactly equal to the optimal ones with high probability. In the future, we plan to design and analyze an algorithm for this purpose assuming the cascading bandit model.




\vskip 0.2in
\bibliography{cascaderef}

\newpage

\appendix

\section{Notations}

\begin{spacing}{1.5}
	\begin{longtable}[!h]{ p{.32\textwidth}  p{.68\textwidth} } 
		   $[L]$ & ground set of size $L$ \\
		   $w(i)$ & click probability of item $i \in [L]$ \\
		   $x(i)$ & feature vector of item $i $ in the linear generalization \\
		   $\beta$ & common latent vector in the linear generalization \\
		   $T$ & time horizon \\
		   $d$ & number of features of any item in the linear generalization\\
		   $K$ & size of recommendation list \\
		   $S_t$ & recommendation list at time $t$\\
		   $i_i^t$ & $i$-th recommended item at time $t$\\
		   $[L]^{(K)}$ & set of all $K$-permutations of $[L]$ \\
		   $W_t(i)$ & reflection of the interest of the user in item $t$ at time $t$\\
		   $R(S_t| {\bm w } )$ & instantaneous reward of the agent at time $t$\\
		   $k_t$ & feedback from the user at time $t$ \\
		   $r(S_t| {\bm w } )$ & expected instantaneous reward of the agent at time $t$\\
		   $S^{ *}$ & optimal ordered $K$-subset for maximizing the expected instantaneous reward  \\
		   $\mathrm{Reg}(T)$ & expected cumulative regret\\
		   $\hat{w}_t(i)$ & empirical mean of item $i$ at time $t$  \\
		   $N_t(i)$ & number of observations of item $i$ before time $t$ plus $1$ \\
		   $Z_t$ & $1$-dimensional random variable drawn from the standard normal distribution at time $t$ \\
		   $\hat{\nu}_t(i)$ & empirical variance of item $i$ at time $t$\\
		   $\sigma_t(i)$ & standard deviation of Thompson sample of item $i$ at time $t$ in {\sc TS-Cascade}\\
		   $\theta_t(i)$ & Thompson sample of item $i$ at time $t$ in {\sc TS-Cascade}\\
		   $\mathcal{E}_{\hat{w},t}$, $\mathcal{E}_{\theta,t}$ & ``nice events'' in {\sc TS-Cascade} \\
		   $ g_t(i)$, $h_t(i) $ & functions used to define ``nice events'' \\
		   $\mathcal{S}_t$ & ``unsaturated" set at time $t$ \\
		   $F(S,t)$ & function used to define ``unsaturated" set \\
		   $\alpha$ & determined by click probabilities $w(i)$'s, used in the analysis \\
		   $\mathcal{H}_{\hat{w},t}$, $\mathcal{F}_{i,t}$, $G(S_t, \mathbf{W}_t ) $ & used in the analysis of {\sc TS-Cascade} \\
           $\alphaAlgBeta_t(i), \betaAlgBeta_t(i)$ & statistics used in {\sc CTS} \\
           $\thetaAlgBeta_t(i)$ & Thompson sample from Beta distribution at time $t$ in {\sc CTS} \\
           $\Delta_{i,j}$ & gap between click probabilities of optimal item $i$ and suboptimal item $j$ \\
           $\Delta=\Delta_{K,K+1}$ & minimum gap between click probabilities of optimal and suboptimal items \\
           $\pi_t$ & a permutation of optimal items used in the analysis of {\sc CTS} \\
           ${\mathcal{A}}_t( \ell,j,k)$, $\tilde{\mathcal{A}}_t( \ell,j,k)$,  ${\mathcal{A}}_t( \ell,j)$ & events defined to help decompose the regret of {\sc CTS} \\
            $\numAlgBeta(i,j)$ & number of time steps that $\calA_t(i, j)$ happens for optimal item $j$ and suboptimal item $i$ within the whole horizon \\
           $\mathcal{B}_t(i), \mathcal{C}_t(i,j)$ & events that measure the concentration of $\thetaAlgBeta_t(i)$, $\hat{w}_t(i)$ around $w(i)$ as  {\sc CTS} processes \\
           $\thetaAlgBeta(-i), W_{i,j},W_{i,-j}$ & variables used in the analysis of {\sc CTS}  \\
		   $\lambda$ & parameter in {\algCasLin}  \\
		   $X$ & collection of feature vector $x(i)$'s \\
		   $\bLin_t$, $\NLin_t$, $v_t$ & statistics used in {\algCasLin} \\
		   $\ZLin_t$ & $d$-dimensional random variable drawn from standard multivariate normal distribution at time $t$ in {\algCasLin} \\
		   $\hat{\muLin}_t$ & empirical mean of latent vector $\beta$ at time $t$\\
		   $\thetaLin_t$ & Thompson sample at time $t$ in {\sc TS-Cascade}\\
		   $\eta_t(i)$ & random variable in the linear generalization\\
		   $\mathcal{E}_{\hat{\muLin},t}$, $\mathcal{E}_{\thetaLin,t}$ & ``nice events'' in {\algCasLin} \\
		   $l_t$, $u_t(i) $ & functions used to define ``nice events''  in {\algCasLin} \\
		   $w^{ (\ell) }(i)$ & click probability of item $i \in [L]$ under instance $\ell~(0\le \ell \le L)$ \\
		   $r(S_t| {\bm w^{ (\ell) } } )$ & expected instantaneous reward of the agent at time $t$ under instance $\ell$ \\
		   $S^{ *, \ell }$ & optimal ordered $K$-subset under instance $\ell$ \\
		   $\pi$ & deterministic and non-anticipatory policy \\
		   $S_t^{\pi, (\ell) }$ & recommendation list at time $t$ (under instance $\ell$) by policy $\pi$ \\
		   $O_t^{\pi, (\ell) }$ & stochastic outcome by pulling $S_t^{\pi, (\ell) }$ at time $t$ (under instance $\ell$) by policy $\pi$ \\
		   $A$ & feedback matrix extracted from real data\\
		   $y_t$, $Q_t$, $\zeta_t$, $\stdLin_t$ & variables used in the analysis of {\algCasLin} \\
		   $Q$, $J_t^{\pi, \ell}$, $\Upsilon_t^\ell(S)$ & variables used in the analysis of the lower bound
	\end{longtable}
\end{spacing}
	\addtocounter{table}{-1}

\section{Useful theorems}
    Here are some basic facts from the literature that we will use:
\begin{theorem}[\cite{AudibertMS09}, speicalized to Berounlli random variables]\label{cascadeTS_thm:emp_berstein}
    Consider $N$ independently and identically distributed Bernoulli random variables $Y_1, \ldots, Y_N\in \{0, 1\}$, which have the common mean $m = \mathbb{E}[Y_1]$. In addition, consider their sample mean $\hat{\xi}$ and their sample variance $\hat{V}$: 
    $$\hat{\xi} = \frac{1}{N}\sum^{N}_{i=1}Y_i\text{, }\quad \hat{V} = \frac{1}{N}\sum^{N}_{i=1}(Y_i - \hat{\xi})^2 = \hat{\xi}(1 - \hat{\xi}).$$  For any $\delta \in (0, 1)$, the following inequality holds:
        $$
            \Pr\left(\left|\hat{\xi} - m \right| \leq \sqrt{\frac{2\hat{V}\log(1/\delta)}{N}} + \frac{3\log(1/\delta)}{N}\right)\geq 1 - 3\delta.
        $$
\end{theorem}

\begin{theorem}[\cite{AbramowitzS64}, Formula 7.1.13; \cite{AgrawalG13b}]  
\label{cascadeTS_thm:gaussian}
	
    Let $Z\sim {\cal N}(\mu, \sigma^2)$. For any $z\geq 1$, the following inequalities hold:
    \begin{align*}   
         \frac{1}{4\sqrt{\pi}}\exp\left(-\frac{7 z^2}{ 2}\right) & \leq \Pr\left(\left| Z - \mu\right| > z\sigma\right)\leq \frac{1}{2}\exp\left(-\frac{z^2}{2}\right)
         , \\
        \frac{1}{2\sqrt{\pi}z}\exp\left(-\frac{z^2}{ 2}\right) & \leq \Pr\left(\left| Z - \mu\right| > z\sigma\right)\leq \frac{1}{ \sqrt{\pi} z }\exp\left(-\frac{z^2}{2}\right).
    \end{align*}
\end{theorem}

\begin{theorem}\label{cascadeTS_thm:linear} 
    {\rm \citep{abbasi2011improved,AgrawalG13b}.}
    Let $(\calF_t'; t \ge 0)$ be a filtration, $(m_t; t \ge 1)$ be an $\bbR^d$-valued stochastic process such that $m_t$ is $(\calF'_{t-1})$-measurable, $(\eta_t; t \ge 1)$ be a real-valued martingale difference process such that $\eta_t$ is $(\calF'_t)$-measurable. For $t \ge 0$, define $\xi_t = \sum_{\tau=1}^t m_\tau \eta_\tau$ and $M_t = I_d + \sum_{\tau_1}^T m_\tau m_\tau^T$, where $I_d$ is the $d$-dimensional identity matrix. Assume $\eta_t$ is conditionally $R$-sub-Gaussian.
    
Then, for any  $\delta' > 0$, $t \ge 0$, with probability at least $1 - \delta'$,
    \begin{align*}
        \| \xi_t \|_{ M_t^{-1} } \le R \sqrt{d \log \big( \frac{t+1}{\delta'}  \big) },
    \end{align*}
where $\| \xi \|_{ M_t^{-1} } = \sqrt{\xi_t^T M_t^{-1} \xi_t }$.
\end{theorem}

\section{Proofs of main results}\label{cascadeTS_sec:proofs}
In this Section, we provide proofs of Lemmas%
~\ref{cascadeTS_lemma:beta_1st_term_part},~\ref{cascadeTS_lemma:beta_3rd_term_part},%
~\ref{cascadeTS_claim:conc},~\ref{cascadeTS_lemma:transferone},~\ref{cascadeTS_lemma:gap_bound},%
~\ref{cascadeTS_claim:concLin},~\ref{cascadeTS_lemma:transfertwoLin},~\ref{cascadeTS_lemma:zong_sysmmetry_reg_decomp},~\ref{cascadeTS_claim:rewarddiff},%
~\ref{cascadeTS_claim:lbregrettoKL},~\ref{cascadeTS_claim:KLdecomp},~\ref{cascadeTS_claim:KLupperbound} as well as deduction of Inequality%
~\eqref{cascadeTS_eq:beta_2nd_term} and~\eqref{cascadeTS_eq:bound_y_other}. 


\subsection{Proof of Lemma~\ref{cascadeTS_lemma:beta_1st_term_part} }
\label{cascadeTS_pf:lemma_beta_1st_term_part}


\begin{proof}
    We let $\tau_{1}, \tau_{2}, \ldots$ be the time slots such that $i \in S(t)$ and $W_t(i)$ is observed, define $\tau_{0}=0$, and abbreviate $\varepsilon_i$ as $\varepsilon$ for simplicity. Then
    \begin{align*} 
        &\mathbb{E}\left[~ \exists t : 1 \leq t \leq T, i \in S_t,| \hat{w}_{t}(i)-  w(i) | >\varepsilon  , \text{Observe $W_t(i)$ at $t$} ~\right]\\
        &=\mathbb{E}\left[\sum_{t=1}^{T} \mathsf{1}\left[i \in S_t,\left|\hat{w}_t(i)-w(i)\right|>\varepsilon, \text{Observe $W_t(i)$ at $t$}  \right] \right] \\ 
        & \leq \mathbb{E}\left[\sum_{s=0}^{T} \mathbb{E}\left[\sum_{t=\tau_{s}}^{\tau_{s+1}-1} \mathsf{1}\left[i \in S_t,\left|\hat{w}_t(i)-w(i)\right|>\varepsilon, \text{Observe $W_t(i)$ at $t$} \right]\right]\right] \\ 
        & \leq \mathbb{E}\left[\sum_{s=0}^{T} \operatorname{Pr}\left[\hat{w}_t(i)-w(i) |>\varepsilon, N_{t}(i) =s\right]\right]\\
        & \leq 1+\sum_{s=1}^{T} \operatorname{Pr}\left[\left|\hat{w}_t(i)-w(i)\right|>\varepsilon, N_{t}(i)=s \right] \\ 
        & \leq 1+\sum_{s=1}^{T} \exp \left[ -w \rmKL\left(  w(i)+\varepsilon , w(i) \right) \right]+\sum_{s=1}^{T} \exp \left[-w \rmKL\left(w(i)-\varepsilon , w(i)\right) \right] \\ 
        & \leq 1+\sum_{s=1}^{\infty}\left\{ \exp \left [ -\rmKL\left(w(i)+\varepsilon , w(i)\right) \right] \right)\}^{s}+\sum_{w=1}^{\infty}\left\{ \exp \left[ -\rmKL\left(w(i)-\varepsilon , w(i)\right) \right] \right\}^{s}\\
        & \leq 1+\frac{\exp \left[-\rmKL\left(w(i)+\varepsilon , w(i)~\right)\right] }{1-\exp \left[ -\rmKL\left(w(i)+\varepsilon , w(i)\right)  \right] }+\frac{\exp \left[ -\rmKL\left(w(i)-\varepsilon , w(i)\right)\right]  }{1-\exp \left[-\rmKL\left(w(i)-\varepsilon , w(i)\right)\right] } \\
        & \leq 1+\frac{1}{\rmKL\left(w(i)+\varepsilon,  w(i)\right)}+\frac{1}{\rmKL\left(w(i)-\varepsilon , w(i)\right)} \\
        & \leq 1+\frac{1}{\varepsilon^{2}}.
    \end{align*}
\end{proof}

\subsection{Upper bound on the regret incurred by the second term of~\eqref{cascadeTS_eq:beta_decomp_N_ij} in Section~\ref{cascadeTS_sec:beta_alg_cas} }  
\label{cascadeTS_pf:eq_beta_2nd_term}
To derive the inequality~\eqref{cascadeTS_eq:beta_2nd_term}: for any  $0<\varepsilon < \Delta_{K,K+1}/2$,
\begin{align*}
    & \sum_{i=K+1}^L \sum_{j=1}^K \Delta_{j,i} \cdot \mathbb{E}\left[\sum_{t=1}^{T} \mathsf{1}\left[\calA_t(i, j) \wedge \neg \mathcal{B}_{t}(i) \wedge \mathcal{C}_{t}(i,j) \right]\right]  
     \le 4\log T \cdot \! \sum_{i=K+1}^L \frac{ \Delta_{K,i} - \varepsilon }{ ( \Delta_{K,i} - 2\varepsilon )^{2} } 
     + \sum_{i=K+1}^L \sum_{j=1}^K \Delta_{j,i}, 
\end{align*}
%
%
%
we use the following result:
\begin{lemma}[Implied by~\cite{WangC18}, Lemma 4]
\label{cascadeTS_lemma:beta_conc}
	In Algorithm~\ref{cascadeTS_alg:ts_beta},for any base item $i$, we have the following two inequalities:
	\begin{align*}
		\operatorname{Pr}\left[\thetaAlgBeta_{i}(t)-\hat{w}_t(i)>\sqrt{\frac{2 \log T}{N_{t}(i)}}\right] \leq \frac{1}{T},\quad
		\operatorname{Pr}\left[\hat{w}_t(i)-\thetaAlgBeta_{i}(t)>\sqrt{\frac{2 \log T}{N_{t}(i)}}\right] \leq \frac{1}{T}.
	\end{align*}
\end{lemma}

Note that \begin{align*}
    \varepsilon_i = \frac{1}{2} \cdot \Big(  \Delta_{K+1,i} +  \frac{ \Delta_{K,K+1} }{2}  \Big)
    \text{ and }
    \varepsilon_j = \frac{1}{2} \cdot \Big(  \Delta_{j,K} +  \frac{ \Delta_{K,K+1} }{2}  \Big) 
\end{align*}
implies $\varepsilon_i + \varepsilon_j = \Delta_{j,i}/2$. Let
\begin{align*}
    M_{i,j} =  \frac{2\log T}{ ( \Delta_{j,i} - \varepsilon_i  -  \varepsilon_j )^2 } 
    =  \frac{ 8 \log T }{ \Delta_{j,i}^2 },
\end{align*}
then when $N_t(i) > M_{i,j}$, $\Delta_{j,i}  - \varepsilon_i  -  \varepsilon_j > \sqrt{\frac{2 \log T}{N_{t}(i)} }$. Thus in expectation, the time steps that $\calA_t(i, j) \wedge \neg \mathcal{B}_{t}(i) \wedge \mathcal{C}_{t}(i,j)$ occurs can be decomposed as follows:

\begin{align*}
    & \mathbb{E}\left[\sum_{t=1}^{T} \mathsf{1}\left[\calA_t(i, j) \wedge \neg \mathcal{B}_{t}(i) \wedge \mathcal{C}_{t}(i,j) \right]\right]  \\
    & = \mathbb{E}\left[\sum_{t=1}^{T} \mathsf{1}\left[\calA_t(i, j) \wedge \neg \mathcal{B}_{t}(i) \wedge \mathcal{C}_{t}(i,j), N_t(i) \le M_{i,j} \right]\right] 
    \\ & \hspace{10em}
    + \mathbb{E}\left[\sum_{t=1}^{T} \mathsf{1}\left[\calA_t(i, j) \wedge \neg \mathcal{B}_{t}(i) \wedge \mathcal{C}_{t}(i,j), N_t(i) > M_{i,j} \right]\right] .
\end{align*}
And further
\begin{align}
    & \mathbb{E}\left[\sum_{t=1}^{T} \mathsf{1}\left[\calA_t(i, j) \wedge \neg \mathcal{B}_{t}(i) \wedge \mathcal{C}_{t}(i,j), N_t(i) > M_{i,j} \right]\right] \nonumber \\
    & \le \mathbb{E}\left[\sum_{t=1}^{T} \mathsf{1}\left[  \hat{w}_t(i)\le w(i) +\varepsilon_i, \thetaAlgBeta_{t}(i)> w(j) - \varepsilon_j,  N_t(i) > M_{i,j}, \text{Observe $W_t(i)$ at $t$}  \right]\right] \nonumber \\
    & \le \mathbb{E}\left[\sum_{t=1}^{T} \mathsf{1}\left[  \thetaAlgBeta_t(i) > \hat{w}_t(i) + \Delta_{j,i} -  \varepsilon_i - \varepsilon_j , N_t(i) > M_{i,j}, \text{Observe $W_t(i)$ at $t$}  \right]\right] \label{cascadeTS_eq:beta_theta_mu_dist} \\
    %
%
    & \le \bbE \left[ \sum_{t=1}^T   \mathsf{1}\left[~  \thetaAlgBeta_t(i) > \hat{w}_t(i) +  \sqrt{\frac{2 \log T}{N_{t}(i)}} ~\right]  \right]  \label{cascadeTS_eq:beta_by_M_ij_def} \\
    & \le 1. \nonumber
\end{align}
To derive line~\eqref{cascadeTS_eq:beta_theta_mu_dist}, we observe that when $\hat{w}_t(i)\le w(i) +\varepsilon_i, \thetaAlgBeta_{t}(i)> w(j) - \varepsilon_j$, we have
\begin{align*}
    \thetaAlgBeta_t(i) > w(j) -\varepsilon_j = w(i) + \Delta_{j,i} - \varepsilon_j \ge \hat{w}_t(i) + \Delta_{j,i} - \varepsilon_i - \varepsilon_j.
\end{align*}
Besides, line~\eqref{cascadeTS_eq:beta_by_M_ij_def} follows from the definition of $M_{i,j}$.
 
Hence, 
\begin{align*}
    & \sum_{j=1}^K \Delta_{j,i} \cdot \mathbb{E}\left[\sum_{t=1}^{T} \mathsf{1}\left[\calA_t(i, j) \wedge \neg \mathcal{B}_{t}(i) \wedge \mathcal{C}_{t}(i,j) \right]\right]  
     \le 
     \underbrace{ \sum_{j=1}^K \Delta_{j,i} \cdot \mathbb{E}\left[\sum_{t=1}^{T} \mathsf{1}\left[\calA_t(i, j), N_t(i) \le M_{i,j} \right]\right]  }_{(*)}
    + \sum_{j=1}^K \Delta_{j,i}.
\end{align*}
Note that (i) $N_t(i)$ of suboptimal item $i$ increases by one when the event $\calA_{i,j,t}$ happens for any optimal item $j$, (ii) the event $\calA_t(i, j)$ happens for at most one optimal item $i$ at any time $t$; and (iii) $M_{1,i} \le \ldots \le M_{K,i}$. Let
\begin{align*}
	Y_{i,j} = \sum_{t=1}^T \mathsf{1}\left[~ \calA_t(i, j), N_t(i)\le M_{i,j} \right].
\end{align*}
Based on these facts, it follows that $Y_{i,j} \le  M_{i,j}$, and moreover $\sum_{j=1}^K Y_{i,j} \le M_{K,i}$. Therefore, ($*$) can be bounded from above by:
	\begin{align*}
		\max \left\{\sum_{j=1}^{K} \Delta_{j,i} y_{i,j}: 0 \leq y_{i,j} \le  M_{i,j}, \sum_{i=1}^K y_{i,j} \le M_{K,j} \right\}.
	\end{align*}
Since the gaps are decreasing, $\Delta_{1,i} \ge \Delta_{2,i} \ge \Delta_{K,i}$, the solution to the above problem is $y_{1,i}^* = M_{1,i}$, $y_{2,i}^* = M_{2,i} - M_{1,i}$, $\ldots$, $y_{K,i}^* = M_{K,i} - M_{K-1,i}$. Therefore, the value of ($*$) is bounded from above by:
\begin{align*}
	8 \log T \left[ \Delta_{1,i} \frac{1}{ \Delta_{1,i} ^{2} }+\sum_{j=2}^{K} \Delta_{j,i}\left(\frac{1}{  \Delta_{j,i}^{2} }-\frac{1}{  \Delta_{j-1,i}^{2} }\right) \right] .
\end{align*}
Further, 
we use the following result:
\begin{lemma}[\cite{KvetonWAEE14}, Lemma 3] \label{cascadeTS_lemma:prod_sum_bd}
	Let $ \Delta_{1} \geq \ldots \geq \Delta_{K} $ be a sequence of $K$ positive numbers, then
	$$
\left[\Delta_{1} \frac{1}{\Delta_{1}^{2}}+\sum_{k=2}^{K} \Delta_{k}\left(\frac{1}{\Delta_{k}^{2}}-\frac{1}{\Delta_{k-1}^{2}}\right)\right] \leq \frac{2}{\Delta_{K}}.
	$$
\end{lemma}
%
Therefore, 
\begin{align*}
    & \sum_{j=1}^K \Delta_{j,i} \cdot \mathbb{E}\left[\sum_{t=1}^{T} \mathsf{1}\left[\calA_t(i, j) \wedge \neg \mathcal{B}_{t}(i) \wedge \mathcal{C}_{t}(i,j) \right]\right]  
     \le  \frac{  16\log T  }{  \Delta_{K,i} } + \sum_{j=1}^K \Delta_{j,i}.
\end{align*}

Lastly, the total regret incurred by the second term in equation~\eqref{cascadeTS_eq:beta_decomp_N_ij} can be bounded as follows:
\begin{align*}
    & \sum_{i=K+1}^L \sum_{j=1}^K \Delta_{j,i} \cdot \mathbb{E}\left[\sum_{t=1}^{T} \mathsf{1}\left[\calA_t(i, j) \wedge \neg \mathcal{B}_{t}(i) \wedge \mathcal{C}_{t}(i,j) \right]\right]  
     \le 16 \log T \cdot \sum_{i=K+1}^L \frac{ 1 }{  \Delta_{K,i}   } 
     + \sum_{i=K+1}^L \sum_{j=1}^K \Delta_{j,i}.
\end{align*}


\subsection{Proof of Lemma~\ref{cascadeTS_lemma:beta_3rd_term_part}  }
\label{cascadeTS_pf:lemma_beta_3rd_term_part}


\begin{proof}
    We abbreviate $\varepsilon_j$ as $\varepsilon$ for simplicity.
    By the definition of $W_{i,j}$ and $W_{i,-j}$, $\thetaAlgBeta_t \in W_j$ yields that  there exists $i$ such that $\{ \thetaAlgBeta_t(i) \le w(j) -\varepsilon, \thetaAlgBeta_{t}(-j) \in W_{i,-j} \}$ , thus we can use
    $$
\mathbb{E}\left[\sum_{t=1}^{T} \operatorname{Pr}\left[ ~\exists i~s.t.~ \thetaAlgBeta_t(j) \le w(j) -\varepsilon, \thetaAlgBeta_{t}(-j) \in W_{i,-j}, \text{Observe $W_t(i)$ at $t$}  ~\right]\right]
    $$
as an upper bound.

Denote the value $p_{j, t}$ as $\operatorname{Pr}\left[\thetaAlgBeta_{t}(j)>w(j)-\varepsilon | \mathcal{F}_{t-1}\right] .$ Notice that according to the cascading model, given $\mathcal{F}_{t-1}$ , the value $\thetaAlgBeta_{t}(j)$ and the set of values $\thetaAlgBeta_{t}(-j)$ are independent. Besides, note that $\neg C_{i,j}(t) = \{ \thetaAlgBeta_t(i) \le w(j) -\varepsilon \}$ holds when $\thetaAlgBeta_t(-j) \in W_{i,-j}$. If we further have $\thetaAlgBeta_t(j) > w(j) - \varepsilon$, then $j \in S_t$.

\vspace{.5em}
\textbf{Observation of fixed $i$.} Fix $\thetaAlgBeta_t(-j) \in W_{i,-j}$, let 
\begin{itemize}
    \item ${S}_t = \{ {i}_1^t, {i}_2^t, \ldots, {i}_K^t  \} $, $i_{k_1}^t = i$ when $  \thetaAlgBeta_t(j) \le w(j) - \varepsilon$,
    \item $ \bar{S}_t = \{ \bar{i}_1^t, \bar{i}_2^t, \ldots, \bar{i}_K^t  \} $, $\bar{i}_{k_2}^t = j$ when $ \thetaAlgBeta_t(j)$ changes to $\bar{\thetaAlgBeta}_t(j)$ such that $ \bar{\thetaAlgBeta}_t(j) > w(j) - \varepsilon $.
\end{itemize}
Since $\thetaAlgBeta_t(i_i^t) \ge \thetaAlgBeta_t(i_2^t) \ge \ldots$, $\thetaAlgBeta_t( \bar{i}_1^t ) \ge \thetaAlgBeta_t( \bar{i}_2^t ) \ge \ldots$ according to Algorithm~\ref{cascadeTS_alg:ts_beta}, 
    $$\bar{\thetaAlgBeta}_t(j) > w(j) - \varepsilon \ge \thetaAlgBeta_t(i) = \thetaAlgBeta_t(i_{k_1}^t) \ge \thetaAlgBeta_t( i_{k_1+1}^t ) \ge \ldots.$$
Therefore, $k_2 \le k_1$ and  $i_l^t = \bar{i}_l^t$ for $1 \le l < k_2$. Besides, the observation of item $i$ or $j$ is determined by $\thetaAlgBeta_t(-j)$

With the observation and notation above, we have
\begin{align*}
    & \operatorname{Pr}\left[ \thetaAlgBeta_t(j) \le w(j) -\varepsilon, \thetaAlgBeta_{t}(-j) \in W_{i,-j}, \text{Observe $W_t(i)$ at $t$}  ~\big|~ \mathcal{F}_{t-1}\right] \\
    & = ( 1 - p_{j, t} ) \cdot \operatorname{Pr}\left[\thetaAlgBeta_{t}(-j) \in W_{i,-j} | \mathcal{F}_{t-1}\right]
    \cdot \text{Pr} \left[  \text{Observe $W_t(i)$ at $t$} ~\big|~ \thetaAlgBeta_t \right]\\
    & = ( 1 - p_{j, t} ) \cdot \operatorname{Pr}\left[\thetaAlgBeta_{t}(-j) \in W_{i,-j} | \mathcal{F}_{t-1}\right]
    \cdot \bbE \left[ \prod_{l=1}^{k_1-1} ( 1-w(i_l^t) ) ~\big|~ \thetaAlgBeta_t(-j) \right],
\end{align*}
and
\begin{align*}
    & \operatorname{Pr}\left[  j \in S_t, \thetaAlgBeta_{t}(-j) \in W_{i,-j}, \text{Observe $W_t(j)$ at $t$}  | \mathcal{F}_{t-1}\right] \\
    & \ge  \operatorname{Pr}\left[ \thetaAlgBeta_t(j) > w(j) -\varepsilon, \thetaAlgBeta_{t}(-j) \in W_{i,-j}, \text{Observe $W_t(j)$ at $t$}  | \mathcal{F}_{t-1}\right] \\
    & = p_{j, t} \cdot \operatorname{Pr}\left[\thetaAlgBeta_{t}(-j) \in W_{i,-j} | \mathcal{F}_{t-1}\right]
    \cdot \text{Pr} \left[  \text{Observe $W_t(j)$ at $t$} ~\big|~ \thetaAlgBeta_t(-j) \right] \\
    & = p_{j, t} \cdot \operatorname{Pr}\left[\thetaAlgBeta_{t}(-j) \in W_{i,-j} | \mathcal{F}_{t-1}\right]
    \cdot \bbE \left[ \prod_{l=1}^{k_2-1} ( 1-w(i_l^t) ) ~\big|~ \thetaAlgBeta_t(-j) \right] \\
    & \ge p_{j, t}\cdot \operatorname{Pr}\left[\thetaAlgBeta_{t}(-j) \in W_{i,-j} | \mathcal{F}_{t-1}\right]
    \cdot \bbE \left[ \prod_{l=1}^{k_1-1} ( 1-w(i_l^t) ) ~\big|~ \thetaAlgBeta_t(-j) \right].
\end{align*}
This leads to 
\begin{align*}
    & \operatorname{Pr}\left[\boldsymbol{\thetaAlgBeta}_t \in W_{i,j}, \text{Observe $W_t(i)$ at $t$}  \right] \\
    & \le ( 1 - p_{j, t} ) \cdot \operatorname{Pr}\left[\thetaAlgBeta_{t}(-j) \in W_{i,-j} ~\big|~  \mathcal{F}_{t-1}\right]
    \cdot \bbE \left[ \prod_{l=1}^{k_1-1} ( 1-w(i_l^t) ) ~\big|~ \thetaAlgBeta_t(-j) \right]\\
    & \le  \frac{  1 - p_{j, t} }{ p_{j,t} } \cdot \operatorname{Pr}\left[  j \in S_t, \thetaAlgBeta_{t}(-j) \in W_{i,-j}, \text{Observe $W_t(j)$ at $t$}  ~\big|~  \mathcal{F}_{t-1}\right].
\end{align*}

\textbf{Consider the overall case.} By definition, $W_{i,j}$ are disjoint sets for different $i$. Hence,
\begin{align}
    & \mathbb{E}\left[\sum_{t=1}^{T} \operatorname{Pr}\left[  ~\exists i~s.t.~  \boldsymbol{\thetaAlgBeta}_t \in W_{i,j}, \text{Observe $W_t(i)$ at $t$}  ~\right]\right] \nonumber \\
    & = \sum_{i=K+1}^{L} \mathbb{E}\left[\sum_{t=1}^{T} \operatorname{Pr}\left[ \boldsymbol{\thetaAlgBeta}_t \in W_{i,j}, \text{Observe $W_t(i)$ at $t$}  ~\right]\right] \nonumber \\
    & \le \sum_{t=1}^{T}   \mathbb{E}\left[ \sum_{i=K+1}^{L}   \mathbb{E}\left[\ \frac{  1 - p_{j, t} }{ p_{j,t} } \cdot \operatorname{Pr}\left[  j \in S_t, \thetaAlgBeta_{t}(-j) \in W_{i,-j}, \text{Observe $W_t(j)$ at $t$}  ~\big|~  \mathcal{F}_{t-1}\right]  \right] \right] \nonumber \\
    & = \sum_{t=1}^{T}   \mathbb{E}\left[ \frac{  1 - p_{j, t} }{ p_{j,t} } \cdot \operatorname{Pr}\left[   ~\exists i~s.t.~   j \in S_t, \thetaAlgBeta_{t}(-j) \in W_{i,-j}, \text{Observe $W_t(j)$ at $t$}  ~\big|~  \mathcal{F}_{t-1}\right] \right] \nonumber \\
    & \le \sum_{t=1}^{T}  \mathbb{E} \left[ \frac{  1 - p_{j, t} }{ p_{j,t} } \cdot \operatorname{Pr}\left[  j \in S_t, \text{Observe $W_t(j)$ at $t$}  ~\big|~  \mathcal{F}_{t-1}\right]  \right] \nonumber \\
    & = \sum_{t=1}^{T}  \mathbb{E} \left[ \frac{  1 - p_{j, t} }{ p_{j,t} } \cdot \mathsf{1} \left[  j \in S_t, \text{Observe $W_t(j)$ at $t$}   \right]  \right]. \nonumber
\end{align}

Let $\tau_{j, q}$ is the time step that item $j$ is observed for the $q$-th time. Then
\begin{align}
     &
     \mathbb{E}\left[\sum_{t=1}^{T} \operatorname{Pr}\left[  ~\exists i~s.t.~  \boldsymbol{\thetaAlgBeta}_t \in W_{i,j}, \text{Observe $W_t(i)$ at $t$}  ~\right]\right]
     \nonumber \\  
    & \leq \sum_{q=0}^{T-1} \mathbb{E}\left[\frac{1-p_{j, \tau_{j, q}+1}}{p_{j, \tau_{j, q}+1}} \sum_{t=\tau_{j, q}+1}^{\tau_{j, q+1}} \mathsf{1}\left[  j \in S_t, \text{Observe $W_t(j)$ at $t$}  \right] \right]  \label{cascadeTS_eq:beta_no_observ_btw} \\
    &= \sum_{q=0}^{T-1} \mathbb{E}\left[\frac{1-p_{j, \tau_{j, q}+1 }}{p_{j, \tau_{j, q} + 1 }}\right].  \nonumber
\end{align}
Line~\eqref{cascadeTS_eq:beta_no_observ_btw} follows from the fact that the probability $\operatorname{Pr}\left[\thetaAlgBeta_t(j)>w(j)-\varepsilon | \mathcal{F}_{t-1}\right]$ only changes when we get a feedback of item $j$, but during time steps in $\left[\tau_{j, q}+1, \tau_{j, q+1} - 1 \right],$ we do not have any such feedback.

Next, we apply a variation of Lemma 2.9 in \cite{agrawal2017near}:
\begin{lemma}[Implied by \cite{agrawal2017near}, Lemma 2.9]
    For any $j$, $p_{j, t}$ is $\operatorname{Pr}\left[\thetaAlgBeta_{t}(j)>w(j)-\varepsilon | \mathcal{F}_{t-1}\right] $, and $\tau_{j, q}$ is the time step that item $j$ is observed for the $q$-th time. Then
    \footnote{A (non-negative) function  $f(T) = O(T)$ if there exists a constant $0<c<\infty$ (dependent on $w$ but not $T$) such that $f(T)\le cT$ for sufficiently large $T$. Similarly $f(T)=\Omega(T) $ if there exists $c>0$ such that $f(T)\ge cT$ for sufficiently large $T$. Finally, $f(T)=\Theta(T)$ if $f(T)=O(T)$ and $f(T)=\Omega(T)$.}
    \begin{align*}
        & \mathbb{E}\left[\frac{1  }{p_{j, \tau_{j, q} +1 }}  -1 \right] \le 
        \left\{
            \begin{array}{ll}
                \displaystyle  \frac{  3  }{  \varepsilon }    
                &  \displaystyle   \text{if }  q < \frac{ 8  }{  \varepsilon },  \\ 
                \displaystyle  \Theta\left(  e^{ -\varepsilon^2 q/2 } + \frac{ 1 }{ (q+1) \varepsilon^2 } e^{-D_j q}  +  \frac{ 1 }{  e^{  \varepsilon^2 q/4 } -1 } \right) 
                & \text{else, } 
            \end{array}
        \right.
    \end{align*}
    where $D_j = \rmKL( w(j), w(j) - \varepsilon ) =  [ w(j) - \varepsilon ] \cdot \log  \left [ \frac{ w(j) - \varepsilon  }{ w(j) } \right] + [ 1 - w(j) + \varepsilon ]  \cdot \log \left[ \frac{  1 - w(j) + \varepsilon }{  1 - w(j)} \right].  $
\end{lemma}

According to Pinsker's and reverse Pinsker's inequality for any two distributions $ P$ and $ Q$ defined in the same finite space $X$, we have
\begin{align*}
	\delta(P,Q)^2 \le \frac{1}{2} \mathrm{KL}(P,Q)
	\le \frac{1}{\alpha_Q} \delta(P,Q)^2
\end{align*}
where
$
	\delta(P,Q) = \sup\{ | P(A)-Q(A)| \big| A \subset X \} \text{ and } \alpha_Q = \min_{x\in X: Q(x)>0} Q(x)
$. Therefore,
\begin{align*}
    D_j \ge 2 \{  w(j) - [ w(j) -  \varepsilon] \}^2  = 2\varepsilon^2.
\end{align*}
Hence, there exists a constant $0<c_1 < \infty$ such that
\begin{align*}
    & \sum_{q=0}^{T-1} \mathbb{E}\left[\frac{1-p_{j, \tau_{j, q}+1 }}{p_{j, \tau_{j, q} + 1 }}\right] \\
    & \le \sum_{ 0 \le q<  8 / \varepsilon } \frac{ 3 }{ \varepsilon }
            + \sum_{ T-1 \ge q\ge 8/ \varepsilon }  c_1 \cdot 
                \left[  e^{ -\varepsilon^2 q/2 } + \frac{ 1 }{ (q+1) \varepsilon^2 } e^{- 2 \varepsilon^2 q}  +  \frac{ 1 }{  e^{  \varepsilon^2 q/4 } -1 } \right]  \\
    & \le  \frac{ 24 }{ \varepsilon^2 }
            + c_1 \cdot \int_{ 8/ \varepsilon - 1 }^{T-1    \vphantom{\Big[ }       }   ~
                 e^{ -\varepsilon^2 x/2 } + \frac{ 1 }{ (x+1) \varepsilon^2 } e^{- 2 \varepsilon^2 x}  +  \frac{ 1 }{  e^{  \varepsilon^2 x/4 } -1 } ~\mathrm{d} x                  
\end{align*}
Note the Taylor series of $e^x$: $e^x = \sum_{n=0}^\infty  x^n/n! $ ($0!=1$). We have
\begin{align*}
    \text{(i) }
    & \int_{ 8/ \varepsilon - 1 }^{T-1 }  ~ e^{ -\varepsilon^2 x/2 }  ~\mathrm{d} x
    = -  \frac{ 2  }{ \varepsilon^2 } e^{ -\varepsilon^2 x/2 }  \bigg|_{ x=  8/ \varepsilon - 1  }^{T-1}
    =  \frac{ 2  }{ \varepsilon^2 } e^{  \varepsilon^2 /2 } \cdot \Big[ e^{ -\varepsilon^2 T/2 }   - e^{ -\varepsilon^2 (8/ \varepsilon)/2 }   \Big]
        \le \frac{  2 \sqrt{e} }{ \varepsilon^2 } ,  
    \\
    \text{(ii) }
    & \int_{ 8/ \varepsilon - 1 }^{T-1 }  ~ \frac{ 1 }{ (x+1) \varepsilon^2 } e^{- 2 \varepsilon^2 x} ~ \mathrm{d} x
    = e^{  \varepsilon^2 /2 } \cdot \int_{ 8/ \varepsilon - 1 }^{T-1 } ~  \frac{ 1 }{ (x+1) \varepsilon^2 } e^{- 2 \varepsilon^2 (x+1) } ~\mathrm{d} x
    = e^{  \varepsilon^2 /2 } \cdot \int_{ 8/ \varepsilon  }^{T }  ~ \frac{ 1 }{ x \varepsilon^2 } e^{- 2 \varepsilon^2 x} ~\mathrm{d} x  
    \\  &     
    =  e^{  \varepsilon^2 /2 }    \cdot 
            \int_{ 16 \varepsilon  }^{ 2T \varepsilon^2  \vphantom{\Big[ }   } ~  \frac{ e^{-x}  }{ x/2 } ~\mathrm{d} \Big( \frac{ x  }{   2 \varepsilon^2 } \Big)
    =  \frac{  e^{  \varepsilon^2 /2 }    }{   \varepsilon^2 }   \cdot 
            \int_{ 16 \varepsilon  }^{ 2T \varepsilon^2 }  ~  \frac{ e^{-x}  }{ x } ~\mathrm{d} x 
    \\ & 
    \le  \frac{  e^{  \varepsilon^2 /2 }    }{   \varepsilon^2 }   \cdot 
            \int_{ 16\varepsilon  }^{ 1 } ~   \frac{ e^{-x}  }{ x } ~\mathrm{d} x 
            \cdot \mathsf{1} [ 16\varepsilon < 1 ] 
            + \frac{  e^{  \varepsilon^2 /2 }    }{   \varepsilon^2 }   \cdot 
            \int_{ 1  }^{ 2T \varepsilon^2 }   ~ \frac{ e^{-x}  }{ x } ~\mathrm{d} x  
    \\ & 
    \le  \frac{  e^{  \varepsilon^2 /2 }    }{   \varepsilon^2 }   \cdot \int_{ 16\varepsilon  }^{ 1 }  ~ \frac{ 1  }{  16\varepsilon}  ~\mathrm{d} x  
            \cdot \mathsf{1} [ 16\varepsilon < 1 ] 
            + \frac{  e^{  \varepsilon^2 /2 }    }{   \varepsilon^2 }   \cdot 
            \int_{ 1  }^{ 2T \varepsilon^2 }  ~ e^{-x}  ~\mathrm{d} x  
    \\ & 
    \le  \frac{  e^{  \varepsilon^2 /2 }    }{   \varepsilon^2 }   \cdot  \frac{ 1  }{  16\varepsilon}  
            \cdot \mathsf{1} [ 16\varepsilon < 1 ] 
            + \frac{  e^{  \varepsilon^2 /2 }    }{   \varepsilon^2 }   \cdot  \frac{1}{ e },
    \\
    \text{(iii) }
    &
    \int_{ 8/ \varepsilon - 1 }^{T-1    \vphantom{\Big[ }       }   ~ \frac{ 1 }{  e^{  \varepsilon^2 x/4 } -1 } ~\mathrm{d} x  
    \le \int_{ 8/ \varepsilon - 1 }^{T-1        }  ~  \frac{ 1 }{    \varepsilon^2 x/4   } ~\mathrm{d} x 
    = \frac{ 4 }{ \varepsilon^2 }  \cdot \log x \Big|_{ 8/ \varepsilon - 1 }^{T-1    \vphantom{\Big[ }       }
    \le \frac{ 4 }{ \varepsilon^2 } \cdot \log T.
\end{align*} 
Let $c = c_1 \cdot ( 2\sqrt{e} + 5 ) + 24$, then (i) if $\epsilon < 1/ 16$,
\begin{align*}
    & \sum_{q=0}^{T-1} \mathbb{E}\left[\frac{1-p_{j, \tau_{j, q}+1 }}{p_{j, \tau_{j, q} + 1 }}\right] 
    \le  \frac{ 24 }{ \varepsilon^2 }
            + c_1 \cdot 
            \Big(
                \frac{ 2\sqrt{e} }{ \varepsilon^2 }  + \frac{ 2 }{ \sqrt{e} \cdot \varepsilon^3 } + \frac{ 4 }{ \varepsilon^2}  \cdot \log T
            \Big)
    \le c \cdot \bigg( \frac{ 1  }{ \varepsilon^3 } + \frac{ \log T }{ \varepsilon^2  }  \bigg);
\end{align*}
(ii) else, we have $\epsilon \ge 1/16$ and
\begin{align*}
    & \sum_{q=0}^{T-1} \mathbb{E}\left[\frac{1-p_{j, \tau_{j, q}+1 }}{p_{j, \tau_{j, q} + 1 }}\right] 
    \le  \frac{ 24 }{ \varepsilon^2 }
            + c_1 \cdot 
            \Big(
                \frac{ 2\sqrt{e} }{ \varepsilon^2 }  + \frac{ 1 }{ \sqrt{e} \cdot \varepsilon^2 } + \frac{ 4 }{ \varepsilon^2}  \cdot \log T
            \Big)
    \le  \frac{ c \log T }{ \varepsilon^2  }  .
\end{align*}

%
%

\end{proof}


\subsection{Proof of Lemma~\ref{cascadeTS_claim:conc}}\label{cascadeTS_pf:claimconc}
\begin{proof}
\textbf{Bounding probability of event ${\cal E}_{\hat{w}, t}: $} We first consider a fixed non-negative integer $N$ and a fixed item $i$. Let $Y_1, \ldots, Y_N$ be i.i.d. Bernoulli random variables, with the common mean $w(i)$. Denote $\hat{\xi}_N = \sum^N_{i=1}Y_i / N$ as the sample mean, and $\hat{V}_N = \hat{\xi}_N(1 - \hat{\xi}_N)$ as the empirical variance. By applying Theorem~\ref{cascadeTS_thm:emp_berstein} with $\delta = 1 / (t+1)^4$, we have
\begin{equation}\label{cascadeTS_eq:emp_bernstein_fixed}
\Pr\left(\left|\hat{\xi}_N - w(i) \right| > \sqrt{\frac{8\hat{V}_N\log(t+1)}{N}} + \frac{12\log(t+1)}{N}\right)\leq \frac{3}{(t+1)^4}.
\end{equation}
By an abuse of notation, let $\hat{\xi}_N = 0$ if $N= 0$. Inequality (\ref{cascadeTS_eq:emp_bernstein_fixed}) implies the following concentration bound when $N$ is non-negative:
\begin{equation}\label{cascadeTS_eq:emp_bernstein_fixed2}
\Pr\left(\left|\hat{\xi}_N - w(i) \right| > \sqrt{\frac{16\hat{V}_N\log(t+1)}{N + 1}} + \frac{24\log(t+1)}{N+1}\right)\leq \frac{3}{(t+1)^4}.
\end{equation}
Subsequently, we can establish the concentration property of $\hat{w}_t(i)$ by a union bound of $N_t(i)$ over $\{0, 1, \ldots, t-1\}$:
\begin{align*}
& \Pr\left(\left| \hat{w}_t(i) - w(i) \right| > \sqrt{\frac{16\hat{\nu}_t(i)\log(t+1)}{ N_t(i) + 1 }} + \frac{24\log(t + 1)}{N_t(i) + 1}\right) \\
&\leq   \Pr\left(\left|\hat{\xi}_N - w(i) \right| > \sqrt{\frac{16\hat{V}_N\log(t+1)}{N + 1}} + \frac{24\log(t+1)}{N+1} \text{ for some $N\in \{0, 1, \ldots t-1\}$}\right)\\
&\leq  \frac{3}{(t+1)^3}.
\end{align*}
Finally, taking union bound over all items $i\in L$, we know that event ${\cal E}_{\hat{w}, t}$ holds true with probability at least $1 - 3L / (t+1)^3$.


\textbf{Bounding probability of event ${\cal E}_{\theta, t} $, conditioned on event ${\cal E}_{\hat{w}, t}$:} Consider an observation trajectory $H_t$ satisfying event ${\cal E}_{\hat{w}, t}$. By the definition of the Thompson sample $\theta_t(i)$ (see Line 7 in Algorithm~\ref{cascadeTS_alg:ts}), we have
\begin{align}
&\Pr\left(\left|\theta_t(i) - \hat{w}_t(i)\right| > h_t(i) \text{ for all $i\in L$} | H_{\hat{w}, t}\right) \nonumber\\
&=   \Pr_{Z_t}\left( \left|Z_t \cdot \max\left\{\sqrt{\frac{\hat{\nu}_t(i)\log (t+1)}{N_t(i) + 1}}, \frac{\log (t+1)}{N_t(i) + 1}\right\}\right| > \right.\nonumber\\
&\left. \qquad \qquad  \sqrt{\log(t+1)} \left[\sqrt{\frac{16\hat{\nu}_t(i)\log (t + 1)}{N_t(i) + 1}} + \frac{24\log(t + 1)}{N_t(i) + 1}\right] \text{ for all $i\in [L]$}~\bigg |~ \hat{w}_t(i), N_t(i)\right)\nonumber\\
&\leq  \Pr\left( \left|Z_t \cdot \max\left\{\sqrt{\frac{\hat{\nu}_t(i)\log (t+1)}{N_t(i) + 1}}, \frac{\log (t+1)}{N_t(i) + 1}\right\}\right| > \right.\nonumber\\
&\left. \qquad \qquad    \sqrt{16\log(t+1)} \max\left\{\sqrt{\frac{\hat{\nu}_t(i)\log (t+1)}{N_t(i) + 1}}, \frac{\log (t+1)}{N_t(i) + 1}\right\} \text{ for all $i\in [L]$}~\bigg |~ \hat{w}_t(i), N_t(i)\right)\nonumber\\
&\leq  \frac{1}{2}\exp\left[-8\log(t+1)\right] \leq \frac{1}{2 (t+1)^3}. \label{cascadeTS_eq:by_gauss_conc}
\end{align}
The inequality in (\ref{cascadeTS_eq:by_gauss_conc}) is by the concentration property of a Gaussian random variable, see Theorem~\ref{cascadeTS_thm:gaussian}. Altogether, the lemma is proved.
\end{proof}

\subsection{Proof of Lemma~\ref{cascadeTS_lemma:transferone}}\label{cascadeTS_pf:lemmatransferone}

\begin{proof}
To start, we denote the shorthand $\theta_t(i)^+ = \max\{\theta_t(i), 0\} $. We demonstrate that, if events ${\cal E}_{\hat{w}, t},{\cal E}_{\theta, t} $ and inequality (\ref{cascadeTS_eq:crucial_ineq}) hold, then for all $\bar{S} = (\bar{i}_1, \ldots, \bar{i}_K)\in \bar{\cal S}_t$ we have:
\begin{equation}\label{cascadeTS_eq:required_ineq}
\sum^K_{k=1}\left[\prod^{k-1}_{j=1}(1-w(\bar{i}_j))\right]\cdot\theta_t(\bar{i}_k)^+ \overset{(\ddagger)}{<} \sum^K_{k=1}\left[\prod^{k-1}_{j=1}(1-w(j))\right]\cdot \theta_t(k)^+ \overset{(\dagger)}{\leq} \sum^K_{k=1}\left[\prod^{k-1}_{j=1}(1-w(i^t_j))\right]\cdot \theta_t(i^t_k)^+,
\end{equation}
where we recall that $S_t= (i^t_1, \ldots, i^t_K)$ in an optimal arm for $\bm \theta_t$, and $\theta_t(i^t_1)\geq \theta_t(i^t_2)\geq \ldots \geq \theta_t(i^t_K)\geq \max_{i\in [L]\setminus \{i^t_1, \ldots, i^t_K\}} \theta_t(i)$.  
The inequalities in (\ref{cascadeTS_eq:required_ineq}) clearly implies that $S_t\in {\cal S}_t$. To justifies these inequalities, we proceed as follows:

\textbf{Showing $(\dagger)$: } This inequality is true even without requiring events ${\cal E}_{\hat{w}, t},{\cal E}_{\theta, t} $ and inequality (\ref{cascadeTS_eq:crucial_ineq}) to be true. Indeed, we argue the following:
\begin{align}
\sum^K_{k=1}\left[\prod^{k-1}_{j=1}(1-w(j))\right]\cdot \theta_t(k)^+ & \leq \sum^K_{k=1}\left[\prod^{k-1}_{j=1}(1-w(j))\right]\cdot \theta_t(i^t_k)^+ \label{cascadeTS_eq:by_opt_theta}\\
& \leq \sum^K_{k=1}\left[\prod^{k-1}_{j=1}(1-w(i^t_j))\right]\cdot \theta_t(i^t_k)^+ \label{cascadeTS_eq:by_opt_w}.
\end{align}
To justify inequality (\ref{cascadeTS_eq:by_opt_theta}), consider function $f : \pi_K(L) \rightarrow \mathbb{R}$ defined as $$f\left((i_k)^K_{k=1}\right) := \sum^K_{k=1}\left[\prod^{k-1}_{j=1}(1-w(j))\right]\cdot \theta_t(i_k)^+.$$ We assert that $S_t\in \text{argmax}_{S\in \pi_K(L)} f(S)$. The assertion can be justified by the following two properties. First, by the choice of $S_t$, we know that $\theta_t(i^t_1)^+\geq \theta_t(i^t_2)^+\geq \ldots \geq \theta_t(i^t_K)^+\geq\max_{i\in [L]\setminus S_t}\theta_t(i)^+$. Second, the linear coefficients in the function $f$ are monotonic and non-negative, in the sense that $1\geq 1-w(1)\geq (1 - w(1))(1 - w(2))\geq \ldots \geq \prod^{K-1}_{k=1} (1 - w(k))\geq 0$. Altogether, we have $f(S_t )\geq f(S^*)$, hence inequality (\ref{cascadeTS_eq:by_opt_theta}) is shown.

Next, inequality (\ref{cascadeTS_eq:by_opt_w}) clearly holds, since for each $k\in [K]$ we know that  $\theta_t(i^t_k)^+ \geq 0$, and $\prod^{k-1}_{j=1}(1-w(j)) \leq \prod^{k-1}_{j=1}(1-w(i^t_j))$. The latter is due to the fact that $1\geq w(1)\geq w(2)\geq \ldots \geq w(K)\geq \max_{i\in [L]\setminus [K]}w(i)$. Altogether, inequality $(\dagger)$ is established.


\textbf{Showing $(\ddagger)$: } The demonstration crucially hinges on events ${\cal E}_{\hat{w}, t},{\cal E}_{\theta, t} $ and inequality (\ref{cascadeTS_eq:crucial_ineq}) being held true. For any $\bar{S} = (\bar{i}_1, \ldots, \bar{i}_K)\in \bar{\cal S}_t$, we have
\begin{align}
\sum^K_{k=1}\left[\prod^{k-1}_{j=1}(1-w(\bar{i}_j))\right] \theta_t(\bar{i}_k)^+ & \leq \sum^K_{k=1}\left[\prod^{k-1}_{j=1}(1-w(\bar{i}_j))\right] \left(w(\bar{i}_k) + g_t(\bar{i}_k) + h_t(\bar{i}_k)\right) \label{cascadeTS_eq:by_events_mu_theta}\\
&< \left\{\sum^K_{k=1}\left[\prod^{k-1}_{j=1}(1-w(\bar{i}_j))\right] w(\bar{i}_k)\right\} + r(S^* | \bm w) - r(\bar{S} | \bm w)  \label{cascadeTS_eq:by_event_bar_S_t}\\
&=  r(S^* | \bm w) = \sum^K_{k=1}\left[\prod^{k-1}_{j=1}(1-w(j))\right] w(k) \nonumber\\
&\leq \sum^K_{k=1}\left[\prod^{k-1}_{j=1}(1-w(j))\right] \theta_t(k) \leq \sum^K_{k=1}\left[\prod^{k-1}_{j=1}(1-w(j))\right] \theta_t(k)^+. \label{cascadeTS_eq:by_crucial_inequality}
\end{align} 
Inequality (\ref{cascadeTS_eq:by_events_mu_theta}) is by the assumption that events ${\cal E}_{\hat{w}, t}, {\cal E}_{\theta, t}$ are true, which means that for all $i\in [L]$ we have $\theta_t(i)^+ \leq \mu(i) + g_t(i) + h_t(i)$. Inequality (\ref{cascadeTS_eq:by_event_bar_S_t}) is by the fact that $S\in \bar{\cal S}_t$. 
Inequality (\ref{cascadeTS_eq:by_crucial_inequality}) is by our assumption that inequality (\ref{cascadeTS_eq:crucial_ineq}) holds.

Altogether, the inequalities $(\dagger, \ddagger)$ in (\ref{cascadeTS_eq:required_ineq}) are shown, and the Lemma is established.
\end{proof}

\subsection{Proof of Lemma~\ref{cascadeTS_lemma:gap_bound}}\label{cascadeTS_pf:lemmagapbound}
The proof of Lemma~\ref{cascadeTS_lemma:gap_bound} crucially uses the following lemma on the expression of the difference in expected reward between two arms:
\begin{restatable}[Implied by \cite{ZongNSNWK16}, Lemma 1]{lemma}{cascadeTSlemmaZongSysmmetryRegDecomp}\label{cascadeTS_lemma:zong_sysmmetry_reg_decomp}
Let $S = (i_1, \ldots, i_K)$, $S' = (i'_1, \ldots, i'_K)$ be two arbitrary ordered $K$-subsets of $[L]$. For any ${\bm w}, {\bm w}'\in \mathbb{R}^L$, the following equalities holds:
\begin{align*}
r(S | {\bm w}) - r(S' | {\bm w}') &= \sum^K_{k=1}\left[\prod^{k-1}_{j=1}(1 - w(i_j))\right] \cdot (w(i_k) - w'(i'_k)) \cdot \left[\prod^{K}_{j=k+1}(1 - w'(i'_j))\right]\\
&= \sum^K_{k=1}\left[\prod^{k-1}_{j=1}(1 - w'(i'_j))\right] \cdot (w(i_k) - w'(i'_k)) \cdot \left[\prod^{K}_{j=k+1}(1 - w(i_j))\right].
\end{align*}
\end{restatable} 
While Lemma~\ref{cascadeTS_lemma:zong_sysmmetry_reg_decomp} is folklore in the {cascading bandit literature}, we provide a proof  in Appendix~\ref{cascadeTS_pf:lemma_zong_sysmmetry_reg_decomp} for the sake of  completeness. Now, we proceed to the proof of Lemma~\ref{cascadeTS_lemma:gap_bound}:
\begin{proof}
In the proof, we always condition to the historical observation $H_t$ stated in the Lemma. To proceed with the analysis, we define $\tilde{S}_t = (\tilde{i}^t_1, \ldots, \tilde{i}^t_K)\in {\cal S}_t$ as an ordered $K$-subset that satisfies the following minimization criterion:
\begin{equation}\label{cascadeTS_eq:minimize_criterion}
\sum^K_{k=1}\left[\prod^{k-1}_{j=1} (1-w(\tilde{i}_j))\right](g_t(\tilde{i}_j) + h_t(\tilde{i}_j)) = \min_{S = (i_1, \ldots, i_K)\in {\cal S}_t} \sum^K_{k=1}\left[\prod^{k-1}_{j=1} (1-w(i_j))\right](g_t(i_j) + h_t(i_j)).
\end{equation}
We emphasize that both $\tilde{S}_t$ and the left hand side of (\ref{cascadeTS_eq:minimize_criterion}) are deterministic in the current discussion, where we condition on $H_t$. To establish tight bounds on the regret, we consider the truncated version, $\tilde{\bm \theta}_t\in [0, 1]^L$, of the Thompson sample $\bm \theta_t$. For each $i\in L$, define 
$$
\tilde{\theta}_t(i) = \min \{1, \max\{0, \theta_t(i)\}\}. 
$$ 
The truncated version $\tilde{\theta}_t(i)$ serves as a correction of $\theta_t(i)$, in the sense that the Thompson sample $\theta_t(i)$, which serves as a Bayesian estimate of click probability $w(i)$, should lie in $[0, 1]$. It is important to observe the following two properties hold under the truncated Thompson sample $\tilde{\bm\theta}_t$:
\begin{enumerate}[leftmargin = 2.3cm, label={\textbf{Property \arabic*}}]
\item Our pulled arm $S_t$ is still optimal under the truncated estimate $\tilde{\bm \theta}_t$, i.e. $$S_t\in \text{argmax}_{S\in \pi_K(L)} ~ r(S  | \tilde{\bm \theta}_t) .$$ Indeed, the truncated Thompson sample can be sorted in a descending order in the same way as for the original Thompson sample\footnote{Recall that that $\theta_t(i^t_1) \geq \theta_t(i^t_2)\geq \ldots \geq \theta_t(i^t_K)\geq \max_{i\in [L]\setminus \{i^t_k\}^K_{k=1}} \theta_t(i)$ for the original Thompson sample $\bm \theta_t$.}, i.e. $\tilde{\theta}_t(i^t_1) \geq \tilde{\theta}_t(i^t_2)\geq \ldots \geq \tilde{\theta}_t (i^t_K)\geq \max_{i\in [L]\setminus \{i^t_1, \ldots , i^t_K \}} \tilde{\theta}_t(i)$. The optimality of $S_t$ thus follows. 
\item For any $t, i$, if it holds that $\left|\theta_t(i) - w(i)\right|\leq g_t(i)  +h_t(i)$, then it also holds that  $|\tilde{\theta}_t(i) - w(i) |\leq g_t(i)  +h_t(i)$. Indeed, we know that $|\tilde{\theta}_t(i) - w(i) |\leq |\theta_t(i) - w(i) |$.
\end{enumerate}

Now, we use the ordered $K$-subset $\tilde{S}_t$ and the truncated Thompson sample $\tilde{\bm \theta}_t$ to decompose the conditionally expected round $t$ regret as follows:
\begin{align}
&r(S^* |{\bm w}) - r(S_t |{\bm w}) = \left[ r(S^* |{\bm w}) - r(\tilde{S}_t |{\bm w})\right] + \left[r(\tilde{S}_t | {\bm w}) - r(S_t | {\bm w}) \right]\nonumber\\
&\leq  \left[ r(S^* | \bm w) - r(\tilde{S}_t |\bm w)\right] + \left[r(\tilde{S}_t |\bm w) - r(S_t |\bm w) \right]\mathsf{1}({\cal E}_{\theta, t}) + \mathsf{1}( \neg {\cal E}_{\theta, t})\nonumber\\
&\leq  \underbrace{\left[ r(S^* | \bm w) - r(\tilde{S}_t |\bm w)\right]}_{(\diamondsuit)} + \underbrace{\left[r(\tilde{S}_t | \tilde{\bm \theta}_t) - r(S_t |\tilde{\bm \theta}_t) \right]}_{(\clubsuit)}\nonumber\\
&\qquad\qquad \qquad \qquad + \underbrace{\left[r(S_t |\tilde{\bm \theta}_t) - r(S_t | \bm w ) \right]\mathsf{1}( {\cal E}_{\theta, t})}_{(\heartsuit)} + \underbrace{\left[r(\tilde{S}_t | {\bm w} ) - r(\tilde{S}_t | \tilde{\bm\theta}_t) \right]\mathsf{1}({\cal E}_{\theta, t})}_{(\spadesuit)} + \mathsf{1}(\neg {\cal E}_{\theta, t}) \label{cascadeTS_eq:decompose}.
\end{align}
We bound $(\diamondsuit, \clubsuit, \heartsuit, \spadesuit)$ from above as follows:

\textbf{Bounding $(\diamondsuit)$: }By the assumption that $\tilde{S}_t = (\tilde{i}^t_1, \ldots, \tilde{i}^t_K)\in {\cal S}_t$, with certainty we have 
\begin{equation}\label{cascadeTS_eq:bound_diamond}
(\diamondsuit) \leq \sum^K_{k=1}\left[\prod^{k-1}_{j=1} (1-w(\tilde{i}^t_j))\right](g_t(\tilde{i}^t_j) + h_t(\tilde{i}^t_j)).
\end{equation}

\textbf{Bounding $(\clubsuit)$: }By \textbf{Property 1} of the truncated Thompson sample $\tilde{\bm \theta}_t$, we know that $r(S_t | \tilde{\bm \theta}_t) = \max_{S\in \pi_K(L)} ~ r(S | \tilde{\bm\theta}_t) \geq r(\tilde{S}_t | \tilde{\bm \theta}_t)$. Therefore, with certainty we have  
\begin{equation}\label{cascadeTS_eq:bound_club}
(\clubsuit) \leq 0.
\end{equation}

\textbf{Bounding $(\heartsuit)$: }. 
We bound the term as follows:
\begin{align}
&\mathsf{1}({\cal E}_{\theta, t}) \left[r(S_t | \tilde{\bm \theta}_t) - r(S_t |\bm w ) \right]\nonumber\\
&=  \mathsf{1}({\cal E}_{\theta, t}) \sum^K_{k=1}\left[\prod^{k-1}_{j=1}(1 - w(i^t_j))\right] \cdot (\tilde{\theta}_t(i^t_k) - w(i^t_k)) \cdot \left[\prod^{K}_{j=k+1}(1 - \tilde{\theta}_t(i^t_j))\right]\label{cascadeTS_eq:heart_by_sym}\\
&\leq  \mathsf{1}({\cal E}_{\theta, t}) \sum^K_{k=1}\left[\prod^{k-1}_{j=1}(1 - w(i^t_j))\right] \cdot \left|\tilde{\theta}_t(i^t_k) - w(i^t_k)\right| \cdot \left[\prod^{K}_{j=k+1}\left|1 - \tilde{\theta}_t(i^t_j)\right|\right]\nonumber\\
&\leq \mathsf{1}({\cal E}_{\theta, t})  \sum^K_{k=1}\left[\prod^{k-1}_{j=1}(1 - w(i^t_j))\right] \cdot \left[ g_t(i^t_k)  +  h_t(i^t_k)\right] \label{cascadeTS_eq:heart_by_warm_start}\\
&\leq  \sum^K_{k=1}\left[\prod^{k-1}_{j=1}(1 - w(i^t_j))\right] \cdot \left[ g_t(i^t_k)  +  h_t(i^t_k)\right] \label{cascadeTS_eq:bound_heart}.
\end{align}
Equality (\ref{cascadeTS_eq:heart_by_sym}) is by applying the second equality in Lemma~\ref{cascadeTS_lemma:zong_sysmmetry_reg_decomp}, with $S = S' = S_t$, as well as $\bm w' \leftarrow \bm w$, $\bm w\leftarrow \bm \theta_t$. Inequality  (\ref{cascadeTS_eq:heart_by_warm_start}) is by the following two facts: (1) By the definition of the truncated Thompson sample $\tilde{\bm \theta}$, we know that
$\left|1-\tilde{\theta}_t(i)\right| \leq 1$ for all $i\in [L]$; (2) By assuming event ${\cal E}_{\theta, t}$ and conditioning on $H_t$ where event ${\cal E}_{\hat{w}, t}$ holds true, \textbf{Property 2} implies that that $|\tilde{\theta}_t(i) - w(i)| \leq g_t(i) + h_t(i)$ for all $i$. 

\textbf{Bounding $(\spadesuit)$: }The analysis is similar to the analysis on $(\heartsuit)$:
\begin{align}
&\mathsf{1}( {\cal E}_{\theta, t}) \left[ r(\tilde{S}_t |{\bm w} ) - r(\tilde{S}_t | \tilde{\bm \theta}_t  ) \right]\nonumber\\
&=  \mathsf{1}({\cal E}_{\theta, t}) \sum^K_{k=1}\left[\prod^{k-1}_{j=1}(1 - w(\tilde{i}^t_j))\right] \cdot (w(\tilde{i}^t_k) - \tilde{\theta}_t(\tilde{i}^t_k)) \cdot \left[\prod^{K}_{j=k+1}(1 - \tilde{\theta}_t(\tilde{i}^t_j))\right]\label{cascadeTS_eq:spade_by_sym}\\
& \leq  \mathsf{1}({\cal E}_{\theta, t})  \sum^K_{k=1}\left[\prod^{k-1}_{j=1}(1 - w(\tilde{i}^t_j))\right] \cdot \left[ g_t(\tilde{i}^t_k) +  h_t(\tilde{i}^t_k)\right] \label{cascadeTS_eq:spade_by_warm_start}\\
&\leq  \sum^K_{k=1}\left[\prod^{k-1}_{j=1}(1 - w(\tilde{i}^t_j))\right] \cdot \left[ g_t(\tilde{i}^t_k) +  h_t(\tilde{i}^t_k)\right] \label{cascadeTS_eq:bound_spade}.
\end{align}
Equality (\ref{cascadeTS_eq:spade_by_sym}) is by applying the first equality in Lemma~\ref{cascadeTS_lemma:zong_sysmmetry_reg_decomp}, with $S = S' = \tilde{S}_t$, and $\bm w \leftarrow \bm w$, $\bm w' \leftarrow \bm \theta_t$. Inequality (\ref{cascadeTS_eq:spade_by_warm_start}) follows the same logic as inequality (\ref{cascadeTS_eq:heart_by_warm_start}).

Altogether, collating the bounds (\ref{cascadeTS_eq:bound_diamond},~\ref{cascadeTS_eq:bound_club},~\ref{cascadeTS_eq:bound_heart},~\ref{cascadeTS_eq:bound_spade}) for $(\diamondsuit, \clubsuit, \heartsuit, \spadesuit)$
respectively, we bound (\ref{cascadeTS_eq:decompose}) from above (conditioned on $H_t$) as follows:
\begin{align}
r(S^* |{\bm w}) - r(S_t |{\bm w})&\leq 2\sum^K_{k=1}\left[\prod^{k-1}_{j=1} (1-w(\tilde{i}^t_j))\right](g_t(\tilde{i}^t_j) + h_t(\tilde{i}^t_j)) \nonumber\\
&\qquad + \sum^K_{k=1}\left[\prod^{k-1}_{j=1}(1 - w(i^t_j))\right] \cdot \left[ g_t(i^t_k)  +  h_t(i^t_k)\right] + \mathsf{1}(\neg {\cal E}_{\theta, t}).\label{cascadeTS_eq:bound_decompose}
\end{align}
Now, observe that 
\begin{align}
& \mathbb{E}_{{\bm \theta}_t}\left[ \sum^K_{k=1}\left[\prod^{k-1}_{j=1} (1-w(i^t_j))\right](g_t(i^t_j) + h_t(i^t_j)) ~ \Big | ~ H_t \right]\nonumber\\ 
&\geq  \mathbb{E}_{{\bm \theta}_t}\left[ \sum^K_{k=1}\left[\prod^{k-1}_{j=1} (1-w(i^t_j))\right](g_t(i^t_j) + h_t(i^t_j)) ~ \Big | ~ H_t, S_t\in {\cal S}_t \right]\Pr_{{\bm \theta}_t}\left[ S_t\in {\cal S}_t \bigg | ~ H_t\right]\nonumber\\
&\geq   \left\{\sum^K_{k=1}\left[\prod^{k-1}_{j=1} (1-w(\tilde{i}_j))\right](g_t(\tilde{i}_j) + h_t(\tilde{i}_j))\right\}\cdot \left(c - \frac{1}{2(t+1)^3}\right).
\end{align}
Thus, taking conditional expectation $\mathbb{E}_{\bm \theta_t}[\cdot | H_t]$ on both sides in inequality (\ref{cascadeTS_eq:bound_decompose}) gives
\begin{align*}
&\mathbb{E}_{\bm \theta_t}[ R(S^* |{\bm w}) - R(S_t |{\bm w}) | H_t]\nonumber\\
&\qquad\leq \left(1 + \frac{2}{c - \frac{1}{2(t + 1)^3}}\right) \mathbb{E}_{\bm \theta_t}\left[ \sum^K_{k=1}\left[\prod^{k-1}_{j=1}(1 - w(i^t_j))\right] \cdot \left[ g_t(i^t_k)  +  h_t(i^t_k)\right] ~\bigg | ~ H_t\right] + \mathbb{E}_{\bm \theta_t}[\mathsf{1}(\neg {\cal E}_{\theta, t})  | H_t] .
\end{align*}
Finally, the Lemma is proved by the assumption that $c > 1/(t+1)^3$, and noting from Lemma~\ref{cascadeTS_claim:conc} that $\mathbb{E}_{\bm \theta_t}[\mathsf{1}(\neg {\cal E}_{\theta, t})  | H_t] \leq 1/(2(t+1)^3)$. 
\end{proof}


\subsection{Proof of Lemma~\ref{cascadeTS_claim:concLin}}\label{cascadeTS_pf:claimconcLin} 

\begin{proof}
    \textbf{Bounding probability of event ${\cal E}_{\hat{\muLin}, t}: $} We use Theorem~\ref{cascadeTS_thm:linear} with $\eta_{t}(i) = W_t(i) - x(i)^T w$, $\calF_t'  = \{S_q\}^{t-1}_{q=1} \cup \{\left(i^q_k, W_q(i^q_k)\right)^{\min\{k_t, \infty\}}_{k=1}\}^{t-1}_{q=1}\cup S_t \cup (x(i))_{i=1}^K$. (Note that effectively, $\calF_t'$ has all the information, including the items observed, until time $t + 1$, except for the clickness of the selected items at time $t + 1$). By the definition of $\calF'_t$, $x(i)$ is $(\calF'_{t-1})$-measurable, and $\eta_t$ is $\calF'_t$-measurable. Also, $\eta_t$ is conditionally $R$-sub-Gaussian due to the the problem settings (refer
to Remark~\ref{cascadeTS_remark:sub_gauss}), and is a martingale difference process:
    \begin{align*}
        \bbE[ \eta_{t}(i) | \calF'_{t-1}] = \bbE[ W_t(i)| x(i) ] - x(i)^T w = 0.
    \end{align*}
Also, let
    \begin{align*}
        Q_t & = I_d + \sum_{q=1}^t \sum_{k=1}^{k_q} x(i_k^q) x(i_k^q)^T,\\
        \zeta_t & = \sum_{q=1}^t \sum_{k=1}^{k_q} x(i_k^q) \eta_q(i_k^q)  = \sum_{q=1}^t \sum_{k=1}^{k_q} x(i_k^q) [ W_q(i_k^q) - x(i_k^q)^T w ],
    \end{align*}
Note that $\NLin_t = Q_{t-1}$, recall $\bLin_{t+1} =  \sum_{q=1}^t \sum_{k=1}^{k_q} x(i_k^q) W(i_k^q)$ and $\hat{\muLin}_t = \NLin_t^{-1} \bLin_t$,
    \begin{align*}
        & Q_{t-1} (\hat{\muLin}_t - w ) = Q_{t-1} (\NLin_{t}^{-1} \bLin_t - w ) = \bLin_t - Q_{t-1}w
        \\ &
            = \sum_{q=1}^{t-1} \sum_{k=1}^{k_q} x(i_k^q) W(i_k^q) - \left[  I_d + \sum_{q=1}^{t-1} \sum_{k=1}^{k_q} x(i_k^q) x(i_k^q)^T \right] w \\
        & = \sum_{q=1}^{t-1} \sum_{k=1}^{k_q} x(i_k^q) [ W_q(i_k^q) - x(i_k^q)^T w ] - w = \zeta_{t-1} -w
    \end{align*}
yields that $\hat{\muLin}_t - w = Q_{t-1}^{-1}(\zeta_{t-1} - w)$. Let for any vector $z \in \bbR^{d\times 1}$ and matrix $A \in \bbR^{d\times d}$, $\|z\|_A:= \sqrt{y^TAy}$. Then, for all $i$,
    \begin{align*}
        & \left| x(i)^T \hat{\muLin}_t - x(i)^T w \right| = \left| x(i)^T Q_{t-1}^{-1}(\zeta_{t-1} - w) \right| 
        \\ &
            \le \|x(i)\|_{Q_{t-1}^{-1}} \| \zeta_{t-1} - w \|_{Q_{t-1}^{-1}} =  \|x(i)\|_{\NLin_t^{-1}} \| \zeta_{t-1} - w \|_{Q_{t-1}^{-1}},
    \end{align*}
where the inequality holds because $Q_{t-1}^{-1}$ is a positive definite matrix. Using Theorem~\ref{cascadeTS_thm:linear}, for any $\delta' > 0$, $t \ge 1$, with probability at least $1 - \delta'$,
    \begin{align*}
        \| \zeta_{t-1} \|_{Q_{t-1}^{-1}} \le R \sqrt{d \log \left( \frac{t}{\delta'} \right) }.
    \end{align*}
Therefore, $\| \zeta_{t-1} - w \|_{Q_{t-1}^{-1}} \le \| \zeta_{t-1}  \|_{Q_{t-1}^{-1} } + \| w \|_{Q_{t-1}^{-1}} \le R \sqrt{d \log \left( \frac{t}{\delta'} \right) } + \| w\| \le R \sqrt{d \log \left( \frac{t}{\delta'} \right) } + \cZeroLin $. Substituting $\delta' = \frac{1}{t^2}$, we get that with probability $1 - \frac{1}{t^2}$, for all $i$,
    \begin{align*}
        & \left| x(i)^T \hat{\muLin}_t - x(i)^T w \right|  \le \|x(i)\|_{\NLin_t^{-1}}  \cdot \left( R \sqrt{d \log \left( \frac{t}{\delta'} \right) } + \cZeroLin \right) 
        \\ &
        = \|x(i)\|_{\NLin_t^{-1}} \cdot \left( R \sqrt{3d \log t  } +  \cZeroLin \right) = l_t u_t(i) = g_t(i).
    \end{align*}
This proves the bound on the probability of ${\cal E}_{\hat{\muLin}, t}$.

\textbf{Bounding probability of event ${\cal E}_{\thetaLin, t} $:} Given any filtration $H_t$, $x(i),\NLin_t$ are fixed. since $\ZLin_t=( \ZLin_t^1, \ZLin_t^2, \ldots, \ZLin_t^d  )^T$ is a standard multivariate normal vector (mean $0_{d\times 1}$ and covariance $I_d$), each $\ZLin_t^l\ (1\le l \le d)$ is a standard univariate normal random variable (mean $0$ and variance $1$). 
Theorem~\ref{cascadeTS_thm:gaussian} implies
    \begin{align*}
        & \Pr( | \ZLin_t^l | > \sqrt{4 \log t} ) \le \frac{1}{2} \exp(-2\log t) = \frac{1}{2t^2}  \ \forall 1\le l\le d, \\
        \text{and then } & \Pr( \| \ZLin_t \|  > \sqrt{4d\log t} )  \le \sum_{l=1}^d Pr( | \ZLin_t^l | > \sqrt{4\log t} ) \le \frac{d}{2t^2}.
    \end{align*}
    Further,
    \begin{align*}
        & \left|  x(i)^T \thetaLin_t -  x(i)^T \hat{\muLin}_t \right| = | x(i)^T [ \thetaLin_t -  \hat{\muLin}_t] |
            = |  x(i)^T \NLin_t^{-1/2} \NLin_t^{-1/2} [ \thetaLin_t -  \hat{\muLin}_t] |\\
        & \le \lambda v_t \sqrt{K} \sqrt{ x(i)^T \NLin_t^{-1} x(i) }  \cdot \left\| \left( \frac{1}{\lambda v_t\sqrt{K} } \NLin_t^{1/2}    [ \thetaLin_t -  \hat{\muLin}_t] \right) \right\| = \lambda v_t \sqrt{K} u_t(i) \|\ZLin_t\| \le \lambda v_t \sqrt{K} u_t(i) \sqrt{4d\log t}
    \end{align*}
    with probability at least $1 - \frac{d}{2t^2}$.
    
    Alternatively, we can bound $ \left|  x(i)^T \thetaLin_t -  x(i)^T \hat{\muLin}_t \right|$ for every $i$ by considering that $x(i)^T\thetaLin_t$ is a Gaussian random variable with mean $x(i)^T\hat{\muLin}(t)$ and variance $\lambda^2 v_t^2K u_t(i)^2$. Therefore, again using Theorem~\ref{cascadeTS_thm:gaussian}, for every $i$, 
    \begin{align*}
        Pr\left( \left|  x(i)^T \thetaLin_t -  x(i)^T \hat{\muLin}_t \right| >   \lambda v_t \sqrt{K} u_t(i)\sqrt{ 4\log(t L) } \right) \le \frac{1}{2} \exp\left[ - \frac{ 4\log(tL ) }{2} \right] = \frac{1}{ 2 t^2 L^2}.
    \end{align*}
Taking union bound over $i = 1,\ldots ,L$, we obtain that $\left|  x(i)^T \thetaLin_t -  x(i)^T \hat{\muLin}_t \right| >  \lambda v_t \sqrt{K} u_t(i)\sqrt{ \log(tL ) }$ holds for all $i\in [L]$ with probability at least $1-\frac{1}{2t^2}$.


   Combined, the two bounds give that  ${\cal E}_{\thetaLin, t} $ holds with probability at least $1 - \frac{d}{t^2}$.
    
    Altogether, the lemma is proved.
    
\end{proof}

\subsection{Proof of Lemma~\ref{cascadeTS_lemma:transfertwoLin} }
\label{cascadeTS_pf:lemmatransfertwoLin}

\begin{proof} 
%
Recall $\alpha(1) := 1$, and $\alpha(k) = \prod^{k-1}_{j=1}(1 - w(j) ) = \prod^{k-1}_{j=1}(1 - x(j)^T \beta )$ for $2\leq k\leq K$. By the second part of Lemma~\ref{cascadeTS_claim:concLin}, we know that $\Pr [ {\cal E}_{\thetaLin, t} | H_t ] \geq 1 - \frac{d}{t^2} $, so to complete this proof, it suffices to show that $\Pr [ \text{(\ref{cascadeTS_eq:crucial_ineqLin}) holds} ~ | ~ H_t]  \geq  c$. For this purpose, consider
\begin{align}
    &\Pr_{{ \thetaLin}_t}\left[\sum^K_{k=1}\alpha(k)x(k)^T\thetaLin_t \geq \sum^K_{k=1}\alpha(k) x(k)^T \beta ~ \bigg | ~ H_t\right] \nonumber\\
    &= \Pr_{\ZLin_t} \left[\sum^K_{k=1}\alpha(k) x(k)^T \big( \hat{\muLin}_t \! +\!   \lambda v_t \sqrt{K}\NLin_t^{-1/2}\ZLin_t \big) 
    	\geq  \sum^K_{k=1}\alpha(k) x(k)^T \beta  ~ \bigg | ~ H_t \right] \label{cascadeTS_eq:equiv_gaussian_lin}\\
    &\geq \Pr_{\ZLin_t}\left[\sum^K_{k=1}\alpha(k) \left[x(k)^T \beta - g_t(k)\right] 
     + \lambda v_t\sqrt{K} \sum^K_{k=1}\alpha(k) x(k)^T  \NLin_t^{-1/2}\ZLin_t\geq\sum^K_{k=1}\alpha(k) x(k)^T \beta ~ \bigg |~ H_t \right] \label{cascadeTS_eq:by_E_mu_hat_lin}\\
   & =\Pr_{\ZLin_t} \left[  \lambda v_t \sum^K_{k=1}\alpha(k) x(k)^T  \NLin_t^{-1/2}\ZLin_t \geq  \sum^K_{k=1}\alpha(k)  g_t(k) ~ \bigg | ~ H_t \right] \nonumber\\
   & \geq  \frac{1}{4\sqrt{\pi}}\exp\left\{-\frac{7}{2}\left[ \sum^K_{k=1}\alpha(k)  g_t(k) \bigg / \stdLin_t  \right]^2 \right\}\label{cascadeTS_eq:by_anti_conc_lin}\\
   & =  \frac{1}{4\sqrt{\pi}}\exp\left\{-\frac{7}{2}\left[ \sum^K_{k=1}\alpha(k)  g_t(k)   \right]^2\bigg / \lambda^2 v_t^2 \sum^K_{k=1}\alpha(k)^2 u_t(k)^2 \right\}\nonumber \\
   & =  \frac{1}{4\sqrt{\pi}}\exp\left\{-\frac{7l_t^2 }{2 \lambda^2  v_t^2K }\left[ \sum^K_{k=1}\alpha(k)  u_t(k)   \right]^2\bigg / \sum^K_{k=1}\alpha(k)^2 u_t(k)^2 \right\}\label{cascadeTS_eq:by_def_g_t_lin}\\
   & \geq  \frac{1}{4\sqrt{\pi}}\exp\left(-\frac{7 }{2\lambda^2 } \right) : = \clin.\label{cascadeTS_eq:cauchy_c_lin} 
\end{align}
Step~\eqref{cascadeTS_eq:equiv_gaussian_lin} is by the definition of  $\thetaLin_t$ in Line 5 in Algorithm~\ref{cascadeTS_alg:casLinTS'}. 
Step~\eqref{cascadeTS_eq:by_E_mu_hat_lin} is by the Lemma assumption that $H_t\in \mathcal{H}_{\hat{\muLin}, t}$, which indicates that $x(k)^T\hat{\muLin}_t \geq x(k)^T \beta - g_t(k)$ for all $k\in [K]$. Noted that $\lambda v_t \sum^K_{k=1}\alpha(k) x(k)^T  \NLin_t^{-1}\ZLin_t$ is a Normal random variable with mean $0$ and variance 
    \begin{align*}
        \stdLin_t^2:=\lambda^2 v_t^2K \sum^K_{k=1}\alpha(k)^2 x(k)^T  \NLin_t^{-1} x(k)=\lambda^2 v_t^2K \sum^K_{k=1}\alpha(k)^2 u_t(k)^2,
    \end{align*}
step~\eqref{cascadeTS_eq:by_anti_conc_lin} is an application of the anti-concentration inequality of a normal random variable in Theorem~\ref{cascadeTS_thm:gaussian}. Step~\eqref{cascadeTS_eq:by_def_g_t_lin} is by applying the definition of $g_t(i)$. Step~\eqref{cascadeTS_eq:cauchy_c_lin} follows from $t\ge 4$ which implying $l_t \le v_t$, and Cauchy-Schwartz inequality.
\end{proof}

\subsection{Deduction of Inequality~\eqref{cascadeTS_eq:bound_y_other} in Section~\ref{cascadeTS_sec:lin_alg_cas} } \label{cascadeTS_sec:bound_y_other}
{
To derive the inequality~\eqref{cascadeTS_eq:bound_y_other}
	$$\sum_{t=1}^T \sqrt{ y_t^T \NLin_t^{-1} y_t } \le 5\sqrt{dT\log T},$$
we use the following result:
}
\begin{lemma} \label{cascadeTS_lemma:bound_y_other}
[Implied by \cite{Auer02}, Lemma 11; \cite{AgrawalG17}, Lemma 9]
Let $A' = A+ y y^T$ , where $y \in \bbR^d$, $A, A' \in \bbR^{d\times d}$, and all the eigenvalues $\lambda_j , j = 1, \ldots, d$ of $A$ are greater than or equal to $1$. Then, the eigenvalues $\lambda'_j , j = 1, \ldots, d$ of $A'$ can be arranged so that $\lambda_j \le \lambda'_j $ for all $j$, and
	$$
	y^T A^{-1} y \le 10 \sum_{j=1}^d \frac{\lambda'_j - \lambda_j}{ \lambda_j }.
	$$
\end{lemma}
Let $\lambda_{j,t}$ denote the eigenvalues of $\NLin_t$. Note that $\NLin_{t + 1} = \NLin_t + y_t y_t^T$, and $\lambda_{j,t} \ge 1, \forall j$. Above implies that
\begin{align*}
	y_t^T \NLin_t^{-1} y_t \le 10 \sum_{j=1}^d \frac{\lambda_{j,t+1} - \lambda_{j,t} }{ \lambda_{j,t} }.
\end{align*}
This allows us to derive the inequality after some algebraic computations following along the lines of Lemma 3 in \cite{chu2011contextual}.

\subsection{Proof of Lemma~\ref{cascadeTS_lemma:zong_sysmmetry_reg_decomp}} \label{cascadeTS_pf:lemma_zong_sysmmetry_reg_decomp} 


\begin{proof}
Observe that 
\begin{align*}
&\sum^K_{k=1}\left[\prod^{k-1}_{j=1}(1 - w(i_j))\right] \cdot (w(i_k) - w'(i'_k)) \cdot \left[\prod^{K}_{j=k+1}(1 - w'(i'_j))\right] \\ 
& =  \sum^K_{k=1}\left\{\left[\prod^{k-1}_{j=1}(1 - w(i_j))\right]\cdot \left[\prod^{K}_{j=k}(1 - w'(i'_j))\right] - \left[\prod^{k}_{j=1}(1 - w(i_j))\right]\cdot \left[\prod^{K}_{j=k+1}(1 - w'(i'_j))\right]\right\} \\
& = \prod^K_{k=1}(1- w'(i'_k)) - \prod^K_{k=1}(1 - w(i_k)) = R(S | \bm w) - R(S' | \bm w'), 
\end{align*}
and also that (actually we can also see this by a symmetry argument)
\begin{align*}
&\sum^K_{k=1}\left[\prod^{k-1}_{j=1}(1 - w'(i'_j))\right] \cdot (w(i_k) - w'(i'_k)) \cdot \left[\prod^{K}_{j=k+1}(1 - w(i_j))\right]\\
& =  \sum^K_{k=1}\left\{\left[\prod^{k}_{j=1}(1 - w'(i'_j))\right]\cdot \left[\prod^{K}_{j=k+1}(1 - w(i_j))\right] - \left[\prod^{k-1}_{j=1}(1 - w'(i'_j))\right]\cdot \left[\prod^{K}_{j=k}(1 - w(i_j))\right]\right\} \\
& = \prod^K_{k=1}(1- w'(i'_k)) - \prod^K_{k=1}(1 - w(i_k))= R(S | \bm w) - R(S' | \bm w').
\end{align*}
This completes the proof.
\end{proof}

\subsection{Proof of Lemma~\ref{cascadeTS_claim:rewarddiff} }
\label{cascadeTS_pf:rewarddiff} 

\begin{proof}
Let $Q = \left| S \setminus S^{*, \ell} \right|$. It is clear that $r(S | w^{(\ell)} ) = 1 - \left[1 - \frac{1 - \epsilon}{K}\right]^Q \left[1 - \frac{1 + \epsilon}{K}\right]^{K -Q}$, and then 
\begin{align}
r(S^{*, \ell} | w^{(\ell)} ) - r(S | w^{(\ell)}) &= \left[1 - \frac{1 + \epsilon}{K}\right]^{K -Q} \left[ \left(1 - \frac{1 - \epsilon}{K}\right)^Q - \left(1 - \frac{1 + \epsilon}{K}\right)^Q  \right]\nonumber\\
&= \left[1 - \frac{1 + \epsilon}{K}\right]^{K -Q} \frac{Q}{K}\left(1 - \frac{1 - \bar{\epsilon}}{K}\right)^{Q - 1} \cdot 2 \epsilon \label{cascadeTS_eq:pf_claim_reward_diff_1}\\
& \geq \frac{2Q}{K} \left(1 - \frac{1 + \epsilon}{K}\right)^{K - 1}\epsilon >  \frac{2Q}{K} \left(1 - \frac{1 + \epsilon}{K}\right)^{K}\epsilon \nonumber\\ 
& \geq \frac{2Q\epsilon}{K} \exp\left[-2(1 + \epsilon)\right] \geq \frac{2Q\epsilon}{e^{4} K} = 
\frac{2  \left| S \setminus S^{*, \ell} \right|\epsilon}{e^{4}K}.
  \label{cascadeTS_eq:pf_claim_reward_diff_2}
\end{align}
Step~\eqref{cascadeTS_eq:pf_claim_reward_diff_1} is by the application of the Mean Value Theorem on function $f:[-\epsilon , \epsilon] \rightarrow \mathbb{R}$ defined as $f(\varepsilon) = \left(1 - \frac{1 - \varepsilon}{K}\right)^Q$. In Step~\eqref{cascadeTS_eq:pf_claim_reward_diff_1}, $\bar{\epsilon}$ is some number lying in $[-\epsilon, \epsilon]$. Step~\eqref{cascadeTS_eq:pf_claim_reward_diff_2} is by the fact that $1 - \delta > \exp(-2\delta)$ for all $\delta \in [0, 1/4]$, and applying $\delta = (1 + \epsilon)/ K \in [0, 1/4]$ to the fact.
\end{proof}

\subsection{Proof of Lemma~\ref{cascadeTS_claim:lbregrettoKL} }
\label{cascadeTS_pf:claimlbregrettoKL} 


\begin{proof}
	To proceed, we define a uni-variate random variable $J^{\pi}(S)$, for $S$ being a $K$-subset of $\{1, \ldots, L\}$:
	$$
J^{\pi}(S) = \sum^T_{t=1}\sum_{j\in [L]}\frac{\mathsf{1}(j\in S^{\pi}_t) \cdot \mathsf{1}(j\in S)}{KT}.
$$
The random variable $J^{\pi}(S) \in [0, 1]$ always, and note that for any fixed $S$, the random variable $J^{\pi}(S)$ is a deterministic function of the stochastic process $\{S^\pi_t, O^\pi_t\}^T_{t=1}$. We denote the probability mass functions of  $J^{\pi}(S)$ , $\{S^\pi_t, O^\pi_t\}^T_{t=1}$ under instance $\ell$ as $P^{(\ell)}_{J^{\pi}(S)}, P^{(\ell)}_{\{S^\pi_t, O^\pi_t\}^T_{t=1}}$ respectively, for each $\ell\in \{0, 1, \ldots, L\}$. Now, we have
\begin{align}
&\frac{1}{L} \sum^L_{\ell = 1} \sum_{j\in [L]\setminus S^{*, \ell}}	\mathbb{E}^{(\ell)} \left[  \sum^T_{t=1} \mathsf{1}(j\in S^{\pi}_t)\right] \nonumber\\
= & \frac{1}{L} \sum^L_{\ell = 1} \sum_{j\in [L]}\mathbb{E}^{(\ell)} \left[  \sum^T_{t=1} \mathsf{1}(j\in S^{\pi}_t)\right]  - \frac{1}{L} \sum^L_{\ell = 1} \sum_{j\in  S^{*, \ell}}	\mathbb{E}^{(\ell)} \left[  \sum^T_{t=1} \mathsf{1}(j\in S^{\pi}_t)\right]\nonumber\\
= & KT - \frac{1}{L} \sum^L_{\ell = 1} \mathbb{E}^{(\ell)} \left[  \sum^T_{t=1} \sum_{j\in [L]}  \mathsf{1}(j\in S^{\pi}_t) \mathsf{1}(j\in S^{*, \ell}) \right] \nonumber\\
= & KT \left\{ 1  - \frac{1}{L} \sum^L_{\ell = 1} \mathbb{E}^{(\ell)}[J^{\pi }( S^{*, \ell}) ] \right\} \nonumber\\
 \geq &  KT \left\{ 1  - \frac{1}{L} \sum^L_{\ell = 1} \mathbb{E}^{(0)}[J^{\pi}( S^{*, \ell}) ] - \frac{1}{L}\sum^L_{\ell = 1} \sqrt{\frac{1}{2}\mathrm{KL}\left(P^{(0)}_{J^{\pi}(S^{*, \ell})} , P^{(\ell)}_{J^{\pi}(S^{*, \ell})}\right)} \right\} \label{cascadeTS_eq:by_pinsker}.\\
= & KT  \left\{ 1  - \frac{K}{L} - \frac{1}{L}\sum^L_{\ell = 1} \sqrt{\frac{1}{2} \mathrm{KL}\left(P^{(0)}_{J^{\pi}(S^{*, \ell})} , P^{(\ell)}_{J^{\pi}(S^{*, \ell})}\right) } \right\} \label{cascadeTS_eq:by_averaging} \\
\geq & K T \left\{ 1  - \frac{K}{L} - \frac{1}{L}\sum^L_{\ell =1} \sqrt{\frac{1}{2} \mathrm{KL}\left( P^{(0)}_{\{S^\pi_t, O^\pi_t\}^T_{t=1}}, P^{(\ell)}_{\{S^\pi_t, O^\pi_t\}^T_{t=1}} \right)}\right\} \label{cascadeTS_eq:by_full_pic}\\\
\geq & K T \left\{1  - \frac{K}{L} - \sqrt{\frac{1}{2  L}\sum^L_{\ell =1} \mathrm{KL}\left( P^{(0)}_{\{S^\pi_t, O^\pi_t\}^T_{t=1}}, P^{(\ell)}_{\{S^\pi_t, O^\pi_t\}^T_{t=1}} \right)}   \right\} \label{cascadeTS_eq:after_jens}.
\end{align}
Step~\eqref{cascadeTS_eq:by_pinsker} is by applying the following version of the Pinsker's inequality:
\begin{theorem}[Pinsker's Inequality]
Consider two probability mass functions $P_X, P_Y$ defined on the same discrete probability space ${\cal A} \subset [0, 1]$. The following inequalities hold:
\begin{align}
\left|\mathbb{E}_{X\sim P_X}[X] - \mathbb{E}_{Y\sim P_Y}[Y]\right| & \leq \sup_{A\subseteq {\cal A} } \left\{\sum_{a\in A}P_X(a) - \sum_{a\in A}P_Y(a)\right\} \leq  \sqrt{\frac{1}{2}\mathrm{KL}(P_X, P_Y)}\nonumber.
\end{align}
\end{theorem}
Step \eqref{cascadeTS_eq:by_averaging} is by the uniform cover property of the size-$K$ sets $\{S^{*, \ell}\}^L_{\ell = 1}$:
\begin{align}
\frac{1}{L}\sum^L_{\ell = 1} J_t^{\pi}(S^{*, \ell})& = \frac{1}{LKT} \sum^T_{t=1} \sum^L_{\ell = 1} \sum_{j\in [L]} \mathsf{1}(j\in S^\pi_t) \mathsf{1}(j\in S^{*, \ell}) \nonumber\\
&= \frac{1}{LKT} \sum^T_{t=1} \sum_{j\in [L]} \mathsf{1}(j\in S^\pi_t) \cdot \left[ \sum^L_{\ell = 1}  \mathsf{1}(j\in S^{*, \ell}) \right]\nonumber\\
&= \frac{1}{LT} \sum^T_{t=1} \sum_{j\in [L]} \mathsf{1}(j\in S^\pi_t) \label{cascadeTS_eq:by_uniform_cover}\\
&= \frac{K}{L}\nonumber.
\end{align}
Step \eqref{cascadeTS_eq:by_uniform_cover} is because each $j\in [L]$ lies in exactly $K$ sets in the collection $\{S^{*, \ell}\}^L_{\ell = 1}$.
Step~\eqref{cascadeTS_eq:by_full_pic} is by the fact that $J_t^{\pi}( S^{*, \ell})$ is $\sigma(\{ S^{\pi}_t, O^{\pi}_t \}^T_{t=1})$-measurable for any $\ell$, so that
$$
\mathrm{KL}\left(P^{(0)}_{J^{\pi}(S^{*, \ell})} , P^{(\ell)}_{J^{\pi}(S^{*, \ell})}\right) \leq  \mathrm{KL}\left( P^{(0)}_{\{S^\pi_t, O^\pi_t\}^T_{t=1}}, P^{(\ell)}_{\{S^\pi_t, O^\pi_t\}^T_{t=1}} \right).
$$
Finally, step~\eqref{cascadeTS_eq:after_jens} follows from Jensen's inequality. Altogether, the Lemma is proved.
\end{proof}

\subsection{Proof of Lemma~\ref{cascadeTS_claim:KLdecomp} }
\label{cascadeTS_pf:claimKLdecompL} 


\begin{proof}
In the following calculation, the notation $\sum_{\{ s_\tau, o_\tau \}^t_{\tau=1}}$ means $\sum_{\{ s_\tau, o_\tau \}^t_{\tau=1} \in ([L]^{(K)}\times {\cal O})^t }$. In addition, to ease the notation, we drop the subscript $\{S^\pi_t, O^\pi_t\}^T_{t=1}$ in the notation $P^{(\ell)}_{\{S^\pi_t, O^\pi_t\}^T_{t=1}}$, since in this section we are only works with the stochastic process $\{S^\pi_t, O^\pi_t\}$, but not $J^\pi_t(S)$. Lastly, we adopt the convention that $0\log (0/0) = 0$. Now,
\begin{align}
	&  \mathrm{KL}\left( P^{(0)}_{\{S^\pi_t, O^\pi_t\}^T_{t=1}}, P^{(\ell)}_{\{S^\pi_t, O^\pi_t\}^T_{t=1}} \right)\nonumber\\
 = & \sum_{\{ s_t, o_t \}^T_{t=1}} P^{(0)}\left[ \{ S^{\pi}_t, O^{\pi}_t \}^T_{t=1}= \{ s_t, o_t \}^T_{t=1}\right] \log \frac{ P ^{(0)}\left[\{ S^{\pi}_t, O^{\pi}_t \}^T_{t=1} = \{ s_t, o_t \}^T_{t=1}\right]}{P^{(\ell)} \left[\{ S^{\pi}_t, O^{\pi}_t \}^T_{t=1} = \{ s_t, o_t \}^T_{t=1}\right]}\nonumber\\
	= & \sum^T_{t=1} \sum_{\{ s_\tau, o_\tau \}^t_{\tau=1}} P ^{(0)}\left[\{ S^{\pi, 0}_\tau, O^{\pi, 0}_\tau \}^t_{\tau =1} = \{ s_\tau, o_\tau \}^t_{\tau=1}\right] 
	\nonumber  \\
	&   \hphantom{ \sum^T_{t=1} \sum_{\{ s_\tau, o_\tau \}^t_{\tau=1}} }  \cdot
	\log \frac{P^{(0)}\left[ (S^{\pi}_t, O^{\pi}_t) = (s_t, o_t) \mid \{S^{\pi}_\tau, O^{\pi}_\tau \}^{t-1}_{\tau=1} = \{ s_\tau, o_\tau \}^{t-1}_{\tau=1}\right]}{P^{(\ell)}\left[ (S^{\pi}_t, O^{\pi}_t) = (s_t, o_t) \mid \{S^{\pi}_\tau, O^{\pi}_\tau \}^{t-1}_{\tau=1} = \{ s_\tau, o_\tau \}^{t-1}_{\tau=1}\right]} \label{cascadeTS_eq:intermediate_kl}
\end{align}
Step \eqref{cascadeTS_eq:intermediate_kl} is by the Chain Rule for the KL divergence. To proceed from step \eqref{cascadeTS_eq:intermediate_kl}, we invoke the assumption that, policy $\pi$ is  non-anticipatory and deterministic, to simplify the conditional probability terms. For any $t, \ell$, we have  
\begin{align}
	&P^{(\ell)}\left[ (S^{\pi}_t, O^{\pi}_t) = (s_t, o_t) \mid \{S^{\pi}_\tau, O^{\pi}_\tau \}^{t-1}_{\tau=1} = \{ s_\tau, o_\tau \}^{t-1}_{\tau=1}\right]\nonumber\\
	= & P^{(\ell)}\left[ (S^{\pi}_t, O^{\pi}_t) = (s_t, o_t) \mid \{S^{\pi}_\tau, O^{\pi}_\tau \}^{t-1}_{\tau=1} = \{ s_\tau, o_\tau \}^{t-1}_{\tau=1}, S^{\pi}_t = \pi_t(\{ s_\tau, o_\tau \}^{t-1}_{\tau=1}) \right] \label{cascadeTS_eq:by_def_of_nad_policy}\\
	 = & P^{(\ell)}\left[ (S^{\pi}_t, O^{\pi}_t) = (s_t, o_t) \mid S^{\pi}_t = \pi_t(\{ s_\tau, o_\tau \}^{t-1}_{\tau=1}) \right] \label{cascadeTS_eq:by_model_assumption}\\
	= & \begin{cases} 
   P^{(\ell)}\left[O^{\pi}_t = o_t \mid S^{\pi}_t = s_t  \right] & \text{if } s_t = \pi_t(\{ s_\tau, o_\tau \}^{t-1}_{\tau=1}) \\
   0       & \text{otherwise } 
  \end{cases} \nonumber.
\end{align}
\noindent Step \eqref{cascadeTS_eq:by_def_of_nad_policy} is by our supposition on $\pi$, which implies $S^{\pi}_t = \pi_t(S^{\pi}_1, O^{\pi}_1, \ldots, S^{\pi}_{t-1}, O^{\pi}_{t-1})$ for each $\ell\in \{0, \ldots, L\}$, where $\pi_t$ is a deterministic function. Step \eqref{cascadeTS_eq:by_model_assumption} is by the model assumption that, conditioned on the arm $S^{\pi}_t$ pulled at time step $t$, the outcome $O^{\pi}_t$ is independent of the history from time $1$ to $t-1$.

By our adopted convention that $0\log (0/0)= 0$ (equivalently, we are removing the terms with $0\log(0/0)$ from the sum), we simplify the sum in \eqref{cascadeTS_eq:intermediate_kl} as follows:
\begin{align}
	& \text{\eqref{cascadeTS_eq:intermediate_kl}} = \sum^T_{t=1} \sum_{\{ s_\tau, o_\tau \}^t_{\tau=1}} \Bigg\{ 
	    P^{(0)}\left[O^{\pi}_t = o_t \mid S^{\pi}_t = s_t  \right] P^{(0)} \left[ \{S^{\pi}_\tau, O^{\pi}_\tau \}^{t-1}_{\tau=1} = \{ s_\tau, o_\tau \}^{t-1}_{\tau=1}\right]    \nonumber\\
	&\qquad \qquad \qquad  
	\cdot \mathsf{1}\left(s_t = \pi_t(\{ s_\tau, o_\tau \}^{t-1}_{\tau=1}) \right) 
	 \cdot \log\frac{P^{(0)}_{O^\pi_t | S^\pi_t }\left[O^{\pi}_t = o_t \mid S^{\pi}_t = s_t  \right]}{P^{(\ell)}_{O^\pi_t | S^\pi_t }\left[O^{\pi}_t = o_t \mid S^{\pi}_t = s_t  \right]}
	 \Bigg\}\nonumber\\
	&= \sum^T_{t=1} \sum_{ s_t, o_t}P^{(0)}[S^{\pi}_t = s_t] P^{(0)}\left[O^{\pi}_t = o_t \mid S^{\pi}_t = s_t  \right] \log\frac{P^{(0)}_{O^\pi_t | S^\pi_t }\left[O^{\pi}_t = o_t \mid S^{\pi}_t = s_t  \right]}{P^{(\ell)}_{O^\pi_t | S^\pi_t }\left[O^{\pi}_t = o_t \mid S^{\pi}_t = s_t  \right]} \nonumber\\
	&= \sum^T_{t=1} \sum_{ s_t \in [L]^{(K)}}P^{(0)}[S^{\pi}_t = s_t] 	\sum_{k=0}^{K} P^{(0)}_{O^\pi_t | S^\pi_t }\left[O^{\pi}_t = o^k \mid S^{\pi}_t = s_t  \right] \log\frac{P^{(0)}_{O^\pi_t | S^\pi_t }\left[O^{\pi}_t = o^k \mid S^{\pi}_t = s_t  \right]}{P^{(\ell)}_{O^\pi_t | S^\pi_t }\left[O^{\pi}_t = o^k \mid S^{\pi}_t = s_t  \right]}. \label{cascadeTS_eq:sum_KLs} \\
& = \sum_{t=1}^T \sum_{s_t\in [L]^{(K)}}P^{(0)}[S^{\pi}_t = s_t]\cdot \mathrm{KL}\left( P^{(0)}_{ O_t^{\pi}  \mid S_t^{\pi}}(\cdot \mid s_t)  \,\Big\|\,  P^{(\ell)}_{ O_t^{\pi }  \mid S_t^{\pi}}(\cdot \mid s_t)  \right). \label{cascadeTS_eq:KLdivs}
\end{align}
Altogether, the Lemma is proved.
\end{proof}

\subsection{Proof of Lemma~\ref{cascadeTS_claim:KLupperbound} }

\begin{proof}
Consider a fixed $\ell\in \{1, \ldots, L\}$, and recall our notation $\underline{a} = a$ if $1\leq a\leq L$, and $\underline{a} = a - L$ if $L < a \leq 2L$. Recall the weigh parameters $\{w^{(0)}(i)\}_{i\in [L]}, \{w^{(\ell)}(i)\}_{i\in [L]}$ of instances $0, \ell$:
\begin{itemize}
\item For $i \in \{\underline{\ell}, \underline{\ell + 1}, \ldots, \underline{\ell +K - 1}\}$, $w^{(\ell)}(i) = \frac{1 + \epsilon}{K}$ , while $w^{(0)}(i) = \frac{1 - \epsilon}{K}$ .
\item For $i \in [L]\setminus \{\underline{\ell}, \underline{\ell + 1}, \ldots, \underline{\ell +K - 1}\}$, $w^{(\ell)}(i) = w^{(0)}(i) = \frac{1 - \epsilon}{K}$ .
\end{itemize}
By an abuse of notation, we see $S^{*, \ell}, S$ as size-$K$ subsets (instead of order sets) when we take the intersection $S^{*, \ell}\cap S$. More precisely, for $S = (i_1, \ldots, i_k)$, we denote $S^{*, \ell}\cap S = \{\iota : \iota = \underline{j}\text{ for some $j\in \{\ell, \ldots, \ell + K - 1\}$, and }\iota = i_j \text{ for some $j\in \{1, \ldots, K\}$}\} $.

It is clear that, for any ordered $K$ set $S\in [L]^{(K)}$, 
\begin{align}
\mathrm{KL}\left( P^{(0)}_{ O_t^{\pi}  \mid S_t^{\pi}}(\cdot \mid S)  \,\Big\|\,  P^{(\ell)}_{ O_t^{\pi }  \mid S_t^{\pi}}(\cdot \mid S)  \right) &= \sum_{i\in S^{*, \ell}\cap S} P^{(0)}(\text{$i$ is observed}) \cdot  \mathrm{KL}\left(\frac{1-\epsilon}{K}, \frac{1 + \epsilon}{K}\right)\nonumber\\
& \leq  | S^{*, \ell} \cap S | \cdot  \mathrm{KL}\left(\frac{1-\epsilon}{K}, \frac{1 + \epsilon}{K}\right)\nonumber.
\end{align}
For any $S\in [L]^{(K)}$,
    \begin{align*}
        & \frac{1}{ L}\sum^L_{\ell =1}  | S^{*, \ell} \cap S |  
          = \frac{1}{ L}\sum^L_{\ell =1}    \sum_{i\in S } \mathsf{1}\{ i \in S^{*, \ell} \}  
        = \sum_{i\in S }    \frac{1}{ L}\sum^L_{\ell =1}  \mathsf{1}\{ i \in S^{*, \ell} \} 
        \\ &
         =  \sum_{i\in S} \frac{K}{L} 
         = K \cdot \frac{K}{L} = \frac{K^2}{L}.
    \end{align*}
Therefore,
	\begin{align*}
	& \frac{1}{ L}\sum^L_{\ell =1} \sum_{s_t\in [L]^{(K)}} P^{(0)} [S^{\pi}_t = s_t]\cdot \mathrm{KL}\left( P^{(0)}_{ O_t^{\pi}  \mid S_t^{\pi}}(\cdot \mid s_t)  \,\Big\|\,  P^{(\ell)}_{ O_t^{\pi }  \mid S_t^{\pi}}(\cdot \mid s_t)  \right)  \\
	 & =\frac{1}{ L}\sum^L_{\ell =1} \sum_{s_t\in [L]^{(K)}} P^{(0)}[S^{\pi, 0}_t = s_t]\cdot 
	    | S^{*, \ell} \cap s_t | \cdot  \mathrm{KL}\left(\frac{1-\epsilon}{K}, \frac{1 + \epsilon}{K}\right)  \\
	& =\mathrm{KL}\left(\frac{1-\epsilon}{K}, \frac{1 + \epsilon}{K}\right) \cdot
	\sum_{s_t\in [L]^{(K)}}P^{(0)}[S^{\pi, 0}_t = s_t]\cdot 
	     \left( \frac{1}{ L}\sum^L_{\ell =1}   | S^{*, \ell} \cap s_t |    \right) \\
	& =\mathrm{KL}\left(\frac{1-\epsilon}{K}, \frac{1 + \epsilon}{K}\right) \cdot  \frac{K^2}{L} 
	    = \frac{K}{L} \left[ (1-\epsilon) \log \left( \frac{ 1-\epsilon }{ 1+\epsilon } \right) + (K-1+\epsilon) \log \left(  \frac{K-1+\epsilon}{K-1-\epsilon} \right) \right].
\end{align*}


\end{proof}

\newpage

\section{Additional numerical results}
\label{cascadeTS_appdix:add_exp}

\begin{spacing}{1.5}
\begin{longtable}[H] 
{  m{0.25\linewidth} m{0.05\linewidth}  m{0.05\linewidth}  m{0.3\linewidth} m{0.2\linewidth} } 
  \caption{The performances of  {\sc TS-Cascade}, {\sc CTS}, {\sc CascadeUCB1}  and {\sc CascadeKL-UCB}  when $L\in \{ 16, 32, 64, 256, 512, 1024, 2048 \}$, $K \in \{2 ,4 \}$.  The first column shows the mean and the standard deviation of $\mathrm{Reg}(T)$ and the second column shows the average running time in seconds.}
  \\
    & $L$ & $K$ & $\mathrm{Reg}(T)$ & Running time \\
    \thickhline
    {\sc  TS-Cascade} & $ 16    $ & $ 2     $ & $ 4.10 \times 10^2 \pm 3.03 \times 10^1$ & $ 1.68  $  \\
    \hline
    {\sc CTS} & $ 16    $ & $ 2     $ & $ 1.61 \times 10^2  \pm 2.07 \times 10^1 $ & $ 4.96  $ \\ 
    \hline
    {\sc CascadeUCB1 }   & $ 16    $ & $ 2     $ & $ 1.35 \times 10^3  \pm 5.77 \times 10^1 $ & $ 1.33  $ \\
    \hline
    {\sc CascadeKL-UCB }& $ 16    $ & $ 2     $ & $ 1.64 \times 10^4  \pm 6.49 \times 10^3 $ & $ 2.18  $ \\
    \hhline{=====}

    {\sc  TS-Cascade} & $ 16    $ & $ 4     $ & $ 3.54 \times 10^2  \pm 2.70 \times 10^1 $ & $ 3.75  $ \\
    \hline
    {\sc CTS} & $ 16    $ & $ 4     $ & $ 1.20 \times 10^2  \pm 1.56 \times 10^1 $ & $ 6.29  $ \\
    \hline
    {\sc CascadeUCB1 }   & $ 16    $ & $ 4     $ & $ 1.10 \times 10^3  \pm 4.60 \times 10^1 $ & $ 1.23  $ \\
    \hline
    {\sc CascadeKL-UCB }& $ 16    $ & $ 4     $ & $ 2.02 \times 10^4  \pm 6.06 \times 10^3 $ & $ 4.84  $ \\
    \hhline{=====}

    {\sc  TS-Cascade} & $ 32    $ & $ 2     $ & $ 7.55 \times 10^2  \pm 2.38 \times 10^1 $ & $ 1.85  $ \\
    \hline
    {\sc CTS} & $ 32    $ & $ 2     $ & $ 3.30 \times 10^2  \pm 2.84 \times 10^1 $ & $ 5.93  $ \\
    \hline
    {\sc CascadeUCB1 }   & $ 32    $ & $ 2     $ & $ 2.77 \times 10^3  \pm 9.72 \times 10^1 $ & $ 1.53  $ \\
    \hline
    {\sc CascadeKL-UCB }& $ 32    $ & $ 2     $ & $ 1.86 \times 10^4  \pm 7.04 \times 10^3 $ & $ 2.33  $ \\
    \hhline{=====}

    {\sc  TS-Cascade} & $ 32    $ & $ 4     $ & $ 6.52 \times 10^2  \pm 2.95 \times 10^1 $ & $ 1.85  $ \\
    \hline
    {\sc CTS} & $ 32    $ & $ 4     $ & $ 2.84 \times 10^2  \pm 8.25 \times 10^1 $ & $ 5.88  $ \\
    \hline
    {\sc CascadeUCB1 }   & $ 32    $ & $ 4     $ & $ 2.36 \times 10^3  \pm 5.53 \times 10^1 $ & $ 1.52  $ \\
    \hline
    {\sc CascadeKL-UCB }& $ 32    $ & $ 4     $ & $ 2.21 \times 10^4  \pm 7.13 \times 10^3 $ & $ 2.90  $ \\
    \hhline{=====}

    {\sc  TS-Cascade} & $ 64    $ & $ 2     $ & $ 1.39 \times 10^3  \pm 3.37 \times 10^1 $ & $ 2.23  $ \\
    \hline
    {\sc CTS} & $ 64    $ & $ 2     $ & $ 6.67 \times 10^2  \pm 3.43 \times 10^1 $ & $ 7.85  $ \\
    \hline
    {\sc CascadeUCB1 }   & $ 64    $ & $ 2     $ & $ 5.52 \times 10^3  \pm 1.11 \times 10^2 $ & $ 2.05  $ \\
    \hline
    {\sc CascadeKL-UCB }& $ 64    $ & $ 2     $ & $ 2.17 \times 10^4  \pm 8.03 \times 10^3 $ & $ 2.94  $ \\
    \hhline{=====}

    {\sc  TS-Cascade} & $ 64    $ & $ 4     $ & $ 1.22 \times 10^3  \pm 3.08 \times 10^1 $ & $ 2.23  $ \\
    \hline
    {\sc CTS} & $ 64    $ & $ 4     $ & $ 5.27 \times 10^2  \pm 2.89 \times 10^1 $ & $ 7.83  $ \\
    \hline
    {\sc CascadeUCB1 }   & $ 64    $ & $ 4     $ & $ 4.75 \times 10^3  \pm 7.54 \times 10^1 $ & $ 2.06  $ \\
    \hline
    {\sc CascadeKL-UCB }& $ 64    $ & $ 4     $ & $ 2.72 \times 10^4  \pm 8.71 \times 10^3 $ & $ 3.32  $ \\
    \hhline{=====}

    {\sc  TS-Cascade} & $ 128   $ & $ 2     $ & $ 2.46 \times 10^3  \pm 4.63 \times 10^1 $ & $ 3.08  $ \\
    \hline
    {\sc CTS} & $ 128   $ & $ 2     $ & $ 1.36 \times 10^3  \pm 4.39 \times 10^1 $ & $ 1.17 \times 10^1 $ \\
    \hline
    {\sc CascadeUCB1 }   & $ 128   $ & $ 2     $ & $ 1.07 \times 10^4  \pm 1.70 \times 10^2 $ & $ 3.74  $ \\
    \hline
    {\sc CascadeKL-UCB }& $ 128   $ & $ 2     $ & $ 2.19 \times 10^4  \pm 8.11 \times 10^3 $ & $ 4.49  $ \\
    \hhline{=====}

    {\sc  TS-Cascade} & $ 128   $ & $ 4     $ & $ 2.25 \times 10^3  \pm 5.95 \times 10^1 $ & $ 3.09  $ \\
    \hline
    {\sc CTS} & $ 128   $ & $ 4     $ & $ 1.06 \times 10^3  \pm 4.95 \times 10^1 $ & $ 1.17 \times 10^1 $ \\
    \hline
    {\sc CascadeUCB1 }   & $ 128   $ & $ 4     $ & $ 9.23 \times 10^3  \pm 1.27 \times 10^2 $ & $ 3.58  $ \\
    \hline
    {\sc CascadeKL-UCB }& $ 128   $ & $ 4     $ & $ 2.95 \times 10^4  \pm 1.05 \times 10^4 $ & $ 4.71  $ \\
    \hhline{=====}

    {\sc  TS-Cascade} & $ 256   $ & $ 2     $ & $ 4.31 \times 10^3  \pm 2.27 \times 10^2 $ & $ 4.73  $ \\
    \hline
    {\sc CTS} & $ 256   $ & $ 2     $ & $ 2.76 \times 10^3  \pm 2.37 \times 10^2 $ & $ 1.93 \times 10^1 $ \\
    \hline
    {\sc CascadeUCB1 }   & $ 256   $ & $ 2     $ & $ 1.92 \times 10^4 \pm 2.73 \times 10^2 $ & $ 7.57  $ \\
    \hline
    {\sc CascadeKL-UCB }& $ 256   $ & $ 2     $ & $ 2.19 \times 10^4  \pm 8.11 \times 10^3 $ & $ 7.24  $ \\
    \hhline{=====}

    {\sc  TS-Cascade} & $ 256   $ & $ 4     $ & $ 3.99 \times 10^3  \pm 1.37 \times 10^2 $ & $ 4.79  $ \\
    \hline
    {\sc CTS} & $ 256   $ & $ 4     $ & $ 2.16 \times 10^3  \pm 8.27 \times 10^1 $ & $ 1.93 \times 10^1 $ \\
    \hline
    {\sc CascadeUCB1 }   & $ 256   $ & $ 4     $ & $ 1.74 \times 10^4 \pm 1.65 \times 10^2 $ & $ 6.59  $ \\
    \hline
    {\sc CascadeKL-UCB }& $ 256   $ & $ 4     $ & $ 2.98 \times 10^4  \pm 1.03 \times 10^4 $ & $ 7.48  $ \\
    \hhline{=====}

    {\sc  TS-Cascade} & $ 512   $ & $ 2     $ & $ 6.83 \times 10^3  \pm 5.28 \times 10^2 $ & $ 8.05  $ \\
    \hline
    {\sc CTS} & $ 512   $ & $ 2     $ & $ 5.39 \times 10^3  \pm 1.92 \times 10^2 $ & $ 3.49 \times 10^1 $ \\
    \hline
    {\sc CascadeUCB1 }   & $ 512   $ & $ 2     $ & $ 2.51 \times 10^4  \pm 1.47 \times 10^2 $ & $ 1.30 \times 10^1 $ \\
    \hline
    {\sc CascadeKL-UCB }& $ 512   $ & $ 2     $ & $ 2.19 \times 10^4  \pm 8.10 \times 10^3 $ & $ 1.31 \times 10^1 $ \\
    \hhline{=====}

    {\sc  TS-Cascade} & $ 512   $ & $ 4     $ & $ 6.77 \times 10^3  \pm 2.00 \times 10^2 $ & $ 8.20  $ \\
    \hline
    {\sc CTS} & $ 512   $ & $ 4     $ & $ 4.27 \times 10^3  \pm 1.73 \times 10^2 $ & $ 3.50 \times 10^1 $ \\
    \hline
    {\sc CascadeUCB1 }   & $ 512   $ & $ 4     $ & $ 3.03 \times 10^4  \pm 3.33 \times 10^2 $ & $ 1.27 \times 10^1 $ \\
    \hline
    {\sc CascadeKL-UCB }& $ 512   $ & $ 4     $ & $ 2.98 \times 10^4  \pm 1.03 \times 10^4 $ & $ 1.34 \times 10^1 $ \\
    \hhline{=====}

    {\sc  TS-Cascade} & $ 1024  $ & $ 2     $ & $ 9.82 \times 10^3  \pm 6.82 \times 10^2 $ & $ 1.47 \times 10^1 $ \\
    \hline
    {\sc CTS} & $ 1024  $ & $ 2     $ & $ 1.06 \times 10^4  \pm 2.26 \times 10^2 $ & $ 6.73 \times 10^1 $ \\
    \hline
    {\sc CascadeUCB1 }   & $ 1024  $ & $ 2     $ & $ 2.60 \times 10^4  \pm 3.32 \times 10^1 $ & $ 1.01 \times 10^2 $ \\
    \hline
    {\sc CascadeKL-UCB }& $ 1024  $ & $ 2     $ & $ 2.19 \times 10^4  \pm 8.05 \times 10^3 $ & $ 2.57 \times 10^1 $ \\
    \hhline{=====}

    {\sc  TS-Cascade} & $ 1024  $ & $ 4     $ & $ 1.06 \times 10^4  \pm 6.23 \times 10^2 $ & $ 1.51 \times 10^1 $ \\
    \hline
    {\sc CTS} & $ 1024  $ & $ 4     $ & $ 8.59 \times 10^3  \pm 2.94 \times 10^2 $ & $ 6.84 \times 10^1 $ \\
    \hline
    {\sc CascadeUCB1 }   & $ 1024  $ & $ 4     $ & $ 3.87 \times 10^4  \pm 1.32 \times 10^2 $ & $ 2.56 \times 10^1 $ \\
    \hline
    {\sc CascadeKL-UCB }& $ 1024  $ & $ 4     $ & $ 3.00 \times 10^4  \pm 1.01 \times 10^4 $ & $ 4.07 \times 10^1 $ \\
    \hhline{=====}

    {\sc  TS-Cascade} & $ 2048  $ & $ 2     $ & $ 1.27 \times 10^4  \pm 1.72 \times 10^3 $ & $ 4.40 \times 10^1 $ \\
    \hline
    {\sc CTS} & $ 2048  $ & $ 2     $ & $ 1.99 \times 10^4  \pm 5.85 \times 10^2 $ & $ 1.25 \times 10^2 $ \\
    \hline
    {\sc CascadeUCB1 }   & $ 2048  $ & $ 2     $ & $ 2.62 \times 10^4  \pm 9.19  $ & $ 5.67 \times 10^1 $ \\
    \hline
    {\sc CascadeKL-UCB }& $ 2048  $ & $ 2     $ & $ 2.19 \times 10^4  \pm 8.05 \times 10^3 $ & $ 1.08 \times 10^2 $ \\
    \hhline{=====}

    {\sc  TS-Cascade} & $ 2048  $ & $ 4     $ & $ 1.50 \times 10^4  \pm 7.27 \times 10^2 $ & $ 4.55 \times 10^1 $ \\
    \hline
    {\sc CTS} & $ 2048  $ & $ 4     $ & $ 1.67 \times 10^4  \pm 3.99 \times 10^2 $ & $ 1.25 \times 10^2 $ \\
    \hline
    {\sc CascadeUCB1 }   & $ 2048  $ & $ 4     $ & $ 4.01 \times 10^4  \pm 4.18 \times 10^1 $ & $ 5.57 \times 10^1 $ \\
    \hline
    {\sc CascadeKL-UCB }& $ 2048  $ & $ 4     $ & $ 3.00 \times 10^4  \pm 1.00 \times 10^4 $ & $ 5.47 \times 10^1 $ \\
    \hline
  \label{cascadeTS_tab:nonlin_toy_probSet5}%
\end{longtable}%
\end{spacing}

\end{document}